%% file: main.tex
\documentclass{article}

\usepackage{microtype}
\usepackage{booktabs} 

\usepackage{hyperref}
\usepackage{xcolor}
\usepackage[normalem]{ulem}
\usepackage{dsfont}

\usepackage{url}
\usepackage{bm}
\usepackage{graphicx}
\usepackage{xspace}
\usepackage{makecell}
\usepackage{array}
\usepackage{colortbl}
\usepackage{multirow}
\usepackage{subcaption}
\usepackage{tabularx}
\usepackage{color}
\usepackage{pifont}
\usepackage{marvosym} 
\usepackage[utf8]{inputenc}
\usepackage[T1]{fontenc}
\usepackage{CJKutf8}
\usepackage{enumitem}
\usepackage{diagbox}
\usepackage{listings}
\usepackage[flushleft]{threeparttable}
\usepackage{booktabs}
\usepackage{wrapfig}

\newcommand{\emoji}[2][1em]{\raisebox{-0.2\height}{\includegraphics[height=#1]{#2}}}
\definecolor{fbApp}{HTML}{ffe4e3}
\definecolor{mydarkblue}{rgb}{0,0.3,0.9}

\newcommand{\icon}{\raisebox{-3.5pt}{\includegraphics[width=1.1em]{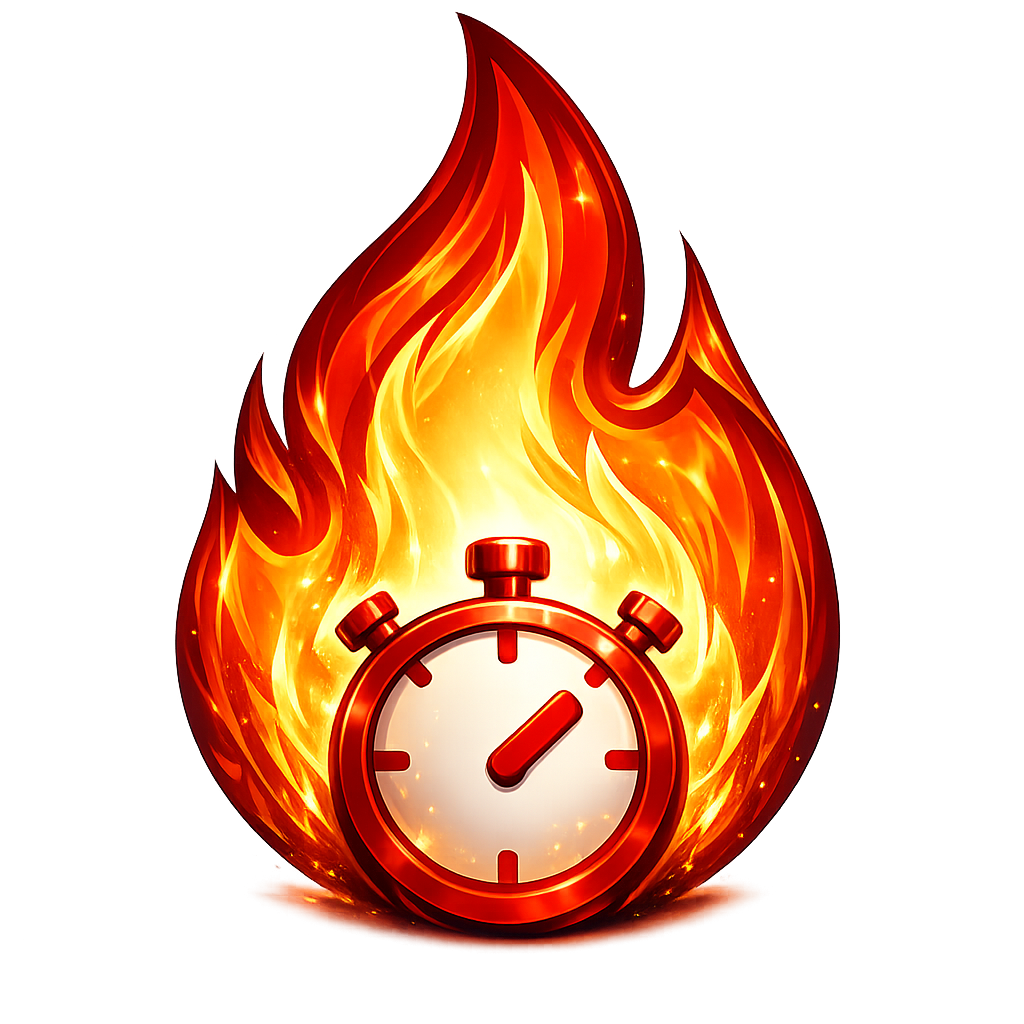}}\xspace}

\newcommand{\rowc}{\rowcolor{fbApp}}

\PassOptionsToPackage{numbers,compress}{natbib}

 \usepackage[main, final]{neurips-2026}

\usepackage{amsmath}
\usepackage{amssymb}
\usepackage{mathtools}
\usepackage{amsthm}

\usepackage[capitalize,noabbrev]{cleveref}

\theoremstyle{plain}
\newtheorem{theorem}{Theorem}[section]
\newtheorem{proposition}[theorem]{Proposition}

\newtheorem{corollary}[theorem]{Corollary}
\theoremstyle{definition}

\theoremstyle{remark}

\usepackage[textsize=tiny]{todonotes}

\title{\icon FLAME: Flow Enhanced Legendre Memory \\ \quad Models for General Time Series Forecasting}

\author{%
  Xingjian Wu\textsuperscript{*}, Zhengyu Li\textsuperscript{*}, Hanyin Cheng\textsuperscript{*}, Xiangfei Qiu\textsuperscript{*}, \\ 
  \textbf{Jilin Hu}, \textbf{Chenjuan Guo}, \textbf{Bin Yang}\textsuperscript{\Letter}\\
  East China Normal University\\
\texttt{\{xjwu,lizhengyu,hycheng,xfqiu\}@stu.ecnu.edu.cn}, \\
\texttt{\{jlhu,cjguo,byang\}@dase.ecnu.edu.cn}
}

\begin{document}

\maketitle

\begin{abstract}
  In this work, we introduce \textbf{FLAME}, a family of extremely lightweight and capable Time Series Foundation Models, which support versatile forecasting tasks via generative probabilistic modeling, while ensuring both efficiency and robustness. FLAME utilizes the Legendre Memory for strong generalization capabilities. By adapting variants of Legendre Memory, i.e., translated Legendre (LegT) and scaled Legendre (LegS), in the Encoding and Decoding phases, FLAME can effectively capture the inherent inductive bias within data and make efficient long-range inferences. To enhance the accuracy of probabilistic forecasting while remaining efficient, FLAME adopts a normalizing-flow-based forecasting head, which can model complex distributions over the forecasting horizon in a generative manner. Comprehensive experiments on three well-recognized benchmarks, including TSFM-Bench, ProbTS, and TFB, demonstrate that FLAME is a strong out-of-the-box tool for decision intelligence.
\end{abstract}
\begin{center}
\emoji{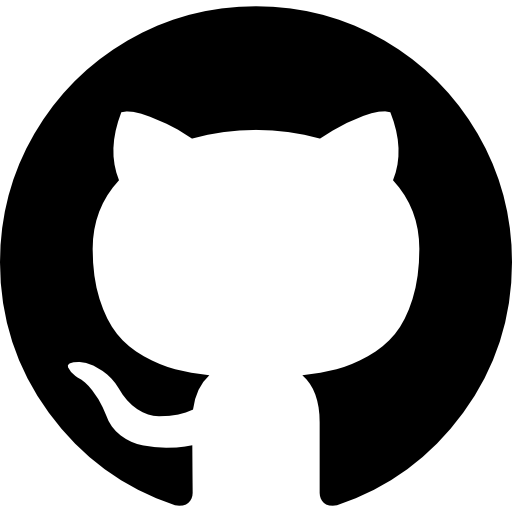} \href{https://github.com/decisionintelligence/FLAME}{https://github.com/decisionintelligence/FLAME}.\\
\emoji{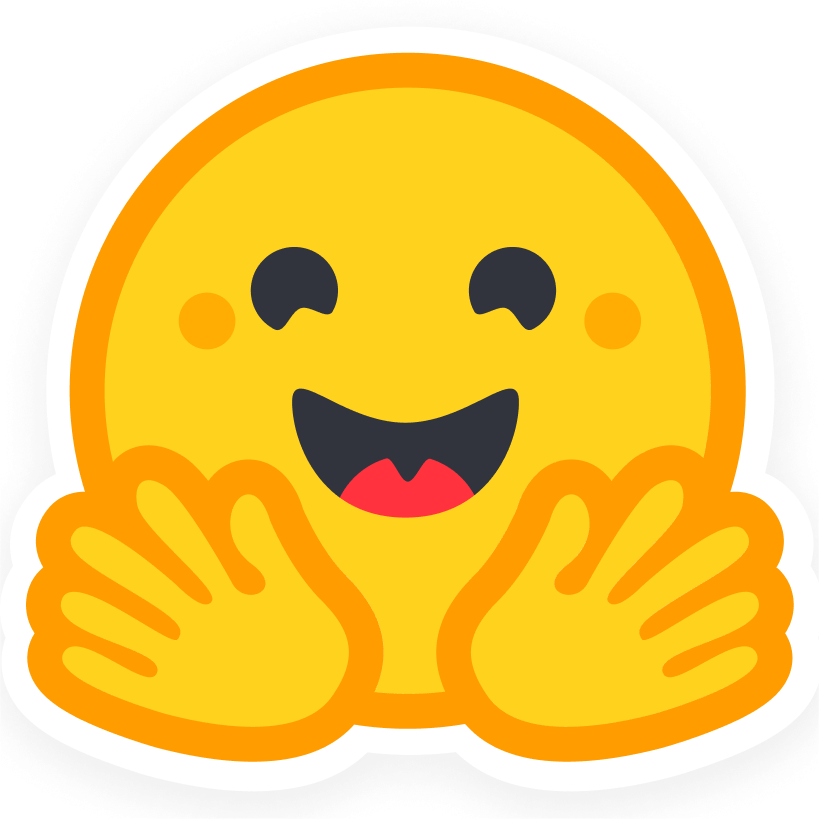} \href{https://huggingface.co/DecisionIntelligence/FLAME}{https://huggingface.co/DecisionIntelligence/FLAME}.
\end{center}

\input{Sections/Introduction}

\input{Sections/Preliminaries}

\input{Sections/Methodology}

\input{Sections/Experiments}

\input{Sections/Related-Works}
\input{Sections/Conclusion}

\clearpage

\begin{ack}
This work was partially supported by National Natural Science Foundation of China (62372179).

Xingjian Wu, Zhengyu Li, Hanyin Cheng, and Xiangfei Qiu contributed equally to this work. Bin Yang is the corresponding author.
\end{ack}

\setlength{\bibhang}{0pt}
\bibliography{reference}
\bibliographystyle{unsrtnat}

\newpage
\appendix
\input{Sections/Appendix}

\newpage
\input{Sections/checklist}

\end{document}

%% file: Sections/Introduction.tex
\section{Introduction}

\begin{wrapfigure}[19]{r}{0.5\textwidth}
\vspace{-4mm}
  \centering
\includegraphics[width=1\linewidth]{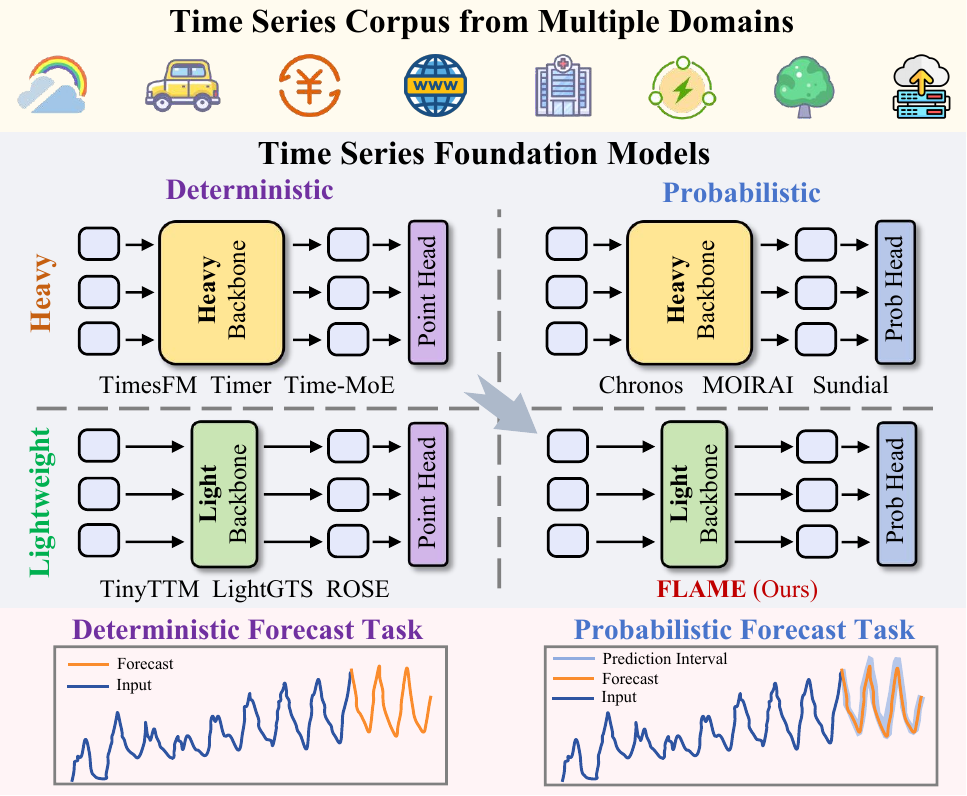}
  \caption{FLAME is pretrained on Time Series Corpus from Multiple Domains, possessing a \textit{lightweight} backbone while supporting generative \textit{probabilistic} forecasting. }
  \label{fig: intro}
  \vspace{-4mm}
\end{wrapfigure}
Time Series Forecasting is a critical and widely studied task due to the attractive charm of peeking into the future, which creates value across fields by supporting planning. Time series can fluctuate like a \textit{flame}, making their underlying dynamics difficult to capture. Therefore, to control risks and make reasonable decisions, quantifying uncertainties through Probabilistic Time Series Forecasting (PTSF) is a more reliable approach, which has been explored for decades from statistical approaches~\citep{godahewa2021monash,arima} with solid theories and interpretability to data-driven approaches~\citep{nie2022time,qiu2025duet, wu2025k2vae} with strong fitting capability.
Among them, deep learning models~\citep{nie2022time,qiu2025duet,wu2025k2vae,FOTraj} show impressive performance on in-distribution tasks through effectively capturing the inherent dynamics within raw data. 
Inspired by the success of pretrained models in other domains such as CV~\citep{khan2022visiontransformers, liu2021swin} and NLP~\citep{liu2024deepseek,achiam2023gpt} with strong generalization capabilities, a surge of pretrained time series foundation models~\citep{timer,timesfm,units,timemoe,wang2025lightgts,wang2025rose} have been developed for out-of-distribution tasks, demonstrating promising trends.

Recent studies on time series foundation models~\citep{woo2024moirai,timemoe,wang2025lightgts,liu2025sundial} have achieved some progress in exploiting the generalization capabilities and exhibit excellent forecasting performance in zero-shot scenarios. Their generalization capabilities mainly come from two perspectives--see Figure~\ref{fig: intro}: 1) some works make efforts in devising appropriate Transformer-based architectures, and expanding the parameters to hundreds of millions or billions of parameters~\citep{timemoe,ansarichronos,woo2024moirai,timesfm}, or with the help of the massive parameters from LLMs~\citep{gpt4ts, timellm} and ViTs~\citep{chen2024visionts,DBLP:conf/icml/ChenS0WSL25}. Though these works achieve excellent performance in zero-shot scenarios, they incur substantial inference costs under limited resources, and some of them are even slower than training end-to-end supervised models from scratch, thus \textit{hindering the efficiency and practicality}; 2) another line of work explores the effective capture of inductive bias within data, with meticulously-designed lightweight structures~\citep{tinyttm,units,wang2025lightgts} but strong generalization capabilities. However, although they demonstrate strong performance in deterministic forecasting tasks, they \textit{lack the support of probabilistic forecasting}. Moreover, predicting accurate distributions over the forecasting horizon is difficult because assuming prior distributions like Isotropic Gaussian Distributions constrains the modeling capabilities of real data distributions. On the other hand, modeling arbitrarily complex distributions relies on generative mechanisms~\citep{li2017diffusion,TSDiff}, which may increase the model complexity and hinder the efficiency.

As shown in Figure~\ref{fig: intro}, our proposed \textbf{FLAME} pioneers the exploration of \textit{lightweight} time series foundation models capable of \textit{both deterministic and probabilistic forecasting}. To enhance the generalization capability while remaining lightweight, we utilize the \textit{Legendre Memory}~\citep{voelker2019legendre,gu2020hippo} in both Encoding and Decoding phases to capture the inherent inductive bias within data, which contributes to a lightweight but strong framework. Originating from ordinary differential equations (ODEs) and characterized by orthogonal Legendre polynomials, the Legendre Memory has been utilized in control theory to model continuous time-delay systems~\citep{richard2003timedelaysystems} for several decades, which has the potential to compress time series of varying lengths into vectors of the same size. This helps keep the model lightweight, and we employ Legendre Memory in both the encoder and decoder.

First, in the pretraining process of time series foundation models, the inherent periods and trends of data from distinct domains can lead to complex local characteristics but the patching technique cannot handle this due to the fixed receptive field. This inspires us to devise a \textit{Local-Perception} module to unlock the power of Legendre Memory in integrating intricate and varying local characteristics within data as supplementary environmental information for patch tokens. Second, we also adopt the Structured State-Space Duality (SSD~\citep{mamba2}) layers in the decoding phase, which utilize Legendre Memory to initialize the state transition matrices~\citep{gu2020hippo}, excelling at modeling long time series based on its selective states and adaptive memory utilization.

Besides the adaptations of Legendre Memory, we also devise a \textit{Flow-based Head} to support generative probabilistic forecasting. Specifically, we focus on token-wise modeling, which constructs the distributions over dozens of time series points at once, and using the Flow-based Head driven by normalizing flow~\citep{normalizing-flow,nsf} can preserve both considerable efficiency and interpretability compared with prevailing diffusion models~\citep{TSDiff,TimeGrad}. Additionally, we also aim to capture the within-token autoregressive dependencies in the Flow-based Head, which is important but neglected in current decoding methods~\citep{liu2025sundial,timer,timemoe}, and this can help preserve temporal dependencies within tokens to boost the fine-grained modeling. Through these adaptations, the Flow-based Head can preserve both the accuracy and efficiency through constructing lightweight autoregressive affine layers to simulate the reversible conditional transfer process from fixed prior Gaussian distributions to flexible predictive distributions. With the Legendre Memory and Flow-based Head facilitating the strong generative paradigm, FLAME can make extremely fast and accurate probabilistic forecasts. Our contributions are summarized as:

\begin{itemize}[left=0.1cm]
\item We present \textbf{FLAME}, a family of \underline{\bf{Fl}}ow enh\underline{\bf{a}}nced Legendre \underline{\bf{Me}}mory Foundation Models, including three parameter versions, i.e., 2M, 6M, 10M. Even the largest version of FLAME (10M) is far more lightweight than other time series foundation models which support probabilistic forecasting.

\item We adapt the Legendre Memory to enhance the generalization capability. Both the Local-Perception Module and SSD-Decoder are driven by Legendre Memory and able to adaptively utilize the inductive bias within data. 

\item We devise a Flow-based Head to model the intricate token-wise distributions over the forecasting horizon, considering the causality within tokens, with strong efficiency and accuracy.

\item Experimentally, FLAME achieves state-of-the-art zero-shot performance on both deterministic and probabilistic forecasting tasks, including TSFM-Bench~\citep{li2025TSFM-Bench}, ProbTS~\citep{ProbTS}, and TFB~\citep{qiu2024tfb}, demonstrating a strong out-of-the-box tool for decision intelligence.

\end{itemize}

%% file: Sections/Preliminaries.tex
\section{Legendre Memory}
\label{sec: legendre}
When modeling sequence data with recurrent networks, ``memory'' is a common concept denoting the online compression capability of historical observations. Legendre Memory~\citep{voelker2019legendre,gu2020hippo} is a method for continuous online function approximation, which utilizes orthogonal Legendre polynomials to compress the continuous-time history $f$ into fixed-size state vectors $m$. From the perspective of \textit{Real Analysis}, given a time-varying \textit{measure} $\mu^{(t)}$ and a \textit{high-order polynomial operator}, recent history $f$ can be compressed into state vectors $m$ with bounded approximation error~\citep{gu2020hippo}. When the operator is the Legendre polynomial, there exist two common time-varying measures: \textit{translated Legendre (LegT)} and \textit{scaled Legendre measure (LegS)}.

LegT assigns uniform weights to the most recent history $[t-\theta, t]$, where $\theta$ denotes the length of the sliding window, or the length of history that is being summarized as
$
    \text{LegT}: \mu^{(t)}(x) = \frac{1}{\theta}\mathbb{I}_{[t-\theta,t]}(x).
$
Under the LegT measure, the process of compressing the historical information $f$ into state vectors $m$ can be represented as solving $d$ coupled linear time-invariant ODEs:
\begin{gather}
    m^\prime(t) = -Am(t) + Bf(t),
    \label{fu: LMU}
    ~
    B_n = \frac{1}{\theta} (2n+1)(-1)^n,\\
    A_{nk} = \frac{1}{\theta}\begin{cases}
(-1)^{n-k}(2n + 1) & \text{if } n > k, \\
2n+1 & \text{if } n\leq k
\end{cases} \ ,\label{fu: init legT}
\end{gather}
where $n,k\in\{0,\ldots,d-1\}$, and $A \in \mathbb{R}^{d\times d}, B\in \mathbb{R}^{d\times 1}$ are parameters derived using a Pad\'{e} approximant~\citep{pade,voelker2019dynamical}. In the forward process, the history over $[t-\theta, t]$ can be compressed into the memory cell $m(t) \in \mathbb{R}^d$ recurrently using the discretized form of Equation~(\ref{fu: LMU}):
\begin{gather}
    m_t = \bar{A}m_{t-1} + \bar{B}f_t,~
    \bar{A} = I - \Delta t A, \quad \bar{B} = \Delta t B, \label{fu: dis2}
\end{gather}
where $\bar{A}$ and $\bar{B}$ are obtained through forward Euler discretization; the factor $1/\theta$ is already included in $A$ and $B$. The ``memory'' lies in the capability of using $m(t)$ to approximate the history $f$ over $[t-\theta, t]$, thereby approximating a $\theta$ time-delay system~\citep{richard2003timedelaysystems}:
\begin{equation}
\begin{gathered}
    f(t-\theta^\prime)\approx \sum_{i=0}^{d-1}\mathcal{P}_i\left(\frac{\theta^\prime}{\theta}\right)m_i(t),\qquad 0\leq\theta^\prime\leq\theta,\\
    \mathcal{P}_i(r) = (-1)^i\sum_{j=0}^i\binom{i}{j}\binom{i+j}{j}(-r)^j,
\end{gathered}
\label{eq:legt_reconstruction}
\end{equation}
where $\mathcal{P}_i(r)$ is the $i$-th shifted Legendre polynomial~\citep{voelker2019legendre}. Here $r=\theta^\prime/\theta$ is the normalized lag: $r=0$ corresponds to the newest point in the window and $r=1$ to the oldest. The same $d$ memory coefficients therefore provide approximate reconstructions at different lags by evaluating the corresponding basis functions. Theorem~\ref{theory: 1}~\citep{gu2020hippo,zhou2022film} bounds the error of the exact polynomial projection as the number $d$ of Legendre basis functions increases. The finite-state recurrence and its discretization introduce additional approximation errors. We provide a detailed proof of the projection bound in Appendix~\ref{app:legt_proof}.

\begin{theorem}
\label{theory: 1}

Let $F: \mathbb{R}_+ \to \mathbb{R}$ be a differentiable function, $\theta>0$, $t\geq\theta$, and $d\geq1$. Let $f_{[t-\theta,t]}$ denote the \textbf{HiPPO-LegT projection} of $F$, defined as its optimal polynomial approximation of degree at most $d-1$ under the uniform probability measure $\mu^{(t)}$ on $[t-\theta,t]$. If $F$ is $L$-Lipschitz on this window, then $\Vert F-f_{[t-\theta,t]}\Vert_{\mu^{(t)}}=\mathcal{O}(\theta L/\sqrt{d})$. Moreover, for a fixed integer $k\geq1$, if $F\in C^k([t-\theta,t])$, $M_k=\sup_{x\in[t-\theta,t]}|F^{(k)}(x)|<\infty$, and $d\geq k$, then $\Vert F-f_{[t-\theta,t]}\Vert_{\mu^{(t)}}=\mathcal{O}(M_k\theta^k d^{-k+1/2})$, with the implied constant depending only on $k$.
\end{theorem}

Compared with LegT, LegS assigns uniform weights to all history $[0, t]$ to support the compression of the entire history: $\text{LegS}: \mu^{(t)}(x) = \frac{1}{t}\mathbb{I}_{[0,t]}(x)$.

Under the LegS measure, the compression process is:
\begin{gather}
    m^\prime(t) = -\frac{1}{t}Am(t) + \frac{1}{t}Bf(t),
    \label{fu: ssd}~
    B_n=(2n+1)^{\frac{1}{2}}, \\
    A_{nk} = \begin{cases}
        (2n+1)^{1/2}(2k+1)^{1/2} & \text{if } n > k,\\
        n+1 & \text{if } n = k,\\
        0 & \text{if } n < k
    \end{cases}\
    \label{fu: init legS}
\end{gather}
where $A\in\mathbb{R}^{d\times d}$ and $B\in \mathbb{R}^{d\times 1}$, with $n,k\in\{0,\ldots,d-1\}$. Unlike LegT, LegS is time-varying: a forward Euler step from time $\tau>0$ uses $\bar{A}_{\tau}=I-(\Delta t/\tau)A$ and $\bar{B}_{\tau}=(\Delta t/\tau)B$.



%% file: Sections/Methodology.tex
\section{FLAME}
\begin{figure*}[!htbp]
    \centering
    \includegraphics[width=0.97\linewidth]{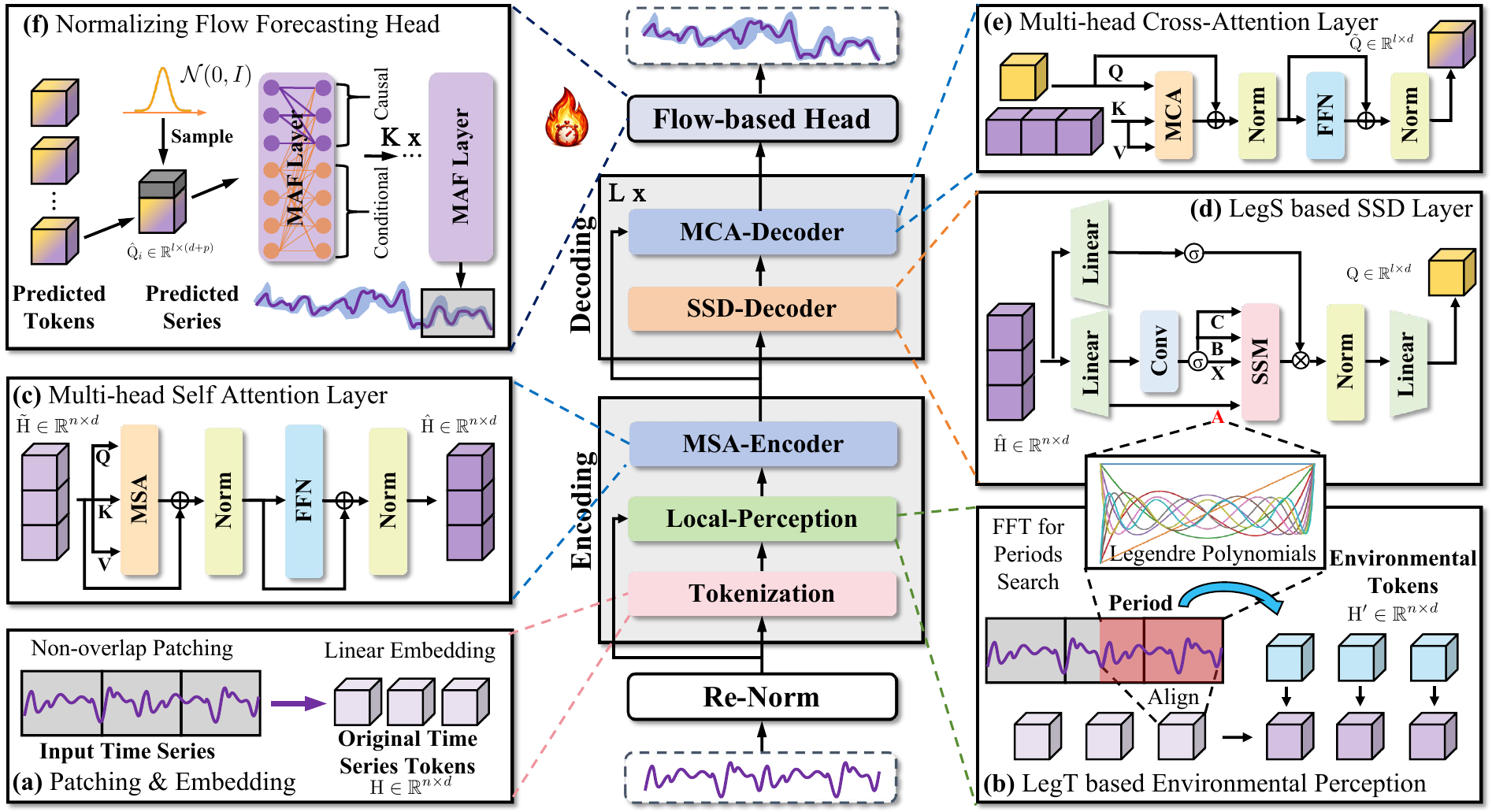}
    \caption{The overall framework of FLAME. The data flow follows the sequence of (a) - (f). }
\label{fig: overview}
\end{figure*}

FLAME adopts the Channel-Independent~\citep{nie2022time} pretraining paradigm, and each variable is preprocessed through Instance Normalization~\citep{ulyanov2016instance} to mitigate the value discrepancy. As shown in Figure~\ref{fig: overview}, FLAME utilizes the Re-Norm~\citep{kim2021reversible} to further mitigate the statistical differences between inputs and forecasts, and its backbone mainly consists of three modules: 1) Encoding, including Time Series Tokenization, Local-Perception, and MSA-Encoder, which tokenize the time series and enhance them through fusing the local environmental information with LegT; 2) Decoding, including LegS based SSD-Decoder and MCA-Decoder, which utilize the SSD layers and MCA layers to make long-term inference; 3) Flow-based Head, which leverages the Normalizing Flow to support generative probabilistic forecasting.

\subsection{Encoding}

\subsubsection{Time Series Tokenization}
In the pretraining Time Series Corpus, all data are preprocessed into univariate time series and received by FLAME in a Channel-Independent paradigm. Following the commonly used patching technique, which was introduced by Triformer~\citep{Triformer}, we first divide each univariate time series $\mathrm{X}\in\mathbb{R}^T$ into $n$ non-overlapping time series patches $ \left[x_1, x_2, \cdots, x_n \right]$,
where $x_i\in\mathbb{R}^{p}$ is a patch of size $p$, and $n=\lceil T/p\rceil$ is the number of patches. If $T$ is not divisible by $p$, the boundary patch is padded to length $p$. Subsequently, we use a linear layer: $\mathbb{R}^p \to \mathbb{R}^d$ to transform the patches to tokens:
\begin{gather}
    \mathrm{H} = \left[\texttt{Linear}(x_1), \texttt{Linear}(x_2),\cdots, \texttt{Linear}(x_n) \right],
\end{gather}
where $\mathrm{H}\in \mathbb{R}^{n \times d}$ are the embedded time series tokens. Patch-wise tokens preserve semantic information and reduce the number compared with point-based tokens, thus enhancing efficiency. 

\subsubsection{Local-Perception}
Although the patching technique~\citep{Triformer,nie2022time} has achieved some progress in time series foundation models, its fixed receptive field may limit models' generalization capabilities, thus requiring heavy backbones. The core reason is that the inherent periods and trends within data vary across datasets from distinct domains, causing the diversity of frequencies, but the fixed patch size in pretrained models only captures the same span without considering the changing surrounding environmental information.

Specifically, in pretraining scenarios, a patch with the fixed length may cover the whole period in one pretraining dataset but only 70\% of a period in another--see Figure~\ref{fig: overview}(b), which makes this semantic variation difficult to model. To tackle this limitation, we manage to recognize the varying periods, then align their information with patch-wise tokens to supplement the environmental semantics. First, we adopt the FFT~\citep{rfft} to find the period of the input series:
\begin{gather}
    \mathcal{A} = \texttt{Amp}(\texttt{FFT}(\mathrm{X})),\quad
    \mathcal{F} = \operatorname*{arg\,max}_{1\leq f\leq\lfloor T/2\rfloor}\mathcal{A}[f],\quad
    \mathrm{P} = \left\lceil \frac{T}{\mathcal{F}} \right\rceil,
\end{gather}
where $\texttt{FFT}(\cdot)$ and $\texttt{Amp}(\cdot)$ denote the FFT and amplitude calculation. The search excludes the zero-frequency (DC) component, and $\mathrm{P}$ is the estimated period length. Then we consider the relative positions of each patch and its surrounding period to align them:
\begin{gather}
    s = \lfloor (p + \mathrm{P} -1 ) / \mathrm{P} \rfloor \cdot \mathrm{P} - p,
   \mathrm{X}^\prime = \texttt{LeftPadding}(\mathrm{X}, s),\\
    \tilde{\mathrm{X}} = \texttt{Patching}(\mathrm{X}^\prime) = [w_1, w_2, \cdots, w_n],
\end{gather}
where $\texttt{Patching}$ uses a window of length $p+s$ and stride $p$, producing local environmental patches $\tilde{\mathrm{X}}\in\mathbb{R}^{n\times(p+s)}$. The padding length $s$ is used to make each local environmental patch $w_i \in \mathbb{R}^{p+s}$ cover the corresponding patch $x_i\in\mathbb{R}^p$, and $p+s$ is the integer multiple of the period $\mathrm{P}$. These operations can help align the intricate and varying local information $w_i$ to the patch $x_i$.

For each $w_i\in \mathbb{R}^{p+s}$, its length varies with the sample during pretraining due to the multiple frequencies of the time series corpus, which is difficult to encode using networks with fixed input dimensions, such as MLPs. To overcome this, we propose to utilize the LegT Legendre Memory--see Section~\ref{sec: legendre}, to encode the varying local environmental information, i.e., the varying periods, into vectors of the fixed size. For each $w_i\in \mathbb{R}^{p+s}$, we use the discretized form of LegT to compress it like a state space model in an autoregressive manner:
\begin{gather}
    \bar{A} = I - \Delta t A, \quad \bar{B}=\Delta t B,\\
    f_t = w_{i,t}, m_0 = 0, m_t = \bar{A}m_{t-1} + \bar{B}f_t, 
\end{gather}
where $\theta=p+s$ is determined by the length of $w_i$, $A\in\mathbb{R}^{d\times d}$ and $B\in\mathbb{R}^{d\times1}$ are fixed and initialized using Equation~(\ref{fu: init legT}), and $\Delta t$ is the predefined step size for forward Euler discretization.
This mechanism compresses each local environmental patch $w_i \in \mathbb{R}^{p+s}$ with \textit{varying lengths} into a fixed-size memory cell $m_t \in \mathbb{R}^d$. The generated environmental tokens are $\mathrm{H}^\prime = [w_1^\prime, w_2^\prime, \cdots, w_n^\prime]$, where each token $w_i^\prime \in \mathbb{R}^{d}$ corresponds to the last state of the memory cell. We then fuse the original tokens $\mathrm{H}$ and the environmental tokens $\mathrm{H}^\prime$ into $\tilde{\mathrm{H}} \in \mathbb{R}^{n \times d}$ via $\tilde{\mathrm{H}} = \mathrm{H} + \mathrm{H}^\prime $.
Theorem~\ref{theory: 1} motivates increasing $d$ to reduce the polynomial projection error; it does not by itself guarantee lossless reconstruction when $d>\theta$. We evaluate reconstruction empirically in Appendix~\ref{app: model analytics} and provide the projection proof in Appendix~\ref{app: theory}.

\subsubsection{MSA-Encoder}
After obtaining the fused tokens, we use the Multi-head Self-Attention Layer~\citep{vaswani2017attention} as the encoder to generate the hidden representations of them. Through fully-connected interaction, the MSA-Encoder aims at learning the similarities and differences among patterns and environmental information within the tokens
$ \hat{\mathrm{H}} = \texttt{MSA-Encoder}(\tilde{\mathrm{H}})$,
where the representations $\hat{\mathrm{H}} \in \mathbb{R}^{n \times d}$ are the outputs of the Encoding phase, and provide rich semantics within data.

\subsection{Decoding}

\subsubsection{SSD-Decoder} 
We utilize LegT in Local-Perception of the Encoding phase to enrich the local environmental information.
In the Decoding phase, the long-range modeling capability gains more attention, which calls for adaptive memory utilization when making long-term forecasts. The Structured State-Space Duality (SSD) layer in Mamba-2~\citep{mamba2}, uses the selective state space modeling mechanism to adaptively process historical information, significantly improving efficiency while achieving similar accuracy to Transformers. Its mechanism takes the form as follows:
\begin{gather}
    m_t = A_t m_{t-1} + B_t x_t, y_t = C^T_tm_t,
\end{gather}
where the form shares similarities with LegS--see Formula~(\ref{fu: ssd}). Previous works~\citep{gu2023mamba,mamba2} demonstrate that adapting $A,B,C$ to selective $A_t, B_t, C_t$ can enhance the non-linearity capability. Among them, $A_t$ is obtained from $A$ through the learnable granularity $\Delta t$ of discretization, so that the initialization of $A$ still matters. To preserve the capability of adaptive memory utilization and ensure the linear complexity of SSD, we use the diagonal of $A$ in LegS to initialize the continuous-time transition parameters before discretization.
We then use the SSD layer with initialized $A_t$ as the Decoder to forecast the future tokens:
\begin{gather}
    \mathrm{Q} = \texttt{SSD-Decoder}(\hat{\mathrm{H}}),
\end{gather}
where $\mathrm{Q}\in\mathbb{R}^{l\times d}$. Supposing the length of forecasting horizon is $L$, $l=\lceil L/p \rceil$. Based on the LegS, SSD-Decoder can efficiently summarize historical information and adaptively use it for forecasting.

\subsubsection{MCA-Decoder}
We then use the Multi-head Cross-Attention~\citep{vaswani2017attention} Layer to further enhance the outputs through recalling the similar representations in the inputs. Specifically, we set $\mathrm{Q}$ as Query, $\hat{\mathrm{H}}$ as Key and Value. 
Through the skip connections in MCA-Decoder, the feature sequence $\mathrm{Q}$ is refined as $\tilde{\mathrm{Q}} = \texttt{MCA-Decoder}(\mathrm{Q},\hat{\mathrm{H}})$, where $\tilde{\mathrm{Q}} \in \mathbb{R}^{l \times d}$ denotes the output of the Decoding phase and captures the high-order semantic information learned from the token transition process in the hidden space.


\subsection{Flow-based Head}
Through pretraining on Time Series Corpus from multiple domains, the Encoding and Decoding phases of FLAME possess strong capability of modeling the token transition principles, which means $\tilde{\mathrm{Q}} = [q_1, q_2, \cdots, q_l] \in \mathbb{R}^{l\times d}$ are high-quality representations for forecasting. In time series foundation models, making token-wise forecasts through relatively lightweight heads can preserve efficiency and ensure the heavy backbones do learn the ``knowledge''~\citep{timer,liu2025sundial}.

However, due to the inherent temporal causality within time series data, patch-based forecasting models only model the token-wise causal relationships in decoding. When transforming these high-dimensional tokens into patch sizes for forecasts, the fine-grained causal relationships within each patch are often neglected. To support generative probabilistic forecasting, remain lightweight, and capture such fine-grained causal relationships, we adopt a Masked Autoregressive Flow (MAF)~\citep{maf} to model the intricate probabilistic distributions of real data. The whole generative process is formulated as:
\begin{gather}
\tilde{\mathrm{Q}} = [q_1, q_2, \cdots, q_l] \in \mathbb{R}^{l\times d},~
   \epsilon_i \sim \mathcal{N}(0,I), \\
   \hat{\mathrm{Y}}_i = T_1^{-1} \circ \cdots \circ T_K^{-1}(\epsilon_i \mid q_i),
\end{gather}
where $q_i\in\mathbb{R}^{d}$ is the conditioning context for the $i$-th predicted patch $\hat{\mathrm{Y}}_i\in\mathbb{R}^{p}$, and $\epsilon_i\in\mathbb{R}^{p}$ is sampled from the base distribution. Each $T_j$ maps data toward latent space; generation applies the inverse layers in reverse order. Define $\mathrm{Z}_0^i=\mathrm{Y}_i$ and $\mathrm{Z}_K^i=\epsilon_i$ in the data-to-latent direction. The $j$-th layer is an autoregressive affine transformation parameterized by a masked MLP:
\begin{gather}
\begin{aligned}
(\alpha_{i,k}^{(j)},\; \beta_{i,k}^{(j)})
    &= \texttt{MLP}^j_{\mathcal{M},k}(\mathrm{Z}_{j-1}^i[1{:}k{-}1],\; q_i), \\
\mathrm{Z}_j^i[k]
    &= \exp(\alpha_{i,k}^{(j)})\,\mathrm{Z}_{j-1}^i[k]+\beta_{i,k}^{(j)},
\end{aligned}
\label{eq:maf_affine}
\end{gather}
where $\texttt{MLP}^j_{\mathcal{M},k}$ denotes the pair of outputs for dimension $k$ of the shared masked MLP. It generates element-wise affine parameters $(\alpha_{i,k}^{(j)},\beta_{i,k}^{(j)})$. The effective input-to-parameter dependency mask is:
\begin{gather}
\mathcal{M}_{r,k} = \begin{cases}
1 & \text{if } k<r, \\
0 & \text{if } k\geq r,
\end{cases}\qquad r,k\in\{1,\ldots,p\},
\end{gather}
where $\mathcal{M}_{r,k}=0$ means that the affine parameters for dimension $r$ cannot depend on input dimension $k$. The transformed value still directly depends on its own input through the affine term. Thus, $(\alpha_{i,k}^{(j)},\beta_{i,k}^{(j)})$ depend only on preceding input dimensions and the full conditioning token $q_i$. This preserves the autoregressive ordering within each patch (Figure~\ref{fig: overview}(f)) and ensures \textbf{exact invertibility}: $\mathrm{Z}_{j-1}^i[k]=(\mathrm{Z}_j^i[k]-\beta_{i,k}^{(j)})/\exp(\alpha_{i,k}^{(j)})$, evaluated sequentially for $k=1,\ldots,p$ within each inverse layer. Applying all $K$ inverse layers gives the predicted patches, which are concatenated as:
\begin{gather}
    \hat{\mathrm{Y}} = [\hat{\mathrm{Y}}_1,\hat{\mathrm{Y}}_2,\cdots, \hat{\mathrm{Y}}_l ],
\end{gather}
where $\hat{\mathrm{Y}} \in \mathbb{R}^{l \times p}$ can be further flattened and truncated to the first $L$ points to obtain forecasts.

With the affine autoregressive layers, the Flow-based Head achieves intricate distributional transformation while remaining lightweight. Due to the probabilistic generative mechanism, multiple samples are recommended to evaluate the forecasted distributions. Let $\mathrm{Y}_i\in\mathbb{R}^{p}$ denote the $i$-th ground-truth target patch. For training, we apply the encoding direction $T = T_K \circ \cdots \circ T_1$ to map data to latent space. Since the input $\mathrm{Y}_i$ is known, the masked MLP computes all affine parameters in parallel. The log-likelihood is:
\begin{equation}
\begin{gathered}
    \epsilon_i=T_K\circ\cdots\circ T_1(\mathrm{Y}_i\mid q_i),
    \log p_\theta(\mathrm{Y}_i\mid q_i)=\log p_\epsilon(\epsilon_i)
      +\sum_{j=1}^K\sum_{k=1}^p\alpha_{i,k}^{(j)},\\
    \mathcal{L}_{\text{NLL}}=-\frac{1}{l}\sum_{i=1}^l\log p_\theta(\mathrm{Y}_i\mid q_i).
\end{gathered}
\label{eq:flow_nll}
\end{equation}
Each autoregressive layer has a triangular Jacobian with diagonal entries $\exp(\alpha_{i,k}^{(j)})$, making the double sum the log absolute determinant of the data-to-latent mapping. Appendix~\ref{app:maf_properties} proves global invertibility, this exact likelihood formula, and normalization of the conditional density.

%% file: Sections/Experiments.tex
\section{Experiments}
\begin{wrapfigure}{r}{0.45\textwidth}
\vspace{-5mm}
    \centering
    \includegraphics[width=1\linewidth]{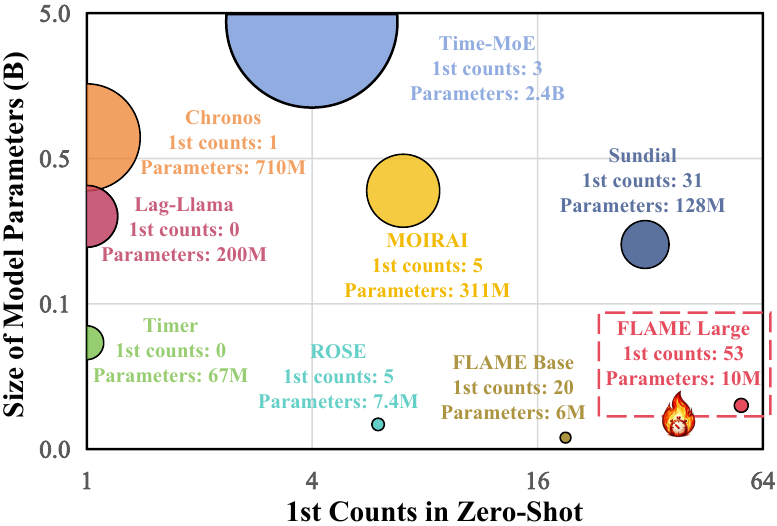}
    \caption{The parameter scale and zero-shot performance of Time Series Foundation models. FLAME Large achieves the most first-place results with approximately 10M parameters.}
\label{fig: efficiency}
\vspace{-6mm}
\end{wrapfigure}
Detailed experimental settings are provided in Appendix~\ref{app: exp}. We evaluate deterministic forecasting on TSFM-Bench and TFB in Section~\ref{sec: point forecast}, and probabilistic forecasting on ProbTS in Section~\ref{sec: prob forecast}. We also conduct a detailed analysis of key components of FLAME to reveal the source of its generalization capability in Section~\ref{sec: model analysis}. To analyze the scalability of FLAME, we make studies on training and inference in Section~\ref{sec: scalability}. As an important property for deployment, we also analyze the efficiency of FLAME in Section~\ref{sec: efficiency analysis}.
In summary--see Figure~\ref{fig: efficiency}, FLAME achieves the most first-place results with a small parameter count. \textit{We provide detailed results in Appendix~\ref{app: full results}, including full results on TSFM-Bench and ProbTS, as well as results under full-shot and few-shot settings.}

\subsection{Deterministic Forecasting}
\label{sec: point forecast}
As shown in Table~\ref{tab: main point forecasting avg}, the FLAME family outperforms other advanced time series foundation models on datasets from TSFM-Bench. Compared with the recent state-of-the-art model Sundial, FLAME Large outperforms it in most settings, achieving the most first-place results in zero-shot deterministic forecasting tasks. Note that even the FLAME Small (2M) can achieve competitive performance with Sundial Base (128M) while only possessing the \textit{1/64} parameter scale. Compared with the heaviest model Time-MoE Ultra (2.4B), FLAME Small possesses the \textit{1/1,200} parameter scale but achieves average MSE reduction of \textit{15.0\%}. Considering lightweight baselines like UniTS and ROSE, FLAME Small also shows excellent advantages in generalization capabilities, achieving significant average MSE reductions of \textit{26.3\%} and \textit{17.4\%}.
\begin{table*}[!htbp]
\caption{Average results of zero-shot deterministic forecasting.
(``-'') denotes datasets included in the model’s pretraining and therefore excluded from testing. \textcolor{red}{\textbf{Red}}: the best, \textcolor{blue}{\underline{Blue}}: the 2nd best. Full results are listed in Table~\ref{tab: main point forecasting all}. We also provide strict comparisons in Table~\ref{tab: strict comparison point}, excluding datasets used in some baselines' pretraining.}
\label{tab: main point forecasting avg}
\resizebox{1\linewidth}{!}{

    \begin{tabular}{c|cc|cc|cc|cc|cc|cc|cc|cc|cc|cc|cc}
    \toprule
        \multirow{2}{*}{Models} & \multicolumn{2}{c|}{\textbf{FLAME}} & \multicolumn{2}{c|}{\textbf{FLAME}} & \multicolumn{2}{c|}{\textbf{FLAME}} & \multicolumn{2}{c|}{\textbf{Sundial}} & \multicolumn{2}{c|}{\textbf{LightGTS}} & \multicolumn{2}{c|}{\textbf{ROSE}} & \multicolumn{2}{c|}{\textbf{Timer}} & \multicolumn{2}{c|}{\textbf{Chronos}} & \multicolumn{2}{c|}{\textbf{Time-MoE}} & \multicolumn{2}{c|}{\textbf{TinyTTM}} & \multicolumn{2}{c}{\textbf{MOIRAI}} \\    
        ~ & \multicolumn{2}{c|}{\textbf{Small}} & \multicolumn{2}{c|}{\textbf{Base}} &\multicolumn{2}{c|}{\textbf{Large}} & \multicolumn{2}{c|}{(2025)} & \multicolumn{2}{c|}{(2025)} & \multicolumn{2}{c|}{(2025)} & \multicolumn{2}{c|}{(2024)} & \multicolumn{2}{c|}{(2024)} & \multicolumn{2}{c|}{(2024)} & \multicolumn{2}{c|}{(2024)} & \multicolumn{2}{c}{(2024)}\\  \cmidrule{1-23}
        Metrics & MSE & MAE & MSE & MAE & MSE & MAE & MSE & MAE & MSE & MAE & MSE & MAE & MSE & MAE & MSE & MAE & MSE & MAE & MSE & MAE & MSE & MAE \\  \cmidrule{1-23}
        ETT (Avg) & 0.339 & 0.382 & \textcolor{blue}{\underline{0.339}} & 0.381 & 0.344 & 0.385 & \textcolor{red}{\textbf{0.335}} & \textcolor{blue}{\underline{0.379}} & 0.339 & \textcolor{red}{\textbf{0.379}} & 0.393 & 0.411 & 0.551 & 0.478 & 0.442 & 0.408 & 0.357 & 0.390 & 0.441 & 0.430 & 0.382 & 0.388 \\ \cmidrule{1-23}
        Weather & \textcolor{red}{\textbf{0.215}} & \textcolor{red}{\textbf{0.261}} & 0.222 & 0.269 & 0.227 & 0.273 & 0.234 & 0.270 & \textcolor{blue}{\underline{0.219}} & \textcolor{blue}{\underline{0.267}} & 0.265 & 0.305 & 0.292 & 0.313 & 0.288 & 0.309 & 0.256 & 0.289 & 0.265 & 0.307 & 0.260 & 0.275 \\ \cmidrule{1-23}
        Electricity & 0.287 & 0.371 & 0.222 & 0.319 & 0.217 & 0.313 & \textcolor{red}{\textbf{0.169}} & \textcolor{red}{\textbf{0.265}} & 0.233 & 0.319 & 0.234 & 0.320 & 0.297 & 0.375 & - & - & - & - & 0.222 & 0.317 & \textcolor{blue}{\underline{0.188}} & \textcolor{blue}{\underline{0.273}} \\ \cmidrule{1-23}
        Traffic & 0.673 & 0.461 & 0.576 & 0.396 & \textcolor{red}{\textbf{0.546}} & \textcolor{red}{\textbf{0.384}} & - & - & 0.610 & 0.399 & 0.588 & 0.412 & 0.613 & 0.407 & 0.615 & 0.421 & - & - & \textcolor{blue}{\underline{0.564}} & \textcolor{blue}{\underline{0.386}} & - & - \\ \cmidrule{1-23}
        Solar & 0.203 & 0.278 & \textcolor{blue}{\underline{0.184}} & 0.277 & \textcolor{red}{\textbf{0.168}} & \textcolor{blue}{\underline{0.272}} & 0.221 & \textcolor{red}{\textbf{0.252}} & 0.219 & 0.305 & 0.505 & 0.549 & 0.771 & 0.604 & 0.393 & 0.319 & 0.411 & 0.428 & 0.815 & 0.710 & 0.714 & 0.704 \\ \cmidrule{1-23}
        PEMS08 & 0.324 & 0.384 & \textcolor{blue}{\underline{0.312}} & \textcolor{blue}{\underline{0.374}} & \textcolor{red}{\textbf{0.293}} & \textcolor{red}{\textbf{0.351}} & - & - & 0.812 & 0.692 & 1.369 & 0.979 & 0.866 & 0.695 & 1.707 & 1.024 & - & - & 1.730 & 1.066 & - & - \\ \cmidrule{1-23}
        Wind & 1.144 & 0.765 & \textcolor{blue}{\underline{1.120}} & \textcolor{blue}{\underline{0.758}} & \textcolor{red}{\textbf{1.042}} & \textcolor{red}{\textbf{0.733}} & 1.186 & 0.772 & 1.292 & 0.836 & 1.251 & 0.820 & 1.201 & 0.783 & 1.478 & 0.834 & - & - & 1.337 & 0.829 & 1.299 & 0.795 \\ \cmidrule{1-23}
        NYSE & 0.527 & 0.553 & \textcolor{red}{\textbf{0.442}} & \textcolor{red}{\textbf{0.430}} & 0.529 & 0.507 & 0.880 & 0.642 & 0.620 & 0.626 & - & - & 0.988 & 0.704 & 1.129 & 0.720 & - & - & - & - & \textcolor{blue}{\underline{0.492}} & \textcolor{blue}{\underline{0.441}} \\ \cmidrule{1-23}
        \rowc
        \textbf{$1^{st}$ Count} & 8 & 8 & 6 & 4 & \textcolor{red}{\textbf{15}} & \textcolor{red}{\textbf{9}} & \textcolor{blue}{\underline{9}} & \textcolor{blue}{\underline{9}} & 0 & 3 & 3 & 2 & 0 & 0 & 0 & 0 & 1 & 2 & 2 & 3 & 0 & 4  \\ 
        \rowc
         \textbf{$2^{nd}$ Count} & 3 & 1 & \textcolor{red}{\textbf{15}} & \textcolor{blue}{\underline{9}} & 2 & \textcolor{red}{\textbf{10}} & 2 & 7 & \textcolor{blue}{\underline{9}} & 6 & 2 & 3 & 0 & 0 & 0 & 0 & 2 & 1 & 5 & 3 & 4 & 4 \\ 
        \bottomrule
    \end{tabular}}
\end{table*}

\begin{figure}[!htbp]
    \centering
    \begin{subfigure}[t]{0.49\linewidth}
        \centering
        \includegraphics[width=\linewidth]{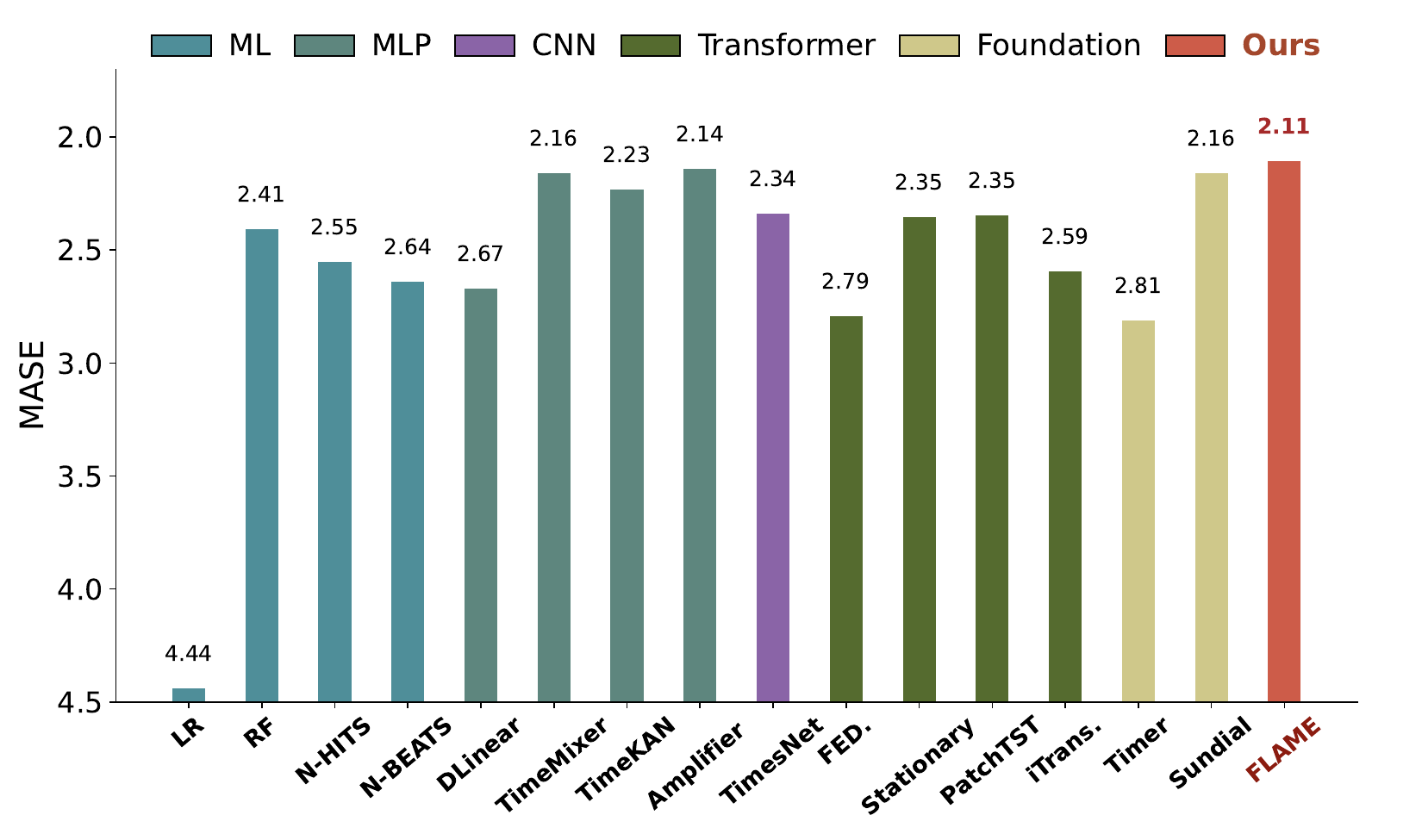}
        \caption{MASE.}
        \label{fig: tfb univariate mase}
    \end{subfigure}
    \hfill
    \begin{subfigure}[t]{0.49\linewidth}
        \centering
        \includegraphics[width=\linewidth]{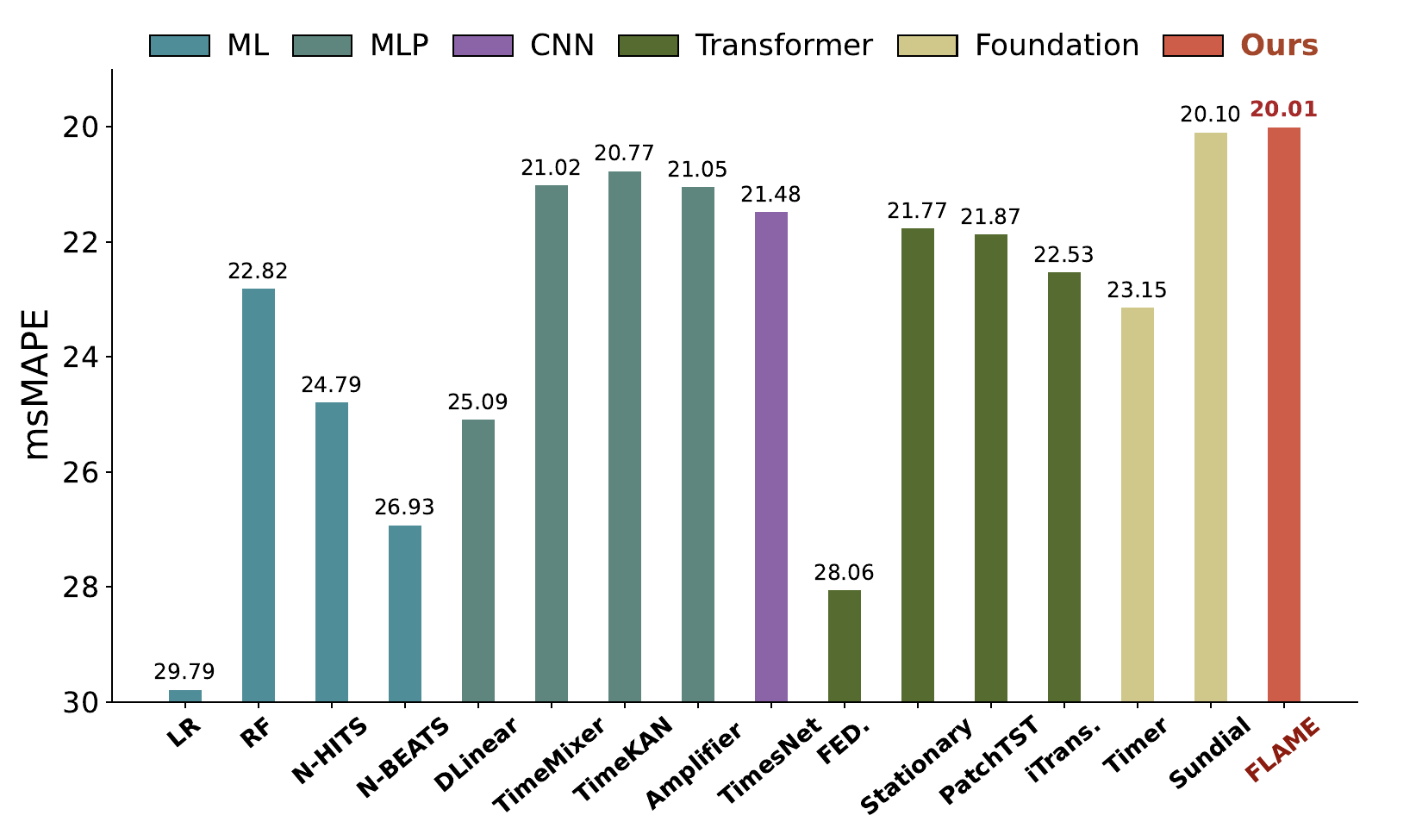}
        \caption{msMAPE.}
        \label{fig: tfb univariate msmape}
    \end{subfigure}
    \caption{Short-term forecasting results averaged over 8,068 univariate time series from TFB. Lower values indicate better performance.}
    \label{fig: tfb short term}
\end{figure}

\begin{samepage}
\paragraph{Short-Term Forecasting.}
We further evaluate FLAME Large on short-term forecasting using 8,068 univariate time series from TFB~\citep{qiu2024tfb}. Figure~\ref{fig: tfb short term} compares FLAME with machine learning methods, end-to-end supervised models, and time series foundation models. FLAME Large achieves the lowest average MASE of \textbf{2.11} and msMAPE of \textbf{20.01} among the compared methods. For comparison, Sundial obtains 2.16 MASE and 20.10 msMAPE.

The baseline rankings differ across the two metrics. For example, Amplifier achieves a lower MASE than TimeKAN (2.14 versus 2.23), whereas TimeKAN obtains a lower msMAPE (20.77 versus 21.05). FLAME ranks first under both criteria, showing that its aggregate advantage holds across the two error measures. These results complement the long-horizon multivariate evaluations on TSFM-Bench, demonstrating that FLAME's forecasting capability also extends to short-term univariate settings.
\par
\end{samepage}

\subsection{Probabilistic Forecasting}
\label{sec: prob forecast}
Table~\ref{tab: main prob forecasting avg} summarizes the results of probabilistic forecasting on ProbTS. FLAME shows strong performance against the other probabilistic foundation models. FLAME Large achieves the most first-place results in zero-shot probabilistic forecasting tasks, excelling in both periodic scenarios like ETT and Weather, and non-stationary scenarios like Exchange and ILI. Compared with the well-established baselines MOIRAI Large and Chronos Large, FLAME Large achieves average CRPS reductions of \textit{34.3\%} and \textit{18.1\%}, respectively. On Weather, FLAME Large reduces CRPS from Sundial's 0.087 to 0.074 and NMAE from 0.102 to 0.090. These paired reductions show improvements in both distributional and point forecast accuracy.

FLAME's zero-shot performance also compares favorably with several full-shot baselines. Compared with diffusion-based CSDI and Flow-based GRU NVP, which need dozens of hours to train from scratch on a large dataset, FLAME Large achieves average \textit{23.5\%} and \textit{35.1\%} CRPS reduction. This demonstrates the effectiveness of Flow-based Head in FLAME, which provides both efficiency and accuracy.
\begin{table*}[!htbp]
\caption{Average results of zero-shot probabilistic forecasting.
(``-'') denotes datasets included in the model’s pretraining and therefore excluded from testing. (``/'') denotes the excessive time consumption. \textcolor{red}{\textbf{Red}}: the best, \textcolor{blue}{\underline{Blue}}: the 2nd best. Full results are in Table~\ref{tab: main prob forecasting all}. We also provide strict comparisons in Table~\ref{tab: strict comparison prob}, excluding datasets used in some baselines' pretraining. }
\label{tab: main prob forecasting avg}
\resizebox{1\linewidth}{!}{
    \begin{tabular}{c|cc|cc|cc|cc|cc|cc|cc|cc|cc|cc}
    \toprule
        \multirow{2}{*}{Models} & \multicolumn{2}{c|}{\textbf{FLAME}} & \multicolumn{2}{c|}{\textbf{FLAME}} & \multicolumn{2}{c|}{\textbf{FLAME}} & \multicolumn{2}{c|}{\textbf{Sundial}} & \multicolumn{2}{c|}{\textbf{Chronos}} & \multicolumn{2}{c|}{\textbf{MOIRAI}} & \multicolumn{2}{c|}{\textbf{Lag-Llama}} & \multicolumn{2}{c|}{\textbf{CSDI}} & \multicolumn{2}{c|}{\textbf{TimeGrad}} & \multicolumn{2}{c}{\textbf{GRU NVP}} \\    
        ~ & \multicolumn{2}{c|}{\textbf{Small}} & \multicolumn{2}{c|}{\textbf{Base}} &\multicolumn{2}{c|}{\textbf{Large}} & \multicolumn{2}{c|}{(2025)} & \multicolumn{2}{c|}{(2024)} & \multicolumn{2}{c|}{(2024)} & \multicolumn{2}{c|}{(2023)} & \multicolumn{2}{c|}{(2022)} & \multicolumn{2}{c|}{(2022)} & \multicolumn{2}{c}{(2021)}\\  \cmidrule{1-21}
        Metrics & CRPS & NMAE & CRPS & NMAE & CRPS & NMAE & CRPS & NMAE & CRPS & NMAE & CRPS & NMAE & CRPS & NMAE & CRPS & NMAE & CRPS & NMAE & CRPS & NMAE \\  \cmidrule{1-21}
        ETT (Avg) &\textcolor{blue}{\underline{0.235}} &\textcolor{blue}{\underline{0.293}} &0.250 &0.317 &0.236 &0.295 &\textcolor{red}{\textbf{0.231}} &\textcolor{red}{\textbf{0.273}} &0.290 &0.316 &0.366 &0.377 &0.273 &0.310 &0.304 &0.389 &0.493 &0.619 &0.476 &0.605 \\ \cmidrule{1-21}
        Weather &0.089 &0.114 &\textcolor{blue}{\underline{0.087}} &0.105 &\textcolor{red}{\textbf{0.074}} &\textcolor{red}{\textbf{0.090}} &0.087 &\textcolor{blue}{\underline{0.102}} &0.142 &0.158 &0.179 &0.143 &0.096 &0.106 &0.077 &0.093 &0.125 &0.155 &0.119 &0.147 \\ \cmidrule{1-21}        
        Electricity &0.093 &0.118 &0.086 &0.110 &\textcolor{red}{\textbf{0.081}} &\textcolor{red}{\textbf{0.097}} &\textcolor{blue}{\underline{0.081}} &\textcolor{blue}{\underline{0.098}} & - & - &0.247 &0.290 & - & - & / & / &0.102 &0.126 &0.101 &0.127 \\ \cmidrule{1-21}        
        Traffic &0.268 &0.337 &\textcolor{blue}{\underline{0.248}} &\textcolor{red}{\textbf{0.306}} &\textcolor{red}{\textbf{0.247}} &\textcolor{blue}{\underline{0.308}} & - & - &0.269 &0.295 & - & - &0.330 &0.385 & / & / &0.211 &0.246 &0.198 &0.245 \\ \cmidrule{1-21}        
        Exchange &0.043 &0.048 &\textcolor{red}{\textbf{0.040}} &\textcolor{red}{\textbf{0.047}} &\textcolor{blue}{\underline{0.042}} &\textcolor{blue}{\underline{0.047}} &0.045 &0.049 &0.044 &0.047 &0.045 &0.050 &0.057 &0.069 &0.069 &0.086 &0.082 &0.095 &0.073 &0.093 \\ \cmidrule{1-21}
        ILI &0.162 &0.198 &\textcolor{red}{\textbf{0.130}} &\textcolor{red}{\textbf{0.166}} &0.154 &0.188 &\textcolor{blue}{\underline{0.148}} &\textcolor{blue}{\underline{0.166}} &0.170 &0.197 &0.159 &0.197 &0.156 &0.211 &0.276 &0.290 &0.284 &0.310 &0.283 &0.309 \\ \cmidrule{1-21}
        \rowc
        \textbf{$1^{st}$ Count} &3 &3 &\textcolor{blue}{\underline{7}} &3 &\textcolor{red}{\textbf{15}} &\textcolor{red}{\textbf{14}} &5 &\textcolor{blue}{\underline{8}} &0 &1 &0 &1 &0 &0 &2 &2 &0 &2 &4 &2 \\   
        \rowc
        \textbf{$2^{nd}$ Count}& \textcolor{blue}{\underline{7}} & 5 & 5 & 5 & \textcolor{red}{\textbf{7}} & \textcolor{blue}{\underline{6}} & 5 & \textcolor{red}{\textbf{8}} & 0 & 2 & 1 & 1 & 1 & 0 & 6 & 5 & 4 & 2 & 0 & 2\\ 
        \bottomrule
    \end{tabular}
    }
\end{table*}

\begin{samepage}
\subsection{Model Analysis}
\label{sec: model analysis}

\begin{wraptable}{r}{0.6\textwidth}
\centering
 \vspace{-4mm}
\caption{Ablation studies on deterministic and probabilistic forecasting.}
\label{tab: ablations}
\resizebox{1\linewidth}{!}{
    \begin{tabular}{c|cc|cc|cc|cc}
    \toprule
        Variants & \multicolumn{2}{c|}{ETT (Avg)} & \multicolumn{2}{c|}{Weather} & \multicolumn{2}{c|}{Electricity} & \multicolumn{2}{c}{Traffic} \\     \cmidrule{1-9} 
        Metrics & MSE & CRPS &MSE & CRPS & MSE & CRPS & MSE & CRPS \\  \cmidrule{1-9} 
        w/o Local-Perception &0.374 &0.266 &0.278 &0.104 &0.293 &0.109 &0.690 &0.283 \\
        w/o Period Alignment &0.368 &0.251 &0.248 &0.083 &0.275 &0.096 &0.627 &0.264 \\
        Learnable LegT &\underline{0.348} &\underline{0.245} &\underline{0.237} &\underline{0.074} &\underline{0.222} &\underline{0.084} &\underline{0.566} &\underline{0.253} \\  \cmidrule{1-9}
        
        Self-Attention Decoder &0.391 &0.300 &\underline{0.253} &0.106 &0.300 &0.131 &0.654 &0.341 \\
        Causal-Attention Decoder &0.369 &0.260 &0.255 &0.103 &0.230 &0.095 &0.560 &\underline{0.266} \\
        Mamba v1 Decoder &\underline{0.362} &\underline{0.254} &0.256 &\underline{0.098} &\underline{0.218} &\underline{0.087} &\underline{0.557} &0.277 \\\cmidrule{1-9}
        VAE Head &0.366 &0.262 &0.264 &0.093 &0.291 &0.138 &0.745 &0.506 \\
        Diffusion Head &0.360 &0.252 &\underline{0.230} &\underline{0.077} &0.235 &0.088 &\underline{0.568} &\underline{0.266} \\
        Flow Match Head &\underline{0.358} &\underline{0.251} &0.233 &0.080 &\underline{0.232} &\underline{0.088} &0.595 &0.297 \\\cmidrule{1-9}

        FLAME Large &\textbf{0.344} &\textbf{0.236} &\textbf{0.227} &\textbf{0.074} &\textbf{0.217} &\textbf{0.081} &\textbf{0.546} &\textbf{0.247} \\
        \bottomrule
    \end{tabular}}
    \vspace{-3mm}
\end{wraptable}

To identify the source of FLAME's generalization capability, we analyze the key components of FLAME including Local-Perception, SSD-Decoder, and Flow-based Head in Table~\ref{tab: ablations}. We choose ETT (Avg), Weather, Electricity and Traffic (shared in TSFM-Bench and ProbTS), and report average MSE and CRPS to evaluate deterministic and probabilistic forecasting.
The results support the effectiveness of the proposed components--see Appendix~\ref{app: model analytics} for full results, analyses, and implementation of variants.
\par
\end{samepage}

\subsection{Scalability Analysis}
\label{sec: scalability}

\begin{figure}[!htb]
    \centering
    \begin{subfigure}[t]{0.41\linewidth}
        \centering
        \includegraphics[width=\linewidth]{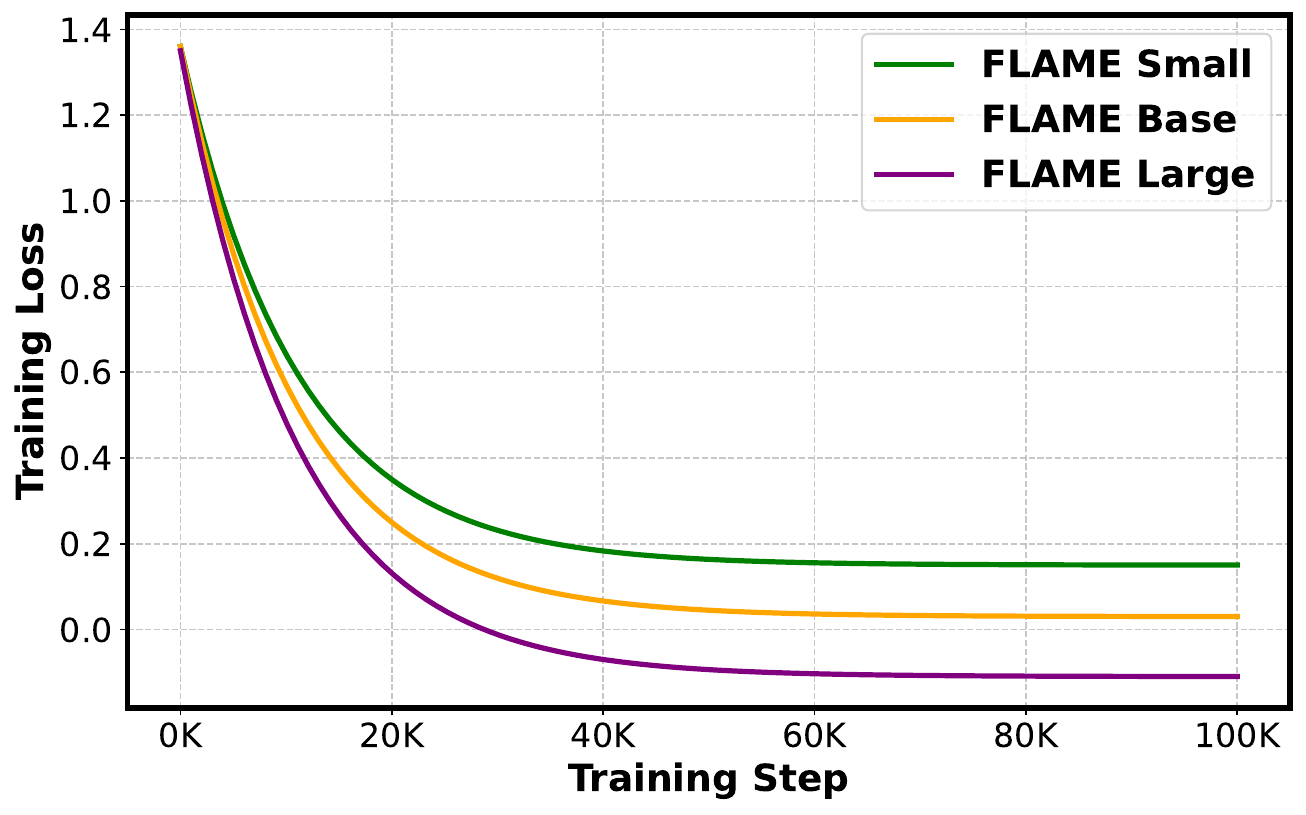}
        \caption{The training curves of the FLAME family.}
        \label{fig: train curve}
    \end{subfigure}
    \hfill
    \begin{subfigure}[t]{0.58\linewidth}
        \centering
        \includegraphics[width=\linewidth]{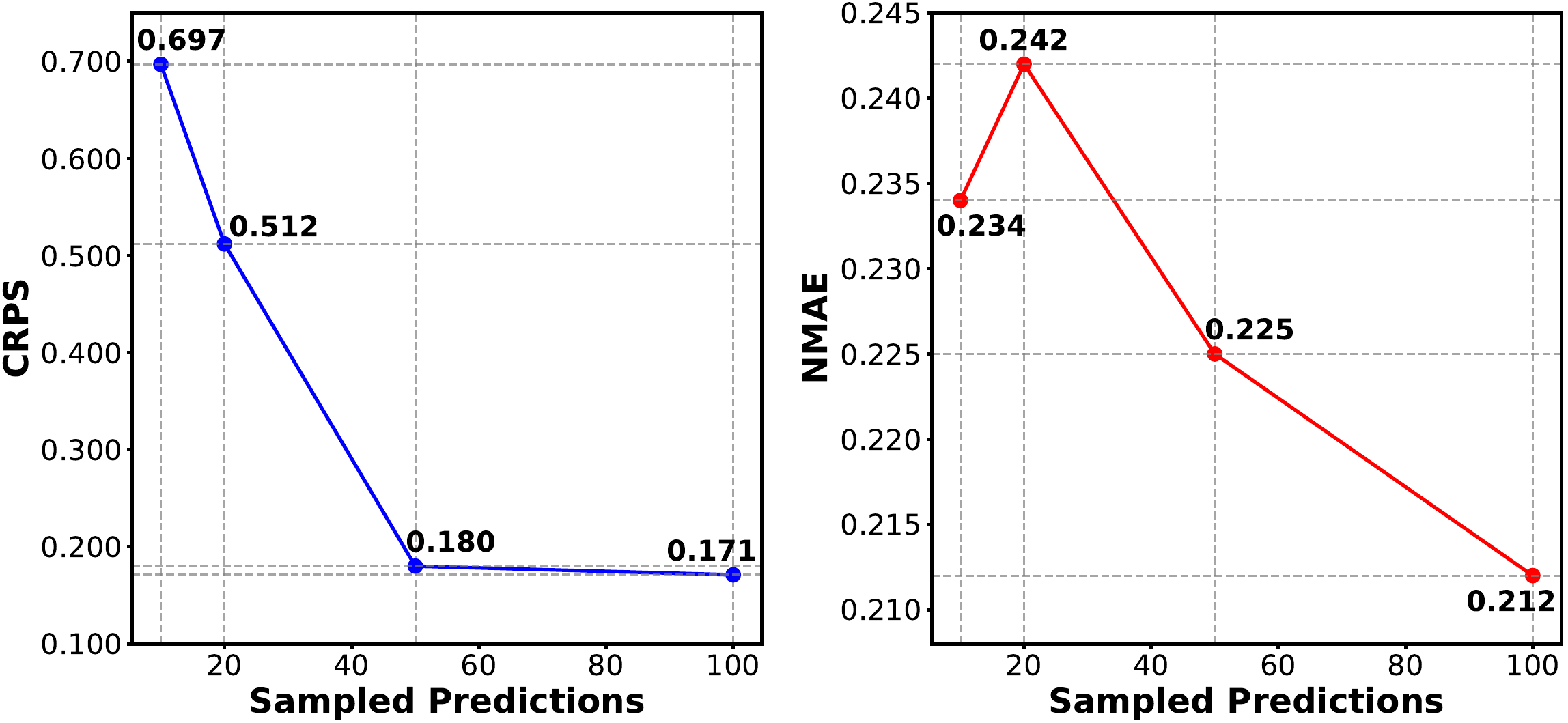}
        \caption{ Analysis on sampling number. Full results are in Table~\ref{tab: sampling num}.}
        \label{fig: sampled crps nmae}
    \end{subfigure}
    \caption{Scalability analysis of the FLAME family.}
    \label{fig: scalability}
\end{figure}

Though the FLAME family possesses extremely lightweight backbones while achieving state-of-the-art performance, we reveal that it also possesses the potential of scalability. 
As shown in Figure~\ref{fig: train curve}, the training curves of the FLAME family indicate that the model capacity increases with the parameter scale. 
According to the results in Table~\ref{tab: main point forecasting avg}--\ref{tab: main prob forecasting avg}, the zero-shot performance also gets better and the FLAME Large achieves the most first-place results in both deterministic and probabilistic forecasting tasks.

Since FLAME models distributions over the forecasting horizon, we also study the scalability of generative forecasting by exploring the correlations between the sampling number and performance--see Figure~\ref{fig: sampled crps nmae}. The results show that CRPS and NMAE generally improve as the number of samples increases, with 100 samples providing the best overall performance among the tested settings.


\noindent\begin{minipage}{\linewidth}
\subsection{Efficiency Analysis}
\begin{wraptable}{r}{0.55\textwidth}
\vspace{-4mm}
\centering
\caption{
Efficiency analysis of FLAME and other baselines on ETTm1 dataset, evaluated with the horizon of 720 and batch size of 1. We report the Parameter scale, MACs, Max GPU Memory, and Inference Time.
}
\label{tab: efficiency}
\resizebox{1\linewidth}{!}{
    \begin{tabular}{c|cccc}
    \toprule
        Models & Parameters & MACs & GPU (MB) & Inference (s) \\     \cmidrule{1-5} 
        Timer & 67.4 M & 52.6 G & 1,435 & 0.08 \\ 
        TimesFM & 200 M & 624.9 G & 1,395 & 0.16 \\ 
        Chronos Large & 700 M & 92327.9 G & 10,269 & 34.33 \\ 
        MOIRAI Large & 300 M & 97.36 G & 2,009 & 0.1 \\
        Time-MoE Ultra& 453 M & 5252.9 G & 14,131 & 2.13 \\ 
        Sundial Base &128 M & 432.3 G &603 &0.1 \\ \cmidrule{1-5} 
        TimeGrad & 12.3 M &19.3 G &522 &28.61 \\
        CSDI &15.9 M &37.5 G &817 & 86.18 \\ \cmidrule{1-5} 
        FLAME Large &\textbf{10.6 M} &\textbf{0.238 G} &\textbf{129} &\textbf{0.037} \\
        \bottomrule
    \end{tabular}}
    \vspace{-3mm}
\end{wraptable}
As a characteristic property of FLAME, the extreme efficiency owes to the lightweight backbone and Flow-based Head. For detailed quantification, we report the Parameter scale, Multiply-Accumulate Operations (MACs), Max GPU Memory Occupation, and Inference Time in Table~\ref{tab: efficiency}. We compare the FLAME Large with time series foundation models and end-to-end supervised probabilistic models.

Results show FLAME Large achieves consistent efficiency advantages over other time series foundation models. Compared with end-to-end supervised models like TimeGrad and CSDI, FLAME Large also has fewer parameters, faster inference speed, and fewer MACs and lower GPU memory usage.

\label{sec: efficiency analysis}
\par
\end{minipage}

%% file: Sections/Related-Works.tex
\section{Related Work}
\subsection{Time Series Forecasting}
Deterministic Time Series Forecasting is central to decision intelligence and has driven many deep learning methods, such as PatchTST~\citep{nie2022time}, DUET~\citep{qiu2025duet}, and SEER~\citep{seer}. LMU~\citep{voelker2019legendre} and FiLM~\citep{zhou2022film} introduced Legendre Memory (LegT) into time series forecasting, but they compress the entire input series into state vectors, which may cause information loss. In contrast, FLAME leverages LegT to compress varying token-wise environmental information and align it with the original tokens, enabling adaptation to multiple frequencies in the pretraining corpus and yielding strong generalization. Another important direction is Probabilistic Time Series Forecasting, which quantifies uncertainty and is more practical in real-world applications. Recently, generative mechanisms such as VAE, Diffusion, and Flow have been widely studied. Among them, $D^3$VAE~\citep{li2022generative} and $K^2$VAE~\citep{wu2025k2vae} are the most advanced VAE models in probabilistic time series forecasting. Considering multi-step generative mechanisms like Diffusion and Normalizing Flow, recent models include GRU NVP~\citep{normalizing-flow}, CSDI~\citep{CSDI}, TimeGrad~\citep{TimeGrad}, and TSDiff~\citep{TSDiff}, showing strong performance.


\subsection{Time Series Foundation Models}
Recent studies on Time Series Foundation Models have explored the source of generalization capability. Many works design specialized Transformer-based architectures and scale them to massive time series corpora to capture the inherent inductive bias in data. Representative deterministic forecasting models, such as TimesFM~\citep{timesfm}, Timer~\citep{timer}, Time-MoE~\citep{timemoe}, and VisionTS~\citep{chen2024visionts,DBLP:conf/icml/ChenS0WSL25} achieve strong zero-shot performance but rely on heavy backbones. In contrast, UniTS~\citep{units}, TinyTTM~\citep{tinyttm}, and ROSE~\citep{wang2025rose} show that lightweight backbones can also achieve state-of-the-art deterministic forecasting performance when inductive bias is captured more efficiently. Some foundation models also support probabilistic forecasting, including Lag-Llama~\citep{lagllama}, Chronos~\citep{ansarichronos}, Moirai~\citep{woo2024moirai}, and Sundial~\citep{liu2025sundial}, which outperform several end-to-end supervised probabilistic models. However, they still rely on heavy backbones, making probabilistic forecasting on large downstream datasets inefficient. In this paper, we propose \textbf{FLAME} to address this bottleneck. 

%% file: Sections/Conclusion.tex
\section{Conclusion}
In this work, we propose FLAME, a family of lightweight yet capable time series foundation models. FLAME adapts Legendre Memory in the backbone to encode local environmental information and support adaptive memory utilization for long-term forecasting, yielding strong generalization capabilities. It also supports probabilistic time series forecasting through a normalizing-flow-based head. Comprehensive experiments show that FLAME is an accurate and efficient out-of-the-box tool for decision intelligence.



%% file: Sections/Appendix.tex
\appendix
\section{Implementation Details}
\label{app: exp}

\subsection{Pretraining Corpus}
To pre-train our model, we utilize a broad collection of time series datasets drawn from multiple sources, including selected subsets from the Monash~\citep{godahewa2021monash}, UEA~\citep{uea}, and UCR~\citep{ucr} repositories, along with several widely-used benchmarks~\citep{prsa, tdbrain, pems, fred, nn5}. A full inventory of these datasets is provided in Table~\ref{tab: pretraining datasets}. We ensure that none of the pre-training datasets overlap with those used in downstream evaluations. The pretraining Weather dataset is univariate, whereas the downstream Weather dataset contains multivariate sequences.
We categorize the pre-training datasets into six domain-specific groups: Energy, Economic, Health, Transport, Web, and Nature. These datasets vary significantly in sampling rates, ranging from milliseconds to monthly intervals, capturing the diversity and richness of real-world time series. All datasets are decomposed into univariate sequences and FLAME is trained using a Channel Independent (CI) strategy.
\begin{table}[!htbp]
  \centering
  \caption{List of pretraining datasets.}
  \resizebox{0.7\linewidth}{!}{
    \begin{tabular}{c|c|c|c|c}
    \toprule
    \textbf{Domain} & \textbf{Dataset} & \textbf{Frequency} & \textbf{Point Number} & \textbf{Source} \\
    \midrule
    \multirow{5}[10]{*}{Energy} & Aus. Electricity Demand & Half Hourly & 1155264 & Monash~\citep{godahewa2021monash} \\
\cmidrule{2-5}          & Wind  & 4 Seconds & 7397147 & Monash~\citep{godahewa2021monash} \\
\cmidrule{2-5}          & Wind Farms & Minutely & 172178060 & Monash~\citep{godahewa2021monash} \\
\cmidrule{2-5}          & Solar Power & 4 Seconds & 7397222 & Monash~\citep{godahewa2021monash} \\
\cmidrule{2-5}          & London Smart Meters & Half Hourly & 166527216 & Monash~\citep{godahewa2021monash} \\
\midrule
    \multirow{3}[6]{*}{Economic} & FRED\_MD & Monthly & 77896 & ~\citep{fred} \\
\cmidrule{2-5}          & Bitcoin & Daily & 75364 & Monash~\citep{godahewa2021monash} \\
\cmidrule{2-5}          & NN5   & Daily & 87801 & ~\citep{nn5} \\
    \midrule
    \multirow{7}[14]{*}{Health} & MotorImagery & 0.001 Seconds & 72576000 & UEA~\citep{uea} \\
\cmidrule{2-5}          & SelfRegulationSCP1 & 0.004 Seconds & 3015936 & UEA~\citep{uea} \\
\cmidrule{2-5}          & SelfRegulationSCP2 & 0.004 Seconds & 3064320 & UEA~\citep{uea} \\
\cmidrule{2-5}          & AtrialFibrillation & 0.008 Seconds & 38400 & UEA~\citep{uea} \\
\cmidrule{2-5}          & PigArtPressure & -     & 624000 & UCR~\citep{ucr} \\
\cmidrule{2-5}          & PIGCVP & -     & 624000 & UCR~\citep{ucr} \\
\cmidrule{2-5}          & TDbrain & 0.002 Seconds & 79232703 & ~\citep{tdbrain} \\
    \midrule
    \multirow{4}[12]{*}{Transport} & Pems03 & 5 Minute & 9382464 & ~\citep{pems} \\
\cmidrule{2-5}          & Pems07 & 5 Minute & 24921792 & ~\citep{pems} \\
\cmidrule{2-5}          & Pems-bay & 5 Minute & 16937700 & ~\citep{pems} \\
\cmidrule{2-5}          & Pedestrian\_Counts & Hourly & 3132346 & Monash~\citep{godahewa2021monash} \\
    \midrule
    Web   & Web Traffic & Daily & 116485589 & Monash~\citep{godahewa2021monash} \\
    \midrule
    \multirow{11}[22]{*}{Nature} & Phoneme & -     & 2160640 & UCR~\citep{ucr} \\
\cmidrule{2-5}          & EigenWorms & -     & 27947136 & UEA~\citep{uea} \\
\cmidrule{2-5}          & PRSA  & Hourly & 4628448 & ~\citep{prsa} \\
\cmidrule{2-5}          & Temperature Rain & Daily & 23252200 & Monash~\citep{godahewa2021monash} \\
\cmidrule{2-5}          & StarLightCurves & -     & 9457664 & UCR~\citep{ucr} \\
\cmidrule{2-5}          & Worms & 0.033 Seconds & 232200 & UCR~\citep{ucr} \\
\cmidrule{2-5}          & Saugeen River Flow & Daily & 23741 & Monash~\citep{godahewa2021monash} \\
\cmidrule{2-5}          & Sunspot & Daily & 73924 & Monash~\citep{godahewa2021monash} \\
\cmidrule{2-5}          & Weather & Daily & 43032000 & Monash~\citep{godahewa2021monash} \\
\cmidrule{2-5}          & KDD Cup 2018 & Daily & 2942364 & Monash~\citep{godahewa2021monash} \\
\cmidrule{2-5}          & US Births & Daily & 7305  & Monash~\citep{godahewa2021monash} \\
    \bottomrule

    \end{tabular}}
  \label{tab: pretraining datasets}
\end{table}

\subsection{Baselines}
For zero-shot deterministic forecasting, we compare the FLAME family against 8 advanced foundation models: Sundial~\citep{liu2025sundial}, ROSE~\citep{wang2025rose}, Timer~\citep{timer}, MOIRAI~\citep{woo2024moirai}, Chronos~\citep{ansarichronos}, LightGTS~\citep{wang2025lightgts}, Time-MoE~\citep{timemoe}, and TinyTTM~\citep{tinyttm}. For zero-shot probabilistic forecasting, we compare the FLAME family with 4 foundation models: Sundial~\citep{liu2025sundial}, MOIRAI~\citep{woo2024moirai}, Chronos~\citep{ansarichronos}, and Lag-Llama~\citep{lagllama}, and 4 advanced end-to-end supervised models including TSDiff~\citep{TSDiff}, TimeGrad~\citep{TimeGrad}, CSDI~\citep{CSDI}, and GRU NVP~\citep{normalizing-flow}. The corresponding codebases and implementation details are summarized in Table~\ref{tab:baselines}.
\begin{table}[!htbp]
  \centering
  \caption{Code repositories for baselines.}
  \resizebox{0.7\linewidth}{!}{
    \begin{tabular}{c|c|l}
    \toprule
    \textbf{Model Types} & \textbf{Models} & \multicolumn{1}{c}{\textbf{Code Repositories}} \\
    \midrule
        \multirow{4}[12]{*}{End2End} & TSDiff   & \textcolor[rgb]{ .267,  .447,  .769}{https://github.com/amazon-science/unconditional-time-series-diffusion} \\
\cmidrule{2-3}          & CSDI & \textcolor[rgb]{ .267,  .447,  .769}{https://github.com/ermongroup/CSDI} \\
\cmidrule{2-3}          & TimeGrad & \textcolor[rgb]{ .267,  .447,  .769}{https://github.com/Zjh152/TimeGrad} \\
\cmidrule{2-3}          & GRU NVP  & \textcolor[rgb]{ .267,  .447,  .769}{https://github.com/microsoft/ProbTS} \\
    \midrule
    \multirow{9}[10]{*}{Foundation} &
    Sundial & \textcolor[rgb]{ .267,  .447,  .769}{https://github.com/thuml/Sundial} \\\cmidrule{2-3}  
    &ROSE & \textcolor[rgb]{ .267,  .447,  .769}{https://github.com/decisionintelligence/TSFM-Bench} \\\cmidrule{2-3}  
    &Timer & \textcolor[rgb]{ .267,  .447,  .769}{https://github.com/thuml/Large-Time-Series-Model} \\\cmidrule{2-3}  
\cmidrule{2-3}          & MOIRAI & \textcolor[rgb]{ .267,  .447,  .769}{https://github.com/redoules/moirai} \\
\cmidrule{2-3}          & Chronos & \textcolor[rgb]{ .267,  .447,  .769}{https://github.com/amazon-science/chronos-forecasting} \\
\cmidrule{2-3}          & LightGTS & \textcolor[rgb]{ .267,  .447,  .769}{https://github.com/decisionintelligence/LightGTS} \\
\cmidrule{2-3}          & Time-MoE & \textcolor[rgb]{ .267,  .447,  .769}{https://github.com/Time-MoE/Time-MoE} \\\cmidrule{2-3}  
&Lag-Llama & \textcolor[rgb]{ .267,  .447,  .769}{https://github.com/time-series-foundation-models/lag-llama} \\\cmidrule{2-3} 
&TinyTTM & \textcolor[rgb]{ .267,  .447,  .769}{https://huggingface.co/ibm-granite/granite-timeseries-ttm-r1} \\

    \bottomrule
    \end{tabular}}
  \label{tab:baselines}
\end{table}

\subsection{Benchmarks}
To thoroughly assess the effectiveness of FLAME, we conduct experiments on TSFM-Bench~\citep{li2025TSFM-Bench} and ProbTS~\citep{ProbTS}. Our evaluation spans 11 deterministic and 9 probabilistic forecasting datasets. In FLAME's flow-based head, samples are drawn i.i.d. from the base distribution and processed through layers to reach the target distribution. For probabilistic forecasting, we draw 100 random samples; for deterministic predictions, we average these 100 samples. 

For deterministic forecasting, we use ETTm1, ETTm2, ETTh1, ETTh2, Weather, Electricity, Traffic, Solar, PEMS08, Wind, and NYSE from TSFM-Bench. For most datasets, the prediction length is set to $L \in \{96, 192, 336, 720\}$, while NYSE uses $L \in \{24, 36, 48, 60\}$. 

For probabilistic forecasting, we adopt ETTm1, ETTm2, ETTh1, ETTh2, Weather, Electricity, Traffic, Exchange, and ILI from ProbTS. The prediction length is set to $L \in \{96, 192, 336, 720\}$ for most datasets, while ILI uses $L \in \{24, 36, 48, 60\}$.

All models are configured with the contextual length of the best performance recommended in their papers. It is important to clarify that datasets shared in both the TSFM-Bench and ProbTS (e.g., ETTh1, Traffic) are different in evaluation settings like strides. A summary of dataset statistics can be found in Table~\ref{Multivariate datasets}.

\begin{table*}[!htbp]
\caption{Statistics of benchmark datasets.}
\label{Multivariate datasets}
\resizebox{\linewidth}{!}{
\begin{tabular}{ccccccccc}
\toprule
Dataset      & Domain      & Frequency & Lengths & Dim & Split &Stride &Benchmark & Description\\ \midrule
ETTm1        & Electricity & 15 mins   & 57,600      & 7        & 6:2:2 & 1 & TSFM-Bench & Power transformer 1, comprising seven indicators such as oil temperature and useful load\\
ETTm2        & Electricity & 15 mins   & 57,600      & 7        & 6:2:2 & 1 & TSFM-Bench& Power transformer 2, comprising seven indicators such as oil temperature and useful load\\
ETTh1        & Electricity & 1 hour     & 14,400      & 7        & 6:2:2 & 1 & TSFM-Bench& Power transformer 1, comprising seven indicators such as oil temperature and useful load\\
ETTh2        & Electricity & 1 hour    & 14,400      & 7        & 6:2:2 & 1 & TSFM-Bench& Power transformer 2, comprising seven indicators such as oil temperature and useful load\\
Weather      & Environment & 10 mins   & 52,696      & 21       & 7:1:2 & 1& TSFM-Bench& Recorded every 10 minutes throughout 2020, with 21 meteorological indicators\\
Electricity  & Electricity & 1 hour    & 26,304      & 321      & 7:1:2 & 1& TSFM-Bench & Electricity records the electricity consumption in kWh every 1 hour from 2012 to 2014\\
Traffic      & Traffic     & 1 hour    & 17,544      & 862      & 7:1:2 & 1 & TSFM-Bench& Road occupancy rates measured by 862 sensors on San Francisco Bay area freeways\\
Solar        & Energy      & 10 mins   & 52,560      & 137      & 6:2:2 & 1 & TSFM-Bench&Solar production records collected from 137 PV plants in Alabama \\
PEMS08       & Traffic     & 5 mins    & 17,856      & 170      & 6:2:2 & 1 & TSFM-Bench&Traffic
flow time series collected from the CalTrans PeMS\\
Wind         & Energy & 15 mins   & 48,673      & 7        & 7:1:2& 1 & TSFM-Bench & Wind power records from 2020-2021 at 15-minute intervals \\
NYSE         & Stock       & 1 day     & 1,243       & 5        & 7:1:2 & 1 & TSFM-Bench& Records opening price, closing price, trading volume, lowest price, and highest price\\
\midrule
ETTm1        & Electricity & 15 mins   & 57,600      & 7        & 6:2:2 & 96 & ProbTS & Power transformer 1, comprising seven indicators such as oil temperature and useful load\\
ETTm2        & Electricity & 15 mins   & 57,600      & 7        & 6:2:2 & 96 & ProbTS & Power transformer 2, comprising seven indicators such as oil temperature and useful load\\
ETTh1        & Electricity & 1 hour     & 14,400      & 7        & 6:2:2 & 96 & ProbTS & Power transformer 1, comprising seven indicators such as oil temperature and useful load\\
ETTh2        & Electricity & 1 hour    & 14,400      & 7        & 6:2:2 & 96 & ProbTS & Power transformer 2, comprising seven indicators such as oil temperature and useful load\\
Weather      & Environment & 10 mins   & 52,696      & 21       & 7:1:2 & 96 & ProbTS& Recorded every 10 minutes throughout 2020, with 21 meteorological indicators\\
Electricity  & Electricity & 1 hour    & 26,304      & 321      & 7:1:2 & 96 & ProbTS & Electricity records the electricity consumption in kWh every 1 hour from 2012 to 2014\\
Traffic      & Traffic     & 1 hour    & 17,544      & 862      & 7:1:2 & 96 & ProbTS & Road occupancy rates measured by 862 sensors on San Francisco Bay area freeways\\
Exchange & Economic    & 1 day      & 7,588       & 8        & 7:1:2 & 96 & ProbTS& ExchangeRate collects the daily exchange rates of eight countries\\
ILI          & Health      & 1 week     & 966        & 7        & 7:1:2 & 96 & ProbTS& Recorded indicators of patient data from Centers for Disease Control and Prevention\\
\bottomrule
\end{tabular}}
\end{table*}

\subsection{Experimental Settings}
\paragraph{Pretraining}
We implement the FLAME family using Distributed Data Parallel in PyTorch~\citep{paszke2019pytorch}, and conduct all experiments on 8 NVIDIA A800 GPUs with 80GB GPU memory. The model is optimized using the AdamW optimizer with an initial learning rate of $5 \times 10^{-5}$. To progressively reduce the learning rate during training, we employ a step decay schedule via the StepLR scheduler.
During pre-training, we use 12 historical tokens and 4 prediction tokens, with a reference patch size $p = 48$. The batch size is set to 8,192. Full details on model configurations and parameter sizes are summarized in Table~\ref{tab:config}.

\begin{table}[htbp]
  \centering
   \captionsetup{width=0.8\linewidth}
  \caption{Detailed model configurations of the FLAME family and corresponding parameter counts. }
  \resizebox{0.8\linewidth}{!}{
    \begin{tabular}{c|c|c|c|c|c|c}
    \toprule
    \textbf{Models} & \textbf{Encoder Layers} & \textbf{Decoder Layers} &\textbf{Flow Layers} & \textbf{Model Dim $d$}& \textbf{FFN Dim $d^\prime$} & \textbf{Parameters} \\
    \midrule
    FLAME Small & 1     & 1  &3   & 256   & 512   & 2M \\
    \midrule
    FLAME Base & 1     & 3  &5   & 256   & 512   & 6M \\    \midrule
    FLAME Large & 1    & 6  &7   & 256   & 512   & 10M \\
    \bottomrule
    \end{tabular}
  \label{tab:config}}
\end{table}

\paragraph{Downstream Forecasting}
For downstream forecasting tasks, we apply periodic patching strategies adapted to each dataset's temporal characteristics. The number of past tokens remains fixed at 11, while the prediction lengths are evaluated at 96, 192, 336, and 720 steps.

We also address the \textit{"Drop Last"} issue highlighted in recent studies~\citep{qiu2024tfb,qiu2025tab,li2025TSFM-Bench}, where setting $\textit{drop\_last}$=True during test evaluation may lead to misleading results due to incomplete batches. To ensure consistency and fairness, we set $\textit{drop\_last}$=False for all baseline models in our experiments.

\subsection{Evaluation Metrics}
Regarding evaluation metrics, following the experimental setup in TSFM-Bench, we adopt Mean Squared Error (MSE) and Mean Absolute Error (MAE) as evaluation metrics for deterministic forecasting. For probabilistic forecasting, we use Continuous Ranked Probability Score (CRPS) and Normalized Mean Absolute Error (NMAE) in ProbTS. Consider a forecast with $K$ variables and a prediction horizon of $T$ steps.

\paragraph{Mean Squared Error (MSE)}
 The Mean Squared Error quantifies the average of the squared differences between predicted values and their corresponding ground truths. It penalizes larger errors more heavily due to the squaring operation, making it sensitive to outliers. Formally, it is defined as:
\begin{gather}
    \textrm{MSE} = \frac{1}{K \times T} \sum_{k=1}^{K} \sum_{t=1}^{T} (x^k_t - \hat{x}^k_t)^2,
\end{gather}
where $K$ denotes the number of variables, $T$ the prediction horizon, $x^k_t$ the true value, and $\hat{x}^k_t$ the predicted value.

\paragraph{Mean Absolute Error (MAE)}
 The Mean Absolute Error calculates the average magnitude of prediction errors without considering their direction. By focusing on the absolute differences, MAE provides a robust and interpretable measure of accuracy:

\begin{gather}
    \textrm{MAE} = \frac{1}{K \times T} \sum_{k=1}^{K} \sum_{t=1}^{T} |x^k_t - \hat{x}^k_t|,
\end{gather}
where all terms follow the same definition as above. Unlike MSE, MAE treats all errors equally and is less sensitive to large deviations.

\paragraph{Continuous Ranked Probability Score (CRPS)}
 The CRPS assesses the quality of probabilistic forecasts by comparing the predicted cumulative distribution function (CDF) $F$ with the observed outcome $x$. It is computed as:

\begin{gather}
    \textrm{CRPS} = \int_{\mathds{R}} (F(z) - \mathds{I}\{x \leq z\})^2 dz,
\end{gather}

where $\mathds{I}\{x \leq z\}$ is the indicator function. CRPS evaluates both calibration and sharpness; its expected value is minimized when the predictive distribution matches the true conditional distribution.
 In practice, we approximate CRPS using the empirical CDF $\hat{F}(z) = \frac{1}{n} \sum_{i=1}^{n} \mathds{I} \{ X_i \leq z \}$ based on $n=100$ scalar samples $X_i$ drawn from the corresponding predictive marginal.

\paragraph{Normalized Mean Absolute Error (NMAE)}
 The NMAE extends MAE by normalizing it against the total magnitude of ground-truth values, thereby enabling fair comparison across datasets with varying scales. Its formula is:

\begin{gather}
    \textrm{NMAE} = \frac{\sum_{k=1}^{K} \sum_{t=1}^{T} |x^k_t - \hat{x}^k_t|}{\sum_{k=1}^{K} \sum_{t=1}^{T} |x^k_t|}
\end{gather}

\section{Full Model Analysis}
\label{app: model analytics}

\subsection{The Convergence of LegT based Local Perception}
\begin{figure}[!htbp] 
  \centering         
  \includegraphics[width=0.97\textwidth]{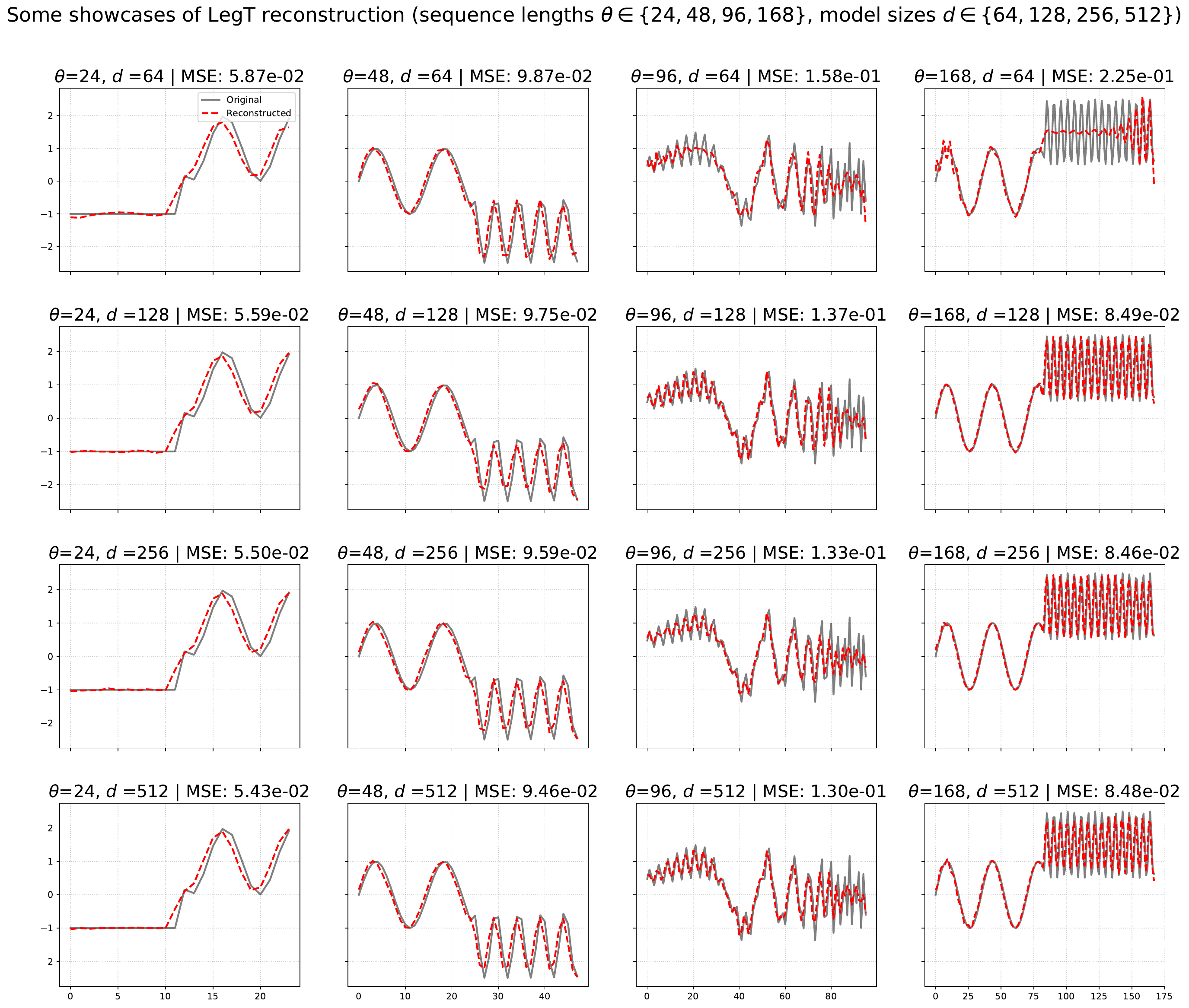} 
  \caption{The reconstruction results of LegT based Local Perception Module.} 
  \label{fig:legt reconstruction}       
\end{figure}

In FLAME, $\theta$ is a varying parameter related to the input's periodicity. Empirically, common dataset periods (e.g., ETT, Electricity, Traffic) are typically 24, 48, 96, or 168, which are smaller than $d=256$. Theorem~\ref{theory: 1} gives a projection error bound that decreases with $d$ for fixed window length and smoothness; the condition $d>\theta$ alone does not guarantee a small error for the discretized memory. To validate this, we randomly select 100 time-series segments for each length $\theta \in \lbrace 24, 48, 96, 168 \rbrace$, compress them using LegT with model sizes $d \in \lbrace 64, 128, 256, 512 \rbrace$, then reconstruct them and report the average Mean Squared Error. The experiments and visual showcases are provided in Table~\ref{tab:legt_reconstruction_mse} and Figure~\ref{fig:legt reconstruction} . We observe that $d=256$ sufficiently maintains a low approximation error.

\begin{table}[!htbp]
    \centering
    \small
    \caption{MSE values of LegT reconstruction results. Each value is calculated over 100 time series samples. We observe that $d=256$ is sufficient to maintain low approximation error under common periods $\theta \in \{24,48,96,168\}$. Although $d=512$ further improves accuracy, the improvement is marginal. Therefore, we choose $d=256$ to balance approximation accuracy and efficiency.}
    \label{tab:legt_reconstruction_mse}
    \begin{tabular}{c|cccc}
        \toprule
        \textbf{Seq. Length $\theta$} & \textbf{$d=64$} & \textbf{$d=128$} & \textbf{$d=256$} & \textbf{$d=512$} \\
        \midrule
        24  & 0.088 & 0.084 & \textbf{0.083} & \textbf{0.083} \\
        48  & 0.124 & 0.121 & 0.118 & \textbf{0.117} \\
        96  & 0.183 & 0.146 & 0.141 & \textbf{0.139} \\
        168 & 0.233 & 0.094 & 0.093 & \textbf{0.092} \\
        \bottomrule
    \end{tabular}
\end{table}

\subsection{Analysis of Sampling Number}
Given that FLAME models distributions across the forecasting horizon, we conduct an in-depth study on the scalability of generative forecasting by examining the correlations between the sampling number and performance, as presented in Table~\ref{tab: sampling num}. Specifically, we choose four sampling numbers: 10, 20, 50, and 100. The findings indicate that CRPS and NMAE generally improve as the number of samples increases, although individual settings do not improve monotonically. Moreover, a sampling number of 100 leads to excellent performance, and this is unlikely to impose significant efficiency burdens in practical applications.

\begin{table}[!htbp]
\centering
 \captionsetup{width=0.5\linewidth}
\caption{Full results of probabilistic forecasting experiments with different sampling numbers.}
\label{tab: sampling num}

\resizebox{0.55\linewidth}{!}{
\begin{tabular}{cc|cc|cc|cc|cc}
\toprule
\multicolumn{2}{c|}{Models} & \multicolumn{2}{c|}{FLAME-10} & \multicolumn{2}{c|}{FLAME-20} & \multicolumn{2}{c|}{FLAME-50} & \multicolumn{2}{c}{FLAME-100} \\
\multicolumn{2}{c|}{Metric} & CRPS & NMAE & CRPS & NMAE & CRPS & NMAE & CRPS & NMAE \\
\midrule
 & 96 & 1.255 & \textcolor{blue}{\underline{ 0.412}} & 0.388 & 0.477 & \textcolor{blue}{\underline{ 0.323}} & 0.418 & \textcolor{red}{ \textbf{0.315}} & \textcolor{red}{ \textbf{0.407}} \\
 & 192 & 0.905 & 0.444 & 0.372 & 0.433 & \textcolor{blue}{\underline{ 0.358}} & \textcolor{blue}{\underline{ 0.432}} & \textcolor{red}{ \textbf{0.337}} & \textcolor{red}{ \textbf{0.426}} \\
 & 336 & 0.369 & 0.468 & 0.448 & 0.682 & \textcolor{blue}{\underline{ 0.352}} & \textcolor{blue}{\underline{ 0.448}} & \textcolor{red}{ \textbf{0.351}} & \textcolor{red}{ \textbf{0.447}} \\
 & 720 & \textcolor{blue}{\underline{ 0.388}} & \textcolor{blue}{\underline{ 0.486}} & 0.394 & 0.491 & 0.394 & 0.488 & \textcolor{red}{ \textbf{0.383}} & \textcolor{red}{ \textbf{0.476}} \\
 \cmidrule{2-10}
\multirow{-5}{*}{\rotatebox{90}{ETTm1}} & avg & 0.729 & 0.453 & 0.401 & 0.521 & \textcolor{blue}{\underline{ 0.357}} & \textcolor{blue}{\underline{ 0.447}} & \textcolor{red}{ \textbf{0.347}} & \textcolor{red}{ \textbf{0.439}} \\
\midrule
 & 96 & 0.134 & 0.173 & 0.129 & 0.167 & \textcolor{blue}{\underline{ 0.125}} & \textcolor{blue}{\underline{ 0.163}} & \textcolor{red}{ \textbf{0.124}} & \textcolor{red}{ \textbf{0.155}} \\
 & 192 & 0.208 & 0.196 & 0.187 & 0.190 & \textcolor{blue}{\underline{ 0.144}} & \textcolor{blue}{\underline{ 0.186}} & \textcolor{red}{ \textbf{0.143}} & \textcolor{red}{ \textbf{0.175}} \\
 & 336 & 0.477 & 0.212 & 0.427 & 0.208 & \textcolor{blue}{\underline{ 0.159}} & \textcolor{blue}{\underline{ 0.204}} & \textcolor{red}{ \textbf{0.158}} & \textcolor{red}{ \textbf{0.193}} \\
 & 720 & 0.662 & 0.236 & 0.631 & 0.232 & \textcolor{blue}{\underline{ 0.187}} & \textcolor{blue}{\underline{ 0.229}} & \textcolor{red}{ \textbf{0.184}} & \textcolor{red}{ \textbf{0.219}} \\
  \cmidrule{2-10}
\multirow{-5}{*}{\rotatebox{90}{ETTm2}} & avg & 0.370 & 0.204 & 0.344 & 0.199 & \textcolor{blue}{\underline{ 0.154}} & \textcolor{blue}{\underline{ 0.196}} & \textcolor{red}{ \textbf{0.152}} & \textcolor{red}{ \textbf{0.186}} \\
\midrule
 & 96 & 0.299 & 0.346 & 0.269 & 0.331 & \textcolor{blue}{\underline{ 0.262}} & \textcolor{blue}{\underline{ 0.324}} & \textcolor{red}{ \textbf{0.259}} & \textcolor{red}{ \textbf{0.310}} \\
 & 192 & 0.294 & 0.365 & 0.285 & 0.354 & \textcolor{blue}{\underline{ 0.279}} & \textcolor{blue}{\underline{ 0.347}} & \textcolor{red}{ \textbf{0.278}} & \textcolor{red}{ \textbf{0.342}} \\
 & 336 & 0.309 & 0.383 & 0.298 & 0.372 & \textcolor{blue}{\underline{ 0.293}} & \textcolor{blue}{\underline{ 0.365}} & \textcolor{red}{ \textbf{0.290}} & \textcolor{red}{ \textbf{0.362}} \\
 & 720 & 0.328 & 0.406 & 0.317 & 0.396 & \textcolor{blue}{\underline{ 0.311}} & \textcolor{blue}{\underline{ 0.390}} & \textcolor{red}{ \textbf{0.303}} & \textcolor{red}{ \textbf{0.367}} \\
  \cmidrule{2-10}
\multirow{-5}{*}{\rotatebox{90}{ETTh1}} & avg & 0.308 & 0.375 & 0.292 & 0.363 & \textcolor{blue}{\underline{ 0.286}} & \textcolor{blue}{\underline{ 0.357}} & \textcolor{red}{ \textbf{0.283}} & \textcolor{red}{ \textbf{0.345}} \\
\midrule
 & 96 & 0.158 & 0.200 & 0.150 & 0.191 & \textcolor{blue}{\underline{ 0.146}} & \textcolor{blue}{\underline{ 0.188}} & \textcolor{red}{ \textbf{0.144}} & \textcolor{red}{ \textbf{0.186}} \\
 & 192 & 0.167 & 0.211 & 0.162 & 0.207 & \textcolor{blue}{\underline{ 0.160}} & \textcolor{blue}{\underline{ 0.206}} & \textcolor{red}{ \textbf{0.159}} & \textcolor{red}{ \textbf{0.204}} \\
 & 336 & 0.181 & 0.226 & 0.176 & 0.222 & \textcolor{blue}{\underline{ 0.174}} & \textcolor{blue}{\underline{ 0.221}} & \textcolor{red}{ \textbf{0.172}} & \textcolor{red}{ \textbf{0.217}} \\
 & 720 & 0.191 & 0.234 & 0.187 & 0.231 & \textcolor{blue}{\underline{ 0.183}} & \textcolor{blue}{\underline{ 0.229}} & \textcolor{red}{ \textbf{0.182}} & \textcolor{red}{ \textbf{0.228}} \\
  \cmidrule{2-10}
\multirow{-5}{*}{\rotatebox{90}{ETTh2}} & avg & 0.174 & 0.218 & 0.169 & 0.213 & \textcolor{blue}{\underline{ 0.166}} & \textcolor{blue}{\underline{ 0.211}} & \textcolor{red}{ \textbf{0.164}} & \textcolor{red}{ \textbf{0.209}} \\
\midrule
 & 96 & 0.080 & 0.098 & 0.178 & 0.093 & \textcolor{blue}{\underline{ 0.073}} & \textcolor{blue}{\underline{ 0.091}} & \textcolor{red}{ \textbf{0.067}} & \textcolor{red}{ \textbf{0.083}} \\
 & 192 & 0.103 & 0.112 & 0.092 & 0.105 & \textcolor{blue}{\underline{ 0.084}} & \textcolor{blue}{\underline{ 0.103}} & \textcolor{red}{ \textbf{0.068}} & \textcolor{red}{ \textbf{0.082}} \\
 & 336 & 1.382 & 0.121 & 0.233 & 0.117 & \textcolor{blue}{\underline{ 0.092}} & \textcolor{blue}{\underline{ 0.114}} & \textcolor{red}{ \textbf{0.079}} & \textcolor{red}{ \textbf{0.094}} \\
 & 720 & 2.471 & 0.133 & 1.384 & 0.129 & \textcolor{blue}{\underline{ 0.101}} & \textcolor{blue}{\underline{ 0.127}} & \textcolor{red}{ \textbf{0.083}} & \textcolor{red}{ \textbf{0.100}} \\
  \cmidrule{2-10}
\multirow{-5}{*}{\rotatebox{90}{Weather}} & avg & 1.009 & 0.116 & 0.472 & 0.111 & \textcolor{blue}{\underline{ 0.088}} & \textcolor{blue}{\underline{ 0.109}} & \textcolor{red}{ \textbf{0.074}} & \textcolor{red}{ \textbf{0.090}} \\
\midrule
 & 96 & 1.734 & 0.097 & 0.072 & 0.092 & \textcolor{blue}{\underline{ 0.070}} & \textcolor{blue}{\underline{ 0.090}} & \textcolor{red}{ \textbf{0.069}} & \textcolor{red}{ \textbf{0.089}} \\
 & 192 & 0.093 & 0.119 & 1.662 & 0.116 & \textcolor{blue}{\underline{ 0.086}} & \textcolor{blue}{\underline{ 0.112}} & \textcolor{red}{ \textbf{0.072}} & \textcolor{red}{ \textbf{0.088}} \\
 & 336 & 0.662 & 0.147 & 0.375 & 0.143 & \textcolor{blue}{\underline{ 0.107}} & \textcolor{blue}{\underline{ 0.140}} & \textcolor{red}{ \textbf{0.081}} & \textcolor{red}{ \textbf{0.096}} \\
 & 720 & 2.406 & 0.173 & 1.354 & 0.168 & \textcolor{blue}{\underline{ 0.125}} & \textcolor{blue}{\underline{ 0.166}} & \textcolor{red}{ \textbf{0.100}} & \textcolor{red}{ \textbf{0.116}} \\
  \cmidrule{2-10}
\multirow{-5}{*}{\rotatebox{90}{Electricity}} & avg & 1.224 & 0.134 & 0.866 & 0.130 & \textcolor{blue}{\underline{ 0.097}} & \textcolor{blue}{\underline{ 0.127}} & \textcolor{red}{ \textbf{0.081}} & \textcolor{red}{ \textbf{0.097}} \\
\midrule
 & 96 & 0.562 & 0.301 & 0.491 & 0.309 & \textcolor{blue}{\underline{ 0.224}} & \textcolor{blue}{\underline{ 0.273}} & \textcolor{red}{ \textbf{0.220}} & \textcolor{red}{ \textbf{0.268}} \\
 & 192 & 0.811 & 0.324 & 0.681 & 0.333 & \textcolor{blue}{\underline{ 0.240}} & \textcolor{blue}{\underline{ 0.298}} & \textcolor{red}{ \textbf{0.238}} & \textcolor{red}{ \textbf{0.294}} \\
 & 336 & 2.911 & 0.348 & 3.284 & 0.364 & \textcolor{blue}{\underline{ 0.257}} & \textcolor{blue}{\underline{ 0.323}} & \textcolor{red}{ \textbf{0.254}} & \textcolor{red}{ \textbf{0.319}} \\
 & 720 & 3.491 & 0.378 & 2.824 & 0.484 & \textcolor{blue}{\underline{ 0.278}} & \textcolor{blue}{\underline{ 0.354}} & \textcolor{red}{ \textbf{0.276}} & \textcolor{red}{ \textbf{0.351}} \\
  \cmidrule{2-10}
\multirow{-5}{*}{\rotatebox{90}{Traffic}} & avg & 1.944 & 0.338 & 1.820 & 0.373 & \textcolor{blue}{\underline{ 0.250}} & \textcolor{blue}{\underline{ 0.312}} & \textcolor{red}{ \textbf{0.247}} & \textcolor{red}{ \textbf{0.308}} \\
\midrule
 & 96 & 0.023 & 0.027 & 0.022 & 0.027 & \textcolor{blue}{\underline{ 0.022}} & \textcolor{blue}{\underline{ 0.027}} & \textcolor{red}{ \textbf{0.020}} & \textcolor{red}{ \textbf{0.026}} \\
 & 192 & 0.032 & 0.037 & 0.032 & 0.037 & \textcolor{blue}{\underline{ 0.031}} & \textcolor{blue}{\underline{ 0.037}} & \textcolor{blue}{ \textbf{0.031}} & \textcolor{red}{ \textbf{0.035}} \\
 & 336 & 0.045 & 0.049 & 0.044 & 0.049 & \textcolor{blue}{\underline{ 0.044}} & \textcolor{blue}{\underline{ 0.049}} & \textcolor{blue}{ \textbf{0.044}} & \textcolor{red}{ \textbf{0.046}} \\
 & 720 & 1.234 & 0.098 & 0.145 & 0.082 & \textcolor{blue}{\underline{ 0.074}} & \textcolor{blue}{\underline{ 0.079}} & \textcolor{red}{ \textbf{0.073}} & \textcolor{red}{ \textbf{0.079}} \\
  \cmidrule{2-10}
\multirow{-5}{*}{\rotatebox{90}{Exchange}} & avg & 0.334 & 0.053 & 0.061 & 0.049 & \textcolor{blue}{\underline{ 0.043}} & \textcolor{blue}{\underline{ 0.048}} & \textcolor{red}{ \textbf{0.042}} & \textcolor{red}{ \textbf{0.047}} \\
 \midrule
 & 24 & \textcolor{blue}{\underline{ 0.156}} & \textcolor{blue}{\underline{ 0.188}} & 0.157 & 0.197 & 0.157 & 0.194 & \textcolor{red}{ \textbf{0.136}} & \textcolor{red}{ \textbf{0.168}} \\
 & 36 & 0.178 & 0.216 & 0.181 & 0.225 & \textcolor{blue}{\underline{ 0.177}} & \textcolor{blue}{\underline{ 0.214}} & \textcolor{red}{ \textbf{0.145}} & \textcolor{red}{ \textbf{0.179}} \\
 & 48 & 0.197 & \textcolor{blue}{\underline{ 0.234}} & 0.201 & 0.244 & \textcolor{blue}{\underline{ 0.196}} & 0.235 & \textcolor{red}{ \textbf{0.164}} & \textcolor{red}{ \textbf{0.200}} \\
 & 60 & 0.196 & 0.234 & \textcolor{blue}{\underline{ 0.190}} & \textcolor{blue}{\underline{ 0.231}} & 0.192 & 0.233 & \textcolor{red}{ \textbf{0.169}} & \textcolor{red}{ \textbf{0.204}} \\
  \cmidrule{2-10}
\multirow{-5}{*}{\rotatebox{90}{ILI}} & avg & 0.182 & \textcolor{blue}{\underline{ 0.218}} & 0.182 & 0.224 & \textcolor{blue}{\underline{ 0.181}} & 0.219 & \textcolor{red}{ \textbf{0.154}} & \textcolor{red}{ \textbf{0.188}} \\
\bottomrule
\end{tabular}}

\end{table}

\subsection{Analysis of Local-Perception}
As shown in Table~\ref{tab: local perception}, we conduct ablation studies to assess the effectiveness of the proposed Local-Perception module.
Specifically, we consider three variants: removing the entire Local-Perception module, denoted as ``w/o Local-Perception''; disabling Period Alignment, denoted as ``w/o Period Alignment''; and making the LegT matrices $A$ and $B$ learnable, denoted as ``Learnable LegT''.

The results demonstrate that Local-Perception plays a crucial role in zero-shot forecasting.
Removing the entire module leads to the most severe degradation, indicating that Period Alignment and LegT jointly contribute to constructing informative local environments for patch-wise tokens.
We further observe that Period Alignment is especially important for datasets with complex periodic structures, such as Electricity and Traffic.
Without Period Alignment, the model fixes the local window length $\theta$ to the patch size and thus fails to anchor each token to its period-aware local environment, resulting in substantially worse zero-shot performance.

In addition, the Learnable LegT variant also underperforms the default design with fixed matrices.
This suggests that the analytically initialized LegT matrices preserve a stable Legendre basis for online function approximation, while making $A$ and $B$ learnable may shift the underlying orthogonal space during large-scale pretraining and weaken the approximation capability.
Overall, these results verify that the effectiveness of Local-Perception comes from the complementary design of Period Alignment and fixed LegT: Period Alignment determines an appropriate period-aware environment boundary, and LegT compresses the aligned environment into a fixed-size representation without losing its online approximation property.

\begin{table}[!htbp]
\centering
 \captionsetup{width=0.5\linewidth}
\caption{Experimental results on Local Perception analysis, comparing configurations without Local Perception (LP), without Period Alignment (P.A.), and with Learnable LegT (L.L.).}

\label{tab: local perception}
\resizebox{0.55\linewidth}{!}{
\begin{tabular}{cc|cc|cc|cc|cc}
\toprule
\multicolumn{2}{c|}{Models} & \multicolumn{2}{c|}{w/o LP} & \multicolumn{2}{c|}{w/o P.A.} & \multicolumn{2}{c|}{with L.L.} & \multicolumn{2}{c}{FLAME Large} \\
\multicolumn{2}{c|}{Metric} & MSE & CRPS & MSE & CRPS & MSE & CRPS & MSE & CRPS \\
\midrule
 & 96 & 0.365 & 0.352 & 0.362 & 0.332 & \textcolor{red}{ \textbf{0.297}} & \textcolor{blue}{\underline{ 0.317}} & \textcolor{blue}{\underline{ 0.301}} & \textcolor{red}{ \textbf{0.315}} \\
 & 192 & 0.389 & 0.385 & 0.387 & 0.364 & \textcolor{red}{ \textbf{0.327}} & \textcolor{red}{ \textbf{0.335}} & \textcolor{blue}{\underline{ 0.328}} & \textcolor{blue}{\underline{ 0.337}} \\
 & 336 & 0.417 & 0.407 & 0.408 & 0.398 & \textcolor{blue}{\underline{ 0.353}} & \textcolor{blue}{\underline{ 0.359}} & \textcolor{red}{ \textbf{0.351}} & \textcolor{red}{ \textbf{0.351}} \\
 & 720 & 0.470 & 0.462 & 0.483 & 0.423 & \textcolor{blue}{\underline{ 0.397}} & \textcolor{blue}{\underline{ 0.390}} & \textcolor{red}{ \textbf{0.392}} & \textcolor{red}{ \textbf{0.383}} \\
   \cmidrule{2-10}
\multirow{-5}{*}{\rotatebox{90}{ETTm1}} & avg & 0.410 & 0.402 & 0.410 & 0.379 & \textcolor{blue}{\underline{ 0.344}} & \textcolor{blue}{\underline{ 0.350}} & \textcolor{red}{ \textbf{0.343}} & \textcolor{red}{ \textbf{0.347}} \\
\midrule
 & 96 & 0.229 & 0.144 & 0.216 & 0.133 & \textcolor{red}{ \textbf{0.178}} & \textcolor{red}{ \textbf{0.121}} & \textcolor{blue}{\underline{ 0.186}} & \textcolor{blue}{\underline{ 0.124}} \\
 & 192 & 0.283 & 0.154 & 0.258 & 0.149 & \textcolor{blue}{\underline{ 0.253}} & \textcolor{blue}{\underline{ 0.146}} & \textcolor{red}{ \textbf{0.247}} & \textcolor{red}{ \textbf{0.143}} \\
 & 336 & 0.311 & 0.186 & 0.304 & 0.171 & \textcolor{red}{ \textbf{0.297}} & \textcolor{blue}{\underline{ 0.164}} & \textcolor{blue}{\underline{ 0.300}} & \textcolor{red}{ \textbf{0.158}} \\
 & 720 & \textcolor{blue}{\underline{ 0.362}} & 0.200 & 0.363 & 0.195 & 0.365 & \textcolor{blue}{\underline{ 0.191}} & \textcolor{red}{ \textbf{0.357}} & \textcolor{red}{ \textbf{0.184}} \\
   \cmidrule{2-10}
\multirow{-5}{*}{\rotatebox{90}{ETTm2}} & avg & 0.296 & 0.171 & 0.285 & 0.162 & \textcolor{blue}{\underline{ 0.273}} & \textcolor{blue}{\underline{ 0.156}} & \textcolor{red}{ \textbf{0.273}} & \textcolor{red}{ \textbf{0.152}} \\
\midrule
 & 96 & 0.372 & 0.283 & 0.362 & 0.279 & \textcolor{red}{ \textbf{0.352}} & \textcolor{red}{ \textbf{0.255}} & \textcolor{blue}{\underline{ 0.358}} & \textcolor{blue}{\underline{ 0.259}} \\
 & 192 & 0.418 & 0.304 & \textcolor{blue}{\underline{ 0.402}} & 0.291 & 0.408 & \textcolor{blue}{\underline{ 0.291}} & \textcolor{red}{ \textbf{0.392}} & \textcolor{red}{ \textbf{0.278}} \\
 & 336 & 0.433 & 0.322 & 0.428 & \textcolor{blue}{\underline{ 0.300}} & \textcolor{blue}{\underline{ 0.413}} & 0.304 & \textcolor{red}{ \textbf{0.411}} & \textcolor{red}{ \textbf{0.290}} \\
 & 720 & 0.450 & 0.376 & 0.459 & \textcolor{blue}{\underline{ 0.318}} & \textcolor{red}{ \textbf{0.438}} & 0.328 & \textcolor{blue}{\underline{ 0.444}} & \textcolor{red}{ \textbf{0.303}} \\
   \cmidrule{2-10}
\multirow{-5}{*}{\rotatebox{90}{ETTh1}} & avg & 0.418 & 0.321 & 0.413 & 0.297 & \textcolor{blue}{\underline{ 0.403}} & \textcolor{blue}{\underline{ 0.295}} & \textcolor{red}{ \textbf{0.401}} & \textcolor{red}{ \textbf{0.283}} \\
\midrule
 & 96 & 0.312 & 0.148 & \textcolor{blue}{\underline{ 0.303}} & \textcolor{blue}{\underline{ 0.145}} & 0.311 & 0.161 & \textcolor{red}{ \textbf{0.300}} & \textcolor{red}{ \textbf{0.144}} \\
 & 192 & 0.368 & 0.167 & \textcolor{blue}{\underline{ 0.358}} & \textcolor{blue}{\underline{ 0.159}} & 0.364 & 0.178 & \textcolor{red}{ \textbf{0.351}} & \textcolor{red}{ \textbf{0.159}} \\
 & 336 & 0.384 & 0.182 & \textcolor{blue}{\underline{ 0.375}} & \textcolor{blue}{\underline{ 0.174}} & 0.391 & 0.199 & \textcolor{red}{ \textbf{0.374}} & \textcolor{red}{ \textbf{0.172}} \\
 & 720 & 0.425 & 0.188 & 0.422 & 0.186 & \textcolor{blue}{\underline{ 0.422}} & \textcolor{blue}{\underline{ 0.185}} & \textcolor{red}{ \textbf{0.416}} & \textcolor{red}{ \textbf{0.182}} \\
   \cmidrule{2-10}
\multirow{-5}{*}{\rotatebox{90}{ETTh2}} & avg & 0.372 & 0.171 & \textcolor{blue}{\underline{ 0.365}} & \textcolor{blue}{\underline{ 0.166}} & 0.372 & 0.181 & \textcolor{red}{ \textbf{0.360}} & \textcolor{red}{ \textbf{0.164}} \\
\midrule
 & 96 & 0.233 & 0.076 & 0.195 & 0.071 & \textcolor{blue}{\underline{ 0.168}} & \textcolor{blue}{\underline{ 0.067}} & \textcolor{red}{ \textbf{0.152}} & \textcolor{red}{ \textbf{0.067}} \\
 & 192 & 0.270 & 0.099 & 0.225 & 0.074 & \textcolor{blue}{\underline{ 0.214}} & \textcolor{blue}{\underline{ 0.069}} & \textcolor{red}{ \textbf{0.199}} & \textcolor{red}{ \textbf{0.068}} \\
 & 336 & 0.295 & 0.112 & 0.261 & 0.088 & \textcolor{blue}{\underline{ 0.255}} & \textcolor{red}{ \textbf{0.078}} & \textcolor{red}{ \textbf{0.250}} & \textcolor{blue}{\underline{ 0.079}} \\
 & 720 & 0.312 & 0.127 & \textcolor{blue}{\underline{ 0.311}} & 0.098 & 0.312 & \textcolor{red}{ \textbf{0.081}} & \textcolor{red}{ \textbf{0.307}} & \textcolor{blue}{\underline{ 0.083}} \\
   \cmidrule{2-10}
\multirow{-5}{*}{\rotatebox{90}{Weather}} & avg & 0.278 & 0.104 & 0.248 & 0.083 & \textcolor{blue}{\underline{ 0.237}} & \textcolor{red}{ \textbf{0.074}} & \textcolor{red}{ \textbf{0.227}} & \textcolor{blue}{\underline{ 0.074}} \\
\midrule
 & 96 & 0.214 & 0.088 & 0.202 & 0.077 & \textcolor{blue}{\underline{ 0.166}} & \textcolor{blue}{\underline{ 0.072}} & \textcolor{red}{ \textbf{0.161}} & \textcolor{red}{ \textbf{0.069}} \\
 & 192 & 0.253 & 0.098 & 0.237 & 0.091 & \textcolor{blue}{\underline{ 0.198}} & \textcolor{blue}{\underline{ 0.073}} & \textcolor{red}{ \textbf{0.190}} & \textcolor{red}{ \textbf{0.072}} \\
 & 336 & 0.309 & 0.112 & 0.288 & 0.104 & \textcolor{blue}{\underline{ 0.233}} & \textcolor{blue}{\underline{ 0.085}} & \textcolor{red}{ \textbf{0.227}} & \textcolor{red}{ \textbf{0.081}} \\
 & 720 & 0.397 & 0.138 & 0.374 & 0.112 & \textcolor{blue}{\underline{ 0.292}} & \textcolor{blue}{\underline{ 0.105}} & \textcolor{red}{ \textbf{0.289}} & \textcolor{red}{ \textbf{0.100}} \\
   \cmidrule{2-10}
\multirow{-5}{*}{\rotatebox{90}{Electricity}} & avg & 0.293 & 0.109 & 0.275 & 0.096 & \textcolor{blue}{\underline{ 0.222}} & \textcolor{blue}{\underline{ 0.084}} & \textcolor{red}{ \textbf{0.217}} & \textcolor{red}{ \textbf{0.081}} \\
\midrule
 & 96 & 0.599 & 0.239 & 0.557 & 0.229 & \textcolor{blue}{\underline{ 0.503}} & \textcolor{blue}{\underline{ 0.222}} & \textcolor{red}{ \textbf{0.480}} & \textcolor{red}{ \textbf{0.220}} \\
 & 192 & 0.631 & 0.275 & 0.592 & 0.251 & \textcolor{blue}{\underline{ 0.526}} & \textcolor{blue}{\underline{ 0.246}} & \textcolor{red}{ \textbf{0.511}} & \textcolor{red}{ \textbf{0.238}} \\
 & 336 & 0.722 & 0.301 & 0.647 & 0.282 & \textcolor{blue}{\underline{ 0.589}} & \textcolor{blue}{\underline{ 0.265}} & \textcolor{red}{ \textbf{0.553}} & \textcolor{red}{ \textbf{0.254}} \\
 & 720 & 0.808 & 0.318 & 0.712 & 0.294 & \textcolor{blue}{\underline{ 0.647}} & \textcolor{blue}{\underline{ 0.279}} & \textcolor{red}{ \textbf{0.639}} & \textcolor{red}{ \textbf{0.276}} \\
   \cmidrule{2-10}
\multirow{-5}{*}{\rotatebox{90}{Traffic}} & avg & 0.690 & 0.283 & 0.627 & 0.264 & \textcolor{blue}{\underline{ 0.566}} & \textcolor{blue}{\underline{ 0.253}} & \textcolor{red}{ \textbf{0.546}} & \textcolor{red}{ \textbf{0.247}} \\
\bottomrule
\end{tabular}}
\end{table}

\subsection{Analysis of SSD-Decoder}

Subsequently, we compare the SSD-Decoder with several representative decoder variants in Table~\ref{tab: ssd decoder}, including a Self-Attention Decoder, a Causal-Attention Decoder, and a Mamba v1 Decoder.
The main objective of these variants is to validate whether LegS provides a more suitable mechanism for the decoding stage, which is expected to support long-term forecasting while maintaining computational efficiency.
For fair comparison, we adopt the standard implementations of these commonly used structures and keep the hidden dimension consistent with that of the SSD-Decoder.

The results show that the SSD-Decoder achieves the best overall performance.
In forecasting tasks, the decoding process should respect temporal causality and effectively utilize historical memory for future prediction.
Therefore, the Self-Attention Decoder, which does not explicitly impose causal constraints, yields the weakest performance.
Compared with the Causal-Attention Decoder and the Mamba v1 Decoder, the SSD-Decoder further improves forecasting accuracy, demonstrating its stronger capability in adaptive memory utilization and long-range temporal modeling.
These results suggest that the LegS-based SSD mechanism is better suited for the decoder design in FLAME.

For transparency, we report the key hyperparameters of each decoder variant in Table~\ref{tab:ssd_decoder_hyperparams}. 
All variants use the same hidden dimension and number of decoder layers to ensure a fair comparison.
\begin{table}[!htbp]
\centering
 \captionsetup{width=0.9\linewidth}
\caption{Forecasting performance comparison of different decoders: Self-Attention (SA), Causal-Attention (CA), Mamba v1, and our SSD-Decoder used in FLAME.}
\label{tab: ssd decoder}
\resizebox{0.7\linewidth}{!}{
\begin{tabular}{cc|cc|cc|cc|cc}
\toprule
\multicolumn{2}{c|}{Models} & \multicolumn{2}{c|}{SA} & \multicolumn{2}{c|}{CA} & \multicolumn{2}{c|}{Mamba v1} & \multicolumn{2}{c}{FLAME} \\
\multicolumn{2}{c|}{Metric} & MSE & CRPS & MSE & CRPS & MSE & CRPS & MSE & CRPS \\
\midrule
 & 96 & 0.389 & 0.448 & 0.377 & 0.358 & \textcolor{blue}{\underline{ 0.337}} & \textcolor{blue}{\underline{ 0.332}} & \textcolor{red}{ \textbf{0.301}} & \textcolor{red}{ \textbf{0.315}} \\
 & 192 & 0.421 & 0.467 & 0.395 & 0.382 & \textcolor{blue}{\underline{ 0.359}} & \textcolor{blue}{\underline{ 0.341}} & \textcolor{red}{ \textbf{0.328}} & \textcolor{red}{ \textbf{0.337}} \\
 & 336 & 0.449 & 0.485 & 0.413 & 0.401 & \textcolor{blue}{\underline{ 0.390}} & \textcolor{blue}{\underline{ 0.367}} & \textcolor{red}{ \textbf{0.351}} & \textcolor{red}{ \textbf{0.351}} \\
 & 720 & 0.491 & 0.508 & 0.461 & 0.442 & \textcolor{blue}{\underline{ 0.444}} & \textcolor{blue}{\underline{ 0.390}} & \textcolor{red}{ \textbf{0.392}} & \textcolor{red}{ \textbf{0.383}} \\
   \cmidrule{2-10}
\multirow{-5}{*}{\rotatebox{90}{ETTm1}} & avg & 0.438 & 0.477 & 0.412 & 0.396 & \textcolor{blue}{\underline{ 0.383}} & \textcolor{blue}{\underline{ 0.358}} & \textcolor{red}{ \textbf{0.343}} & \textcolor{red}{ \textbf{0.347}} \\
\midrule
 & 96 & 0.235 & 0.167 & \textcolor{blue}{\underline{ 0.209}} & 0.141 & 0.212 & \textcolor{blue}{\underline{ 0.139}} & \textcolor{red}{ \textbf{0.186}} & \textcolor{red}{ \textbf{0.124}} \\
 & 192 & 0.272 & 0.198 & \textcolor{blue}{\underline{ 0.251}} & 0.167 & 0.256 & \textcolor{blue}{\underline{ 0.162}} & \textcolor{red}{ \textbf{0.247}} & \textcolor{red}{ \textbf{0.143}} \\
 & 336 & 0.352 & 0.223 & 0.313 & 0.182 & \textcolor{red}{ \textbf{0.292}} & \textcolor{blue}{\underline{ 0.179}} & \textcolor{blue}{\underline{ 0.300}} & \textcolor{red}{ \textbf{0.158}} \\
 & 720 & 0.398 & 0.257 & 0.368 & 0.224 & \textcolor{red}{ \textbf{0.356}} & \textcolor{blue}{\underline{ 0.198}} & \textcolor{blue}{\underline{ 0.357}} & \textcolor{red}{ \textbf{0.184}} \\
   \cmidrule{2-10}
\multirow{-5}{*}{\rotatebox{90}{ETTm2}} & avg & 0.314 & 0.211 & 0.285 & 0.179 & \textcolor{blue}{\underline{ 0.279}} & \textcolor{blue}{\underline{ 0.170}} & \textcolor{red}{ \textbf{0.273}} & \textcolor{red}{ \textbf{0.152}} \\
\midrule
 & 96 & 0.358 & 0.269 & \textcolor{red}{ \textbf{0.354}} & \textcolor{blue}{\underline{ 0.263}} & 0.363 & 0.271 & \textcolor{blue}{\underline{ 0.358}} & \textcolor{red}{ \textbf{0.259}} \\
 & 192 & 0.404 & 0.302 & \textcolor{blue}{\underline{ 0.395}} & \textcolor{blue}{\underline{ 0.286}} & 0.404 & 0.303 & \textcolor{red}{ \textbf{0.392}} & \textcolor{red}{ \textbf{0.278}} \\
 & 336 & \textcolor{blue}{\underline{ 0.420}} & 0.331 & 0.423 & \textcolor{blue}{\underline{ 0.297}} & 0.428 & 0.325 & \textcolor{red}{ \textbf{0.411}} & \textcolor{red}{ \textbf{0.290}} \\
 & 720 & \textcolor{blue}{\underline{ 0.455}} & 0.352 & 0.475 & \textcolor{blue}{\underline{ 0.325}} & 0.468 & 0.344 & \textcolor{red}{ \textbf{0.444}} & \textcolor{red}{ \textbf{0.303}} \\
   \cmidrule{2-10}
\multirow{-5}{*}{\rotatebox{90}{ETTh1}} & avg & \textcolor{blue}{\underline{ 0.409}} & 0.314 & 0.412 & \textcolor{blue}{\underline{ 0.293}} & 0.416 & 0.311 & \textcolor{red}{ \textbf{0.401}} & \textcolor{red}{ \textbf{0.283}} \\
\midrule
 & 96 & 0.336 & 0.175 & \textcolor{red}{ \textbf{0.288}} & \textcolor{red}{ \textbf{0.139}} & 0.303 & 0.148 & \textcolor{blue}{\underline{ 0.300}} & \textcolor{blue}{\underline{ 0.144}} \\
 & 192 & 0.382 & 0.203 & \textcolor{blue}{\underline{ 0.362}} & \textcolor{blue}{\underline{ 0.170}} & 0.370 & 0.172 & \textcolor{red}{ \textbf{0.351}} & \textcolor{red}{ \textbf{0.159}} \\
 & 336 & 0.414 & 0.212 & \textcolor{blue}{\underline{ 0.389}} & \textcolor{blue}{\underline{ 0.177}} & 0.401 & 0.186 & \textcolor{red}{ \textbf{0.374}} & \textcolor{red}{ \textbf{0.172}} \\
 & 720 & 0.472 & 0.209 & 0.432 & \textcolor{blue}{\underline{ 0.198}} & \textcolor{red}{ \textbf{0.409}} & 0.202 & \textcolor{blue}{\underline{ 0.416}} & \textcolor{red}{ \textbf{0.182}} \\
   \cmidrule{2-10}
\multirow{-5}{*}{\rotatebox{90}{ETTh2}} & avg & 0.401 & 0.200 & \textcolor{blue}{\underline{ 0.368}} & \textcolor{blue}{\underline{ 0.171}} & 0.371 & 0.177 & \textcolor{red}{ \textbf{0.360}} & \textcolor{red}{ \textbf{0.164}} \\
\midrule
 & 96 & \textcolor{blue}{\underline{ 0.188}} & 0.088 & 0.209 & 0.093 & 0.207 & \textcolor{blue}{\underline{ 0.087}} & \textcolor{red}{ \textbf{0.152}} & \textcolor{red}{ \textbf{0.067}} \\
 & 192 & \textcolor{blue}{\underline{ 0.221}} & 0.094 & 0.237 & 0.097 & 0.237 & \textcolor{blue}{\underline{ 0.091}} & \textcolor{red}{ \textbf{0.199}} & \textcolor{red}{ \textbf{0.068}} \\
 & 336 & 0.275 & 0.113 & \textcolor{blue}{\underline{ 0.261}} & 0.105 & 0.269 & \textcolor{blue}{\underline{ 0.102}} & \textcolor{red}{ \textbf{0.250}} & \textcolor{red}{ \textbf{0.079}} \\
 & 720 & 0.328 & 0.127 & 0.311 & 0.117 & \textcolor{blue}{\underline{ 0.311}} & \textcolor{blue}{\underline{ 0.112}} & \textcolor{red}{ \textbf{0.307}} & \textcolor{red}{ \textbf{0.083}} \\
   \cmidrule{2-10}
\multirow{-5}{*}{\rotatebox{90}{Weather}} & avg & \textcolor{blue}{\underline{ 0.253}} & 0.106 & 0.255 & 0.103 & 0.256 & \textcolor{blue}{\underline{ 0.098}} & \textcolor{red}{ \textbf{0.227}} & \textcolor{red}{ \textbf{0.074}} \\
\midrule
 & 96 & 0.210 & 0.115 & 0.172 & 0.075 & \textcolor{blue}{\underline{ 0.163}} & \textcolor{blue}{\underline{ 0.072}} & \textcolor{red}{ \textbf{0.161}} & \textcolor{red}{ \textbf{0.069}} \\
 & 192 & 0.255 & 0.127 & 0.204 & 0.084 & \textcolor{blue}{\underline{ 0.191}} & \textcolor{blue}{\underline{ 0.077}} & \textcolor{red}{ \textbf{0.190}} & \textcolor{red}{ \textbf{0.072}} \\
 & 336 & 0.318 & 0.132 & 0.243 & 0.098 & \textcolor{blue}{\underline{ 0.234}} & \textcolor{blue}{\underline{ 0.090}} & \textcolor{red}{ \textbf{0.227}} & \textcolor{red}{ \textbf{0.081}} \\
 & 720 & 0.415 & 0.149 & 0.302 & 0.122 & \textcolor{red}{ \textbf{0.285}} & \textcolor{blue}{\underline{ 0.108}} & \textcolor{blue}{\underline{ 0.289}} & \textcolor{red}{ \textbf{0.100}} \\
   \cmidrule{2-10}
\multirow{-5}{*}{\rotatebox{90}{Electricity}} & avg & 0.300 & 0.131 & 0.230 & 0.095 & \textcolor{blue}{\underline{ 0.218}} & \textcolor{blue}{\underline{ 0.087}} & \textcolor{red}{ \textbf{0.217}} & \textcolor{red}{ \textbf{0.081}} \\
\midrule
 & 96 & 0.587 & 0.311 & \textcolor{blue}{\underline{ 0.492}} & \textcolor{blue}{\underline{ 0.238}} & 0.501 & 0.244 & \textcolor{red}{ \textbf{0.480}} & \textcolor{red}{ \textbf{0.220}} \\
 & 192 & 0.617 & 0.338 & \textcolor{blue}{\underline{ 0.521}} & \textcolor{blue}{\underline{ 0.256}} & 0.522 & 0.270 & \textcolor{red}{ \textbf{0.511}} & \textcolor{red}{ \textbf{0.238}} \\
 & 336 & 0.662 & 0.343 & 0.572 & \textcolor{blue}{\underline{ 0.275}} & \textcolor{blue}{\underline{ 0.562}} & 0.294 & \textcolor{red}{ \textbf{0.553}} & \textcolor{red}{ \textbf{0.254}} \\
 & 720 & 0.749 & 0.372 & 0.655 & \textcolor{blue}{\underline{ 0.294}} & \textcolor{blue}{\underline{ 0.642}} & 0.298 & \textcolor{red}{ \textbf{0.639}} & \textcolor{red}{ \textbf{0.276}} \\
   \cmidrule{2-10}
\multirow{-5}{*}{\rotatebox{90}{Traffic}} & avg & 0.654 & 0.341 & 0.560 & \textcolor{blue}{\underline{ 0.266}} & \textcolor{blue}{\underline{ 0.557}} & 0.277 & \textcolor{red}{ \textbf{0.546}} & \textcolor{red}{ \textbf{0.247}}\\
\bottomrule
\end{tabular}
}
\end{table}

\begin{table*}[!htbp]
\centering
\small
\caption{Key hyperparameters of different decoder variants. All variants use the same hidden dimension and number of decoder layers for fair comparison.}
\label{tab:ssd_decoder_hyperparams}

\begin{minipage}[t]{0.24\textwidth}
\centering
\resizebox{\linewidth}{!}{
\begin{tabular}{lc}
\toprule
\multicolumn{2}{c}{Self-Attention} \\
\midrule
Hyperparameter & Value \\
\midrule
Decoder Type & Self-Attention \\
Layers & 6 \\
Hidden Dim. & 256 \\
FFN Dim. & 512 \\
Num Heads & 8 \\
Head Dim. & 32 \\
Dropout & 0.1 \\
Causal Mask & No \\
Pos. Encoding & Learnable \\
Norm. & LayerNorm \\
\bottomrule
\end{tabular}}
\end{minipage}
\hfill
\begin{minipage}[t]{0.24\textwidth}
\centering
\resizebox{\linewidth}{!}{
\begin{tabular}{lc}
\toprule
\multicolumn{2}{c}{Causal-Attention} \\
\midrule
Hyperparameter & Value \\
\midrule
Decoder Type & Causal-Attention \\
Layers & 6 \\
Hidden Dim. & 256 \\
FFN Dim. & 512 \\
Num Heads & 8 \\
Head Dim. & 32 \\
Dropout & 0.1 \\
Causal Mask & Yes \\
Pos. Encoding & Learnable \\
Norm. & LayerNorm \\
\bottomrule
\end{tabular}}
\end{minipage}
\hfill
\begin{minipage}[t]{0.24\textwidth}
\centering
\resizebox{\linewidth}{!}{
\begin{tabular}{lc}
\toprule
\multicolumn{2}{c}{Mamba v1} \\
\midrule
Hyperparameter & Value \\
\midrule
Decoder Type & Mamba v1 \\
Layers & 6 \\
Hidden Dim. & 256 \\
State Dim. & 16 \\
Conv. Kernel & 4 \\
Expansion & 2 \\
Num Heads & -- \\
Head Dim. & -- \\
Dropout & 0.1 \\
Causal Modeling & Yes \\
Norm. & RMSNorm \\
\bottomrule
\end{tabular}}
\end{minipage}
\hfill
\begin{minipage}[t]{0.24\textwidth}
\centering
\resizebox{\linewidth}{!}{
\begin{tabular}{lc}
\toprule
\multicolumn{2}{c}{SSD-Decoder} \\
\midrule
Hyperparameter & Value \\
\midrule
Decoder Type & Mamba v2 / SSD \\
Layers & 6 \\
Hidden Dim. & 256 \\
State Dim. & 64 \\
Conv. Kernel & 4 \\
Expansion & 2 \\
SSM Heads & 8 \\
Head Dim. & 32 \\
LegS Order & 256 \\
Dropout & 0.1 \\
Causal Modeling & Yes \\
Memory & LegS-based SSD \\
Norm. & RMSNorm \\
\bottomrule
\end{tabular}}
\end{minipage}

\end{table*}

\subsection{Analysis of Flow-based Head}

To assess the efficacy of the proposed Flow-based Head, we compare it with several representative generative forecasting heads in Table~\ref{tab: flow head}, including Variational Autoencoder (VAE)~\citep{bVAE}, Diffusion~\citep{ho2020denoisingdiffusionprobabilisticmodels}, and Flow Matching (FM)~\citep{FlowMatch}.
The main objective of this ablation is to examine whether the proposed normalizing-flow-based design is more suitable for probabilistic forecasting than other commonly used generative mechanisms, under comparable conditioning and hidden-size settings.

For a fair and informative comparison, we implement all alternative heads with concise but standard designs.
Specifically, the VAE baseline is implemented as a conditional VAE (CVAE) decoder~\citep{bVAE}, where the decoder receives the forecasting condition token from the backbone and generates the target patch from the latent variable, and it adopts the MLPs as network structures.
For Diffusion~\citep{ho2020denoisingdiffusionprobabilisticmodels} and Flow Matching~\citep{FlowMatch}, we adopt a DiT-style conditional architecture~\citep{DBLP:conf/iccv/PeeblesX23}, where the condition is injected through AdaLN-Zero, while the Transformer blocks are replaced by lightweight MLP blocks to keep the parameter scale comparable to our forecasting head.
For the Diffusion baseline, we use a DDPM-style probability path.
For the Flow Matching baseline, we adopt the optimal-transport path used in Sundial~\citep{liu2025sundial} and Aurora~\citep{wu2026aurora}, which is a commonly used probability path for learning continuous transformations between the base and data distributions.
These settings ensure that the baselines follow standard generative modeling principles while remaining lightweight enough to serve as forecasting heads. Note that the VAE, Flow Matching, and Diffusion all adopt the same hidden dimension $d=256$ as the FLAME's Flow-based Head, all other parameters follow the standard design and maintain an appropriate multiple relationship with the hidden dimension.

In the proposed Flow-based Head, samples are independently drawn from the base Gaussian distribution and then transformed through a sequence of autoregressive affine layers to obtain samples from the target forecasting distribution.
For probabilistic forecasting, we randomly sample 100 times from the base distribution.
The deterministic prediction is then obtained by averaging the 100 generated samples, which provides an empirical estimate of the predictive mean.

The results show that the proposed Flow-based Head achieves the best overall performance.
The VAE baseline follows a one-step generation paradigm, which makes it less effective on datasets with complex real-world temporal dynamics, such as Electricity and Traffic.
Although Diffusion and Flow Matching provide more expressive multi-step generation mechanisms, they introduce additional sampling dynamics and rely heavily on the choice of probability paths, solvers, and velocity or denoising networks.
In contrast, our Flow-based Head directly learns an invertible transformation from the base distribution to the target forecasting distribution through lightweight autoregressive affine layers.
This design supports efficient probabilistic sampling while preserving fine-grained dependencies within each predicted patch.
These results suggest that the normalizing-flow-based head provides a more suitable balance between generative expressiveness, inference efficiency, and forecasting accuracy for FLAME.

To further clarify the implementation details of the generative forecasting heads, we report their key hyperparameters in Table~\ref{tab:flow_head_hyperparams}. 
For fair comparison, all variants adopt the same hidden dimension $d=256$, the same forecasting condition token, and the same number of network layers. 
The main differences lie in their generative mechanisms, conditioning strategies, and sampling procedures.

\begin{table*}[!htbp]
\centering
\small
\caption{Key hyperparameters of different generative forecasting heads. All variants use the same hidden dimension $d=256$, the same forecasting condition token, and the same number of layers for fair comparison.}
\label{tab:flow_head_hyperparams}

\begin{minipage}[t]{0.24\textwidth}
\centering
\resizebox{\linewidth}{!}{
\begin{tabular}{lc}
\toprule
\multicolumn{2}{c}{VAE Head} \\
\midrule
Hyperparameter & Value \\
\midrule
Generative Type & CVAE \\
Layers & 7 \\
Hidden Dim. & 256 \\
Latent Dim. & 128 \\
Encoder Network & MLP \\
Decoder Network & MLP \\
Condition Injection & Concatenation \\
Prior Distribution & Gaussian \\
Sampling Steps & 1 \\
Training Objective & ELBO \\
Output Samples & 100 \\
\bottomrule
\end{tabular}}
\end{minipage}
\hfill
\begin{minipage}[t]{0.24\textwidth}
\centering
\resizebox{\linewidth}{!}{
\begin{tabular}{lc}
\toprule
\multicolumn{2}{c}{Diffusion Head} \\
\midrule
Hyperparameter & Value \\
\midrule
Generative Type & DDPM \\
Layers & 7 \\
Hidden Dim. & 256 \\
Network Type & MLP Blocks \\
Condition Injection & AdaLN-Zero \\
Probability Path & DDPM-style \\
Noise Schedule & Linear \\
Prediction Target & Noise \\
Sampling Steps & 100 \\
Training Objective & Denoising Loss \\
Output Samples & 100 \\
\bottomrule
\end{tabular}}
\end{minipage}
\hfill
\begin{minipage}[t]{0.24\textwidth}
\centering
\resizebox{\linewidth}{!}{
\begin{tabular}{lc}
\toprule
\multicolumn{2}{c}{Flow Matching Head} \\
\midrule
Hyperparameter & Value \\
\midrule
Generative Type & Flow Matching \\
Layers & 7 \\
Hidden Dim. & 256 \\
Network Type & MLP Blocks \\
Condition Injection & AdaLN-Zero \\
Probability Path & OT Path \\
Solver & Euler \\
Prediction Target & Velocity \\
Sampling Steps & 100 \\
Training Objective & FM Loss \\
Output Samples & 100 \\
\bottomrule
\end{tabular}}
\end{minipage}
\hfill
\begin{minipage}[t]{0.24\textwidth}
\centering
\resizebox{\linewidth}{!}{
\begin{tabular}{lc}
\toprule
\multicolumn{2}{c}{Flow-based Head} \\
\midrule
Hyperparameter & Value \\
\midrule
Generative Type & Normalizing Flow \\
Layers & 7 \\
Hidden Dim. & 256 \\
Base Distribution & Gaussian \\
Flow Transform & Autoregressive Affine \\
Condition Injection & Masked MLP \\
Jacobian Type & Triangular \\
Sampling Steps & $p$ per flow layer \\
Training Objective & NLL \\
Output Samples & 100 \\
\bottomrule
\end{tabular}}
\end{minipage}

\end{table*}

\begin{table}[!htbp]
\centering
 \captionsetup{width=0.7\linewidth}
\caption{Forecasting performance comparison of various generative heads: VAE, Diffusion, Flow Matching (FM), and our Flow-based Head used in FLAME. We also compare FLAME with vanilla Gaussian Head.}
\label{tab: flow head}
\resizebox{0.55\linewidth}{!}{
\begin{tabular}{cc|cc|cc|cc|cc|cc}
\toprule
\multicolumn{2}{c|}{Models} & \multicolumn{2}{c|}{Gaussian}& \multicolumn{2}{c|}{VAE} & \multicolumn{2}{c|}{Diffusion} & \multicolumn{2}{c|}{FM} & \multicolumn{2}{c}{FLAME} \\
\multicolumn{2}{c|}{Metric}  & MSE & CRPS & MSE & CRPS & MSE & CRPS & MSE & CRPS & MSE & CRPS \\
\midrule
 & 96 & 0.342& 0.357  & 0.334 & 0.348 & \textcolor{blue}{\underline{ 0.305}} & \textcolor{blue}{\underline{ 0.318}} & 0.311 & 0.323 & \textcolor{red}{ \textbf{0.301}} & \textcolor{red}{ \textbf{0.315}} \\
 & 192 &0.362 &0.379 & 0.364 & 0.382 & 0.340 & 0.345 & \textcolor{blue}{\underline{ 0.331}} & \textcolor{blue}{\underline{ 0.340}} & \textcolor{red}{ \textbf{0.328}} & \textcolor{red}{ \textbf{0.337}} \\
 & 336 &0.395 &0.413 & 0.388 & 0.397 & \textcolor{blue}{\underline{ 0.359}} & 0.362 & 0.361 & \textcolor{blue}{\underline{ 0.360}} & \textcolor{red}{ \textbf{0.351}} & \textcolor{red}{ \textbf{0.351}} \\
 & 720 &0.427 & 0.463 & 0.421 & 0.448 & 0.401 & 0.390 & \textcolor{blue}{\underline{ 0.398}} & \textcolor{blue}{\underline{ 0.386}} & \textcolor{red}{ \textbf{0.392}} & \textcolor{red}{ \textbf{0.383}} \\
   \cmidrule{2-12}
\multirow{-5}{*}{\rotatebox{90}{ETTm1}} & avg & 0.382 &0.403 &0.377 & 0.394 & 0.351 & 0.354 & \textcolor{blue}{\underline{ 0.350}} & \textcolor{blue}{\underline{ 0.352}} & \textcolor{red}{ \textbf{0.343}} & \textcolor{red}{ \textbf{0.347}} \\
\midrule
 & 96 &0.231&0.166 & 0.223 & 0.152 & \textcolor{blue}{\underline{ 0.194}} & \textcolor{blue}{\underline{ 0.128}} & 0.215 & 0.139 & \textcolor{red}{ \textbf{0.186}} & \textcolor{red}{ \textbf{0.124}} \\
 & 192 &0.312 &0.198 & 0.292 & 0.179 & 0.261 & 0.145 & \textcolor{blue}{\underline{ 0.257}} & \textcolor{blue}{\underline{ 0.144}} & \textcolor{red}{ \textbf{0.247}} & \textcolor{red}{ \textbf{0.143}} \\
 & 336 &0.372 &0.225 & 0.345 & 0.197 & 0.322 & \textcolor{blue}{\underline{ 0.163}} & \textcolor{blue}{\underline{ 0.309}} & 0.168 & \textcolor{red}{ \textbf{0.300}} & \textcolor{red}{ \textbf{0.158}} \\
 & 720 &0.413 &0.231 & 0.392 & 0.215 & \textcolor{blue}{\underline{ 0.370}} & \textcolor{blue}{\underline{ 0.200}} & 0.381 & 0.205 & \textcolor{red}{ \textbf{0.357}} & \textcolor{red}{ \textbf{0.184}} \\
   \cmidrule{2-12}
\multirow{-5}{*}{\rotatebox{90}{ETTm2}} & avg & 0.332 & 0.205& 0.313 & 0.186 & \textcolor{blue}{\underline{ 0.287}} & \textcolor{blue}{\underline{ 0.159}} & 0.291 & 0.164 & \textcolor{red}{ \textbf{0.273}} & \textcolor{red}{ \textbf{0.152}} \\
\midrule
 & 96 &0.374&0.291 & 0.366 & 0.279 & \textcolor{blue}{\underline{ 0.359}} & \textcolor{blue}{\underline{ 0.261}} & 0.363 & 0.267 & \textcolor{red}{ \textbf{0.358}} & \textcolor{red}{ \textbf{0.259}} \\
 & 192 &0.408 & 0.304 & 0.398 & \textcolor{blue}{\underline{ 0.285}} & \textcolor{blue}{\underline{ 0.394}} & 0.288 & 0.404 & 0.295 & \textcolor{red}{ \textbf{0.392}} & \textcolor{red}{ \textbf{0.278}} \\
 & 336 &0.441&0.344 & \textcolor{blue}{\underline{ 0.423}} & \textcolor{blue}{\underline{ 0.301}} & 0.428 & 0.318 & 0.427 & 0.330 & \textcolor{red}{ \textbf{0.411}} & \textcolor{red}{ \textbf{0.290}} \\
 & 720 &0.488&0.372 & \textcolor{blue}{\underline{ 0.457}} & \textcolor{blue}{\underline{ 0.320}} & 0.479 & 0.348 & 0.462 & 0.327 & \textcolor{red}{ \textbf{0.444}} & \textcolor{red}{ \textbf{0.303}} \\
   \cmidrule{2-12}
\multirow{-5}{*}{\rotatebox{90}{ETTh1}} & avg & 0.420 & 0.328 & \textcolor{blue}{\underline{ 0.411}} & \textcolor{blue}{\underline{ 0.296}} & 0.415 & 0.304 & 0.414 & 0.305 & \textcolor{red}{ \textbf{0.401}} & \textcolor{red}{ \textbf{0.283}} \\
\midrule
 & 96 &0.354 &0.193 & \textcolor{red}{ \textbf{0.296}} & \textcolor{red}{ \textbf{0.142}} & 0.332 & 0.169 & 0.328 & 0.164 & \textcolor{blue}{\underline{ 0.300}} & \textcolor{blue}{\underline{ 0.144}} \\
 & 192 &0.395&0.206 & \textcolor{blue}{\underline{ 0.358}} & \textcolor{blue}{\underline{ 0.167}} & 0.378 & 0.180 & 0.364 & 0.168 & \textcolor{red}{ \textbf{0.351}} & \textcolor{red}{ \textbf{0.159}} \\
 & 336 &0.405&0.217 & \textcolor{blue}{\underline{ 0.379}} & \textcolor{blue}{\underline{ 0.182}} & 0.394 & 0.198 & 0.384 & 0.191 & \textcolor{red}{ \textbf{0.374}} & \textcolor{red}{ \textbf{0.172}} \\
 & 720 &0.472&0.246 & \textcolor{blue}{\underline{ 0.425}} & \textcolor{blue}{\underline{ 0.197}} & 0.448 & 0.215 & 0.428 & 0.202 & \textcolor{red}{ \textbf{0.416}} & \textcolor{red}{ \textbf{0.182}} \\
   \cmidrule{2-12}
\multirow{-5}{*}{\rotatebox{90}{ETTh2}} & avg &0.407 &0.216 &\textcolor{blue}{\underline{ 0.365}} & \textcolor{blue}{\underline{ 0.172}} & 0.388 & 0.191 & 0.376 & 0.181 & \textcolor{red}{ \textbf{0.360}} & \textcolor{red}{ \textbf{0.164}} \\
\midrule
 & 96 &0.186 & 0.097 & 0.176 & 0.083 & \textcolor{red}{ \textbf{0.149}} & \textcolor{blue}{\underline{ 0.068}} & 0.155 & 0.072 & \textcolor{blue}{\underline{ 0.152}} & \textcolor{red}{ \textbf{0.067}} \\
 & 192 &0.237 &0.098 & 0.229 & 0.086 & \textcolor{blue}{\underline{ 0.199}} & \textcolor{blue}{\underline{ 0.069}} & 0.206 & 0.074 & \textcolor{red}{ \textbf{0.199}} & \textcolor{red}{ \textbf{0.068}} \\
 & 336 &0.317 &0.114 & 0.292 & 0.098 & \textcolor{blue}{\underline{ 0.256}} & \textcolor{blue}{\underline{ 0.083}} & 0.257 & 0.084 & \textcolor{red}{ \textbf{0.250}} & \textcolor{red}{ \textbf{0.079}} \\
 & 720 &0.369 &0.118 & 0.357 & 0.105 & 0.314 & 0.089 & \textcolor{blue}{\underline{ 0.312}} & \textcolor{blue}{\underline{ 0.088}} & \textcolor{red}{ \textbf{0.307}} & \textcolor{red}{ \textbf{0.083}} \\
   \cmidrule{2-12}
\multirow{-5}{*}{\rotatebox{90}{Weather}} & avg &0.277&0.107 & 0.264 & 0.093 & \textcolor{blue}{\underline{ 0.230}} & \textcolor{blue}{\underline{ 0.077}} & 0.233 & 0.080 & \textcolor{red}{ \textbf{0.227}} & \textcolor{red}{ \textbf{0.074}} \\
\midrule
 & 96 & 0.255&0.141  & 0.241 & 0.129 & 0.172 & 0.078 & \textcolor{blue}{\underline{ 0.167}} & \textcolor{blue}{\underline{ 0.075}} & \textcolor{red}{ \textbf{0.161}} & \textcolor{red}{ \textbf{0.069}} \\
 & 192 &0.263&0.144 & 0.260 & 0.134 & 0.199 &  0.079 & \textcolor{blue}{\underline{ 0.195}} & \textcolor{blue}{\underline{ 0.079}} & \textcolor{red}{ \textbf{0.190}} & \textcolor{red}{ \textbf{0.072}} \\
 & 336 &0.312 &0.148 & 0.304 & 0.142 & 0.251 & 0.088 & \textcolor{blue}{\underline{ 0.246}} & \textcolor{blue}{\underline{ 0.086}} & \textcolor{red}{ \textbf{0.227}} & \textcolor{red}{ \textbf{0.081}} \\
 & 720 &0.367& 0.155 & 0.357 & 0.147 & 0.318 & \textcolor{blue}{\underline{ 0.105}} & \textcolor{blue}{\underline{ 0.318}} & 0.110 & \textcolor{red}{ \textbf{0.289}} & \textcolor{red}{ \textbf{0.100}} \\
   \cmidrule{2-12}
\multirow{-5}{*}{\rotatebox{90}{Electricity}} & avg &0.299&0.147 & 0.291 & 0.138 & 0.235 & 0.088 & \textcolor{blue}{\underline{ 0.232}} & \textcolor{blue}{\underline{ 0.088}} & \textcolor{red}{ \textbf{0.217}} & \textcolor{red}{ \textbf{0.081}} \\
\midrule
 & 96 &0.785&0.533 & 0.721 & 0.482 & \textcolor{blue}{\underline{ 0.488}} & \textcolor{blue}{\underline{ 0.236}} & 0.507 & 0.248 & \textcolor{red}{ \textbf{0.480}} & \textcolor{red}{ \textbf{0.220}} \\
 & 192 &0.796&0.542 & 0.742 & 0.498 & \textcolor{blue}{\underline{ 0.528}} & \textcolor{blue}{\underline{ 0.264}} & 0.557 & 0.296 & \textcolor{red}{ \textbf{0.511}} & \textcolor{red}{ \textbf{0.238}} \\
 & 336 &0.817&0.579  & 0.747 & 0.512 & \textcolor{blue}{\underline{ 0.572}} & \textcolor{blue}{\underline{ 0.270}} & 0.603 & 0.305 & \textcolor{red}{ \textbf{0.553}} & \textcolor{red}{ \textbf{0.254}} \\
 & 720 &0.838&0.601  & 0.769 & 0.532 & \textcolor{blue}{\underline{ 0.682}} & \textcolor{blue}{\underline{ 0.293}} & 0.712 & 0.337 & \textcolor{red}{ \textbf{0.639}} & \textcolor{red}{ \textbf{0.276}} \\
   \cmidrule{2-12}
\multirow{-5}{*}{\rotatebox{90}{Traffic}} & avg &0.809 &0.564  & 0.745 & 0.506 & \textcolor{blue}{\underline{ 0.568}} & \textcolor{blue}{\underline{ 0.266}} & 0.595 & 0.297 & \textcolor{red}{ \textbf{0.546}} & \textcolor{red}{ \textbf{0.247}}\\
\bottomrule
\end{tabular}
}
\end{table}

\section{Full Results}
\label{app: full results}

\subsection{Full zero-shot results}
We provide all the results of the zero-shot deterministic forecasting in Table~\ref{tab: main point forecasting all}. As shown in Table~\ref{tab: main point forecasting all}, we include 11 representative real-world datasets from TSFM-Bench, demonstrating that FLAME achieves state-of-the-art forecasting performance.
\renewcommand{\arraystretch}{1.2}
\begin{table*}[!htbp]
\centering
\caption{Full results of zero-shot deterministic forecasting experiments on datasets from TSFM-Bench. Lower MSE or MAE values indicate better predictions. (``-'') denotes datasets included in the model’s pretraining and therefore excluded from testing. \textcolor{red}{\textbf{Red}}: the best, \textcolor{blue}{\underline{Blue}}: the 2nd best.}
\label{tab: main point forecasting all}
\resizebox{\textwidth}{!}{
\begin{tabular}{cc|cc|cc|cc|cc|cc|cc|cc|cc|cc|cc|cc}
\toprule
\multicolumn{2}{c|}{\multirow{2}{*}{Models}} & \multicolumn{2}{c|}{\textbf{FLAME}} & \multicolumn{2}{c|}{\textbf{FLAME}} & \multicolumn{2}{c|}{\textbf{FLAME}} & \multicolumn{2}{c|}{Sundial} & \multicolumn{2}{c|}{LightGTS} & \multicolumn{2}{c|}{ROSE} & \multicolumn{2}{c|}{Timer} & \multicolumn{2}{c|}{Chronos} & \multicolumn{2}{c|}{Time-MoE} & \multicolumn{2}{c|}{TinyTTM} & \multicolumn{2}{c}{MOIRAI} \\
~ & ~ & \multicolumn{2}{c|}{\textbf{Small}} & \multicolumn{2}{c|}{\textbf{Base}} & \multicolumn{2}{c|}{\textbf{Large}} & \multicolumn{2}{c|}{(2025)} & \multicolumn{2}{c|}{(2025)} & \multicolumn{2}{c|}{(2025)} & \multicolumn{2}{c|}{(2024)} & \multicolumn{2}{c|}{(2024)} & \multicolumn{2}{c|}{(2024)} & \multicolumn{2}{c|}{(2024)} & \multicolumn{2}{c}{(2024)} \\ \midrule
\multicolumn{2}{c|}{Metric} & \multicolumn{1}{c}{MSE} & \multicolumn{1}{c|}{MAE} & \multicolumn{1}{c}{MSE} & \multicolumn{1}{c|}{MAE} & \multicolumn{1}{c}{MSE} & \multicolumn{1}{c|}{MAE} & \multicolumn{1}{c}{MSE} & \multicolumn{1}{c|}{MAE} & \multicolumn{1}{c}{MSE} & \multicolumn{1}{c}{MAE} & \multicolumn{1}{c}{MSE} & \multicolumn{1}{c|}{MAE} & \multicolumn{1}{c}{MSE} & \multicolumn{1}{c|}{MAE} & \multicolumn{1}{c}{MSE} & \multicolumn{1}{c|}{MAE} & \multicolumn{1}{c}{MSE} & \multicolumn{1}{c|}{MAE} & \multicolumn{1}{c}{MSE} & \multicolumn{1}{c|}{MAE} & \multicolumn{1}{c}{MSE} & \multicolumn{1}{c}{MAE} \\
\midrule
\multirow[c]{5}{*}{\rotatebox{90}{ETTm1}} & 96 & 0.305 & 0.353 & 0.298 & 0.351 & 0.301 & 0.353 & \textcolor{red}{\textbf{0.280}} & \textcolor{red}{\textbf{0.334}} & 0.307 & 0.359 & 0.512 & 0.460 & 0.817 & 0.611 & 0.402 & 0.373 & \textcolor{blue}{\underline{0.281}} & \textcolor{blue}{\underline{0.341}} & 0.738 & 0.541 & 0.353 & 0.363 \\
 & 192 & 0.328 & 0.370 & 0.324 & 0.369 & 0.328 & 0.373 & \textcolor{blue}{\underline{0.321}} & \textcolor{blue}{\underline{0.366}} & 0.332 & 0.374 & 0.512 & 0.462 & 0.927 & 0.659 & 0.510 & 0.435 & \textcolor{red}{\textbf{0.305}} & \textcolor{red}{\textbf{0.358}} & 0.698 & 0.547 & 0.376 & 0.380 \\
 & 336 & \textcolor{blue}{\underline{0.349}} & \textcolor{red}{\textbf{0.384}} & \textcolor{red}{\textbf{0.346}} & \textcolor{blue}{\underline{0.386}} & 0.351 & 0.389 & 0.350 & 0.389 & 0.353 & 0.393 & 0.523 & 0.470 & 1.043 & 0.704 & 0.590 & 0.477 & 0.369 & 0.395 & 0.670 & 0.533 & 0.399 & 0.395 \\
 & 720 & 0.392 & 0.416 & \textcolor{red}{\textbf{0.385}} & \textcolor{blue}{\underline{0.414}} & 0.392 & 0.419 & 0.394 & 0.418 & \textcolor{blue}{\underline{0.388}} & \textcolor{red}{\textbf{0.386}} & 0.552 & 0.490 & 1.044 & 0.722 & 0.703 & 0.525 & 0.469 & 0.472 & 0.660 & 0.550 & 0.432 & 0.417 \\ \cmidrule{2-24}
 & Avg & 0.344 & 0.381 & \textcolor{blue}{\underline{0.338}} & 0.380 & 0.343 & 0.384 & \textcolor{red}{\textbf{0.336}} & \textcolor{red}{\textbf{0.377}} & 0.345 & \textcolor{blue}{\underline{0.378}} & 0.525 & 0.471 & 0.958 & 0.674 & 0.551 & 0.453 & 0.356 & 0.392 & 0.692 & 0.543 & 0.390 & 0.389 \\ \midrule
\multirow[c]{5}{*}{\rotatebox{90}{ETTm2}} & 96 & \textcolor{red}{\textbf{0.165}} & \textcolor{red}{\textbf{0.255}} & 0.171 & 0.262 & 0.186 & 0.272 & 0.170 & \textcolor{blue}{\underline{0.256}} & \textcolor{blue}{\underline{0.166}} & 0.260 & 0.224 & 0.309 & 0.225 & 0.300 & 0.192 & 0.263 & 0.198 & 0.288 & 0.226 & 0.309 & 0.189 & 0.260 \\
 & 192 & \textcolor{red}{\textbf{0.217}} & \textcolor{red}{\textbf{0.297}} & 0.227 & 0.302 & 0.247 & 0.315 & 0.229 & \textcolor{blue}{\underline{0.300}} & \textcolor{blue}{\underline{0.222}} & 0.300 & 0.266 & 0.333 & 0.286 & 0.339 & 0.256 & 0.308 & 0.235 & 0.312 & 0.311 & 0.360 & 0.247 & 0.300 \\
 & 336 & \textcolor{red}{\textbf{0.262}} & \textcolor{red}{\textbf{0.330}} & 0.280 & 0.340 & 0.300 & 0.352 & 0.281 & 0.337 & \textcolor{blue}{\underline{0.267}} & \textcolor{blue}{\underline{0.333}} & 0.310 & 0.358 & 0.335 & 0.369 & 0.315 & 0.346 & 0.293 & 0.348 & 0.350 & 0.383 & 0.295 & 0.334 \\
 & 720 & \textcolor{red}{\textbf{0.333}} & \textcolor{blue}{\underline{0.379}} & 0.348 & 0.387 & 0.357 & 0.394 & 0.351 & 0.387 & \textcolor{blue}{\underline{0.340}} & \textcolor{red}{\textbf{0.378}} & 0.395 & 0.407 & 0.414 & 0.416 & 0.409 & 0.405 & 0.427 & 0.428 & 0.446 & 0.435 & 0.372 & 0.386 \\ \cmidrule{2-24}
 & Avg & \textcolor{red}{\textbf{0.244}} & \textcolor{red}{\textbf{0.315}} & 0.257 & 0.323 & 0.273 & 0.333 & 0.258 & 0.320 & \textcolor{blue}{\underline{0.249}} & \textcolor{blue}{\underline{0.318}} & 0.299 & 0.352 & 0.315 & 0.356 & 0.293 & 0.331 & 0.288 & 0.344 & 0.333 & 0.372 & 0.276 & 0.320 \\ \midrule
\multirow[c]{5}{*}{\rotatebox{90}{ETTh1}} & 96 & 0.362 & 0.395 & 0.361 & 0.392 & 0.358 & 0.389 & \textcolor{red}{\textbf{0.348}} & \textcolor{blue}{\underline{0.385}} & 0.359 & 0.390 & 0.382 & 0.408 & 0.454 & 0.434 & 0.444 & 0.409 & \textcolor{blue}{\underline{0.349}} & \textcolor{red}{\textbf{0.379}} & 0.364 & 0.389 & 0.380 & 0.398 \\
 & 192 & 0.406 & 0.427 & 0.400 & 0.422 & 0.392 & 0.416 & 0.393 & 0.418 & \textcolor{blue}{\underline{0.390}} & \textcolor{blue}{\underline{0.409}} & 0.400 & 0.420 & 0.522 & 0.465 & 0.502 & 0.443 & 0.395 & 0.413 & \textcolor{red}{\textbf{0.386}} & \textcolor{red}{\textbf{0.407}} & 0.440 & 0.434 \\
 & 336 & 0.436 & 0.449 & 0.417 & 0.436 & 0.411 & 0.432 & 0.422 & 0.440 & 0.420 & 0.436 & \textcolor{blue}{\underline{0.404}} & \textcolor{blue}{\underline{0.426}} & 0.559 & 0.484 & 0.580 & 0.460 & 0.447 & 0.453 & \textcolor{red}{\textbf{0.404}} & \textcolor{red}{\textbf{0.422}} & 0.514 & 0.474 \\
 & 720 & 0.471 & 0.482 & 0.433 & 0.459 & 0.444 & 0.463 & 0.481 & 0.493 & 0.437 & 0.460 & \textcolor{red}{\textbf{0.420}} & \textcolor{red}{\textbf{0.447}} & 0.714 & 0.549 & 0.605 & 0.495 & 0.457 & 0.462 & \textcolor{blue}{\underline{0.424}} & \textcolor{blue}{\underline{0.448}} & 0.705 & 0.568 \\ \cmidrule{2-24}
 & Avg & 0.419 & 0.438 & 0.403 & 0.427 & \textcolor{blue}{\underline{0.401}} & 0.425 & 0.411 & 0.434 & 0.402 & \textcolor{blue}{\underline{0.424}} & 0.402 & 0.425 & 0.562 & 0.483 & 0.533 & 0.452 & 0.412 & 0.427 & \textcolor{red}{\textbf{0.395}} & \textcolor{red}{\textbf{0.417}} & 0.510 & 0.469 \\ \midrule
\multirow[c]{5}{*}{\rotatebox{90}{ETTh2}} & 96 & 0.278 & 0.339 & 0.295 & 0.345 & 0.300 & 0.350 & \textcolor{red}{\textbf{0.271}} & \textcolor{blue}{\underline{0.333}} & 0.278 & 0.342 & 0.298 & 0.362 & 0.316 & 0.359 & 0.306 & 0.338 & 0.292 & 0.352 & \textcolor{blue}{\underline{0.277}} & 0.335 & 0.287 & \textcolor{red}{\textbf{0.325}} \\
 & 192 & 0.336 & 0.381 & 0.353 & 0.388 & 0.351 & 0.390 & \textcolor{red}{\textbf{0.327}} & 0.376 & 0.345 & 0.383 & 0.336 & 0.385 & 0.374 & 0.398 & 0.396 & 0.394 & 0.347 & 0.379 & \textcolor{blue}{\underline{0.334}} & \textcolor{blue}{\underline{0.373}} & 0.347 & \textcolor{red}{\textbf{0.367}} \\
 & 336 & 0.373 & 0.410 & 0.373 & 0.409 & 0.374 & 0.411 & \textcolor{blue}{\underline{0.354}} & 0.402 & 0.394 & 0.416 & \textcolor{red}{\textbf{0.353}} & \textcolor{blue}{\underline{0.399}} & 0.381 & 0.410 & 0.423 & 0.417 & 0.406 & 0.419 & 0.362 & 0.402 & 0.377 & \textcolor{red}{\textbf{0.393}} \\
 & 720 & 0.411 & 0.441 & 0.413 & 0.440 & 0.416 & 0.444 & \textcolor{red}{\textbf{0.381}} & 0.435 & 0.430 & 0.445 & \textcolor{blue}{\underline{0.395}} & \textcolor{blue}{\underline{0.432}} & 0.408 & 0.434 & 0.442 & 0.439 & 0.439 & 0.447 & 0.408 & 0.444 & 0.404 & \textcolor{red}{\textbf{0.421}} \\ \cmidrule{2-24}
 & Avg & 0.350 & 0.393 & 0.359 & 0.396 & 0.360 & 0.399 & \textcolor{red}{\textbf{0.333}} & \textcolor{blue}{\underline{0.387}} & 0.362 & 0.397 & 0.346 & 0.395 & 0.370 & 0.400 & 0.392 & 0.397 & 0.371 & 0.399 & \textcolor{blue}{\underline{0.345}} & 0.389 & 0.354 & \textcolor{red}{\textbf{0.377}} \\ \midrule
\multirow[c]{5}{*}{\rotatebox{90}{Weather}} & 96 & \textcolor{red}{\textbf{0.148}} & \textcolor{red}{\textbf{0.204}} & \textcolor{blue}{\underline{0.151}} & 0.209 & 0.152 & 0.211 & 0.157 & \textcolor{blue}{\underline{0.205}} & 0.151 & 0.211 & 0.200 & 0.260 & 0.190 & 0.236 & 0.186 & 0.208 & 0.157 & 0.211 & 0.183 & 0.242 & 0.177 & 0.208 \\
 & 192 & \textcolor{red}{\textbf{0.189}} & \textcolor{red}{\textbf{0.243}} & 0.194 & 0.250 & 0.199 & 0.254 & 0.205 & 0.251 & \textcolor{blue}{\underline{0.191}} & \textcolor{blue}{\underline{0.248}} & 0.239 & 0.288 & 0.261 & 0.293 & 0.238 & 0.258 & 0.208 & 0.256 & 0.229 & 0.285 & 0.219 & 0.249 \\
 & 336 & \textcolor{red}{\textbf{0.233}} & \textcolor{red}{\textbf{0.278}} & 0.241 & 0.286 & 0.250 & 0.293 & 0.253 & 0.289 & \textcolor{blue}{\underline{0.236}} & \textcolor{blue}{\underline{0.284}} & 0.279 & 0.315 & 0.332 & 0.340 & 0.313 & 0.353 & 0.255 & 0.290 & 0.289 & 0.330 & 0.277 & 0.292 \\
 & 720 & \textcolor{red}{\textbf{0.289}} & \textcolor{red}{\textbf{0.320}} & 0.301 & 0.330 & 0.307 & 0.333 & 0.320 & 0.336 & \textcolor{blue}{\underline{0.299}} & \textcolor{blue}{\underline{0.323}} & 0.340 & 0.357 & 0.385 & 0.381 & 0.416 & 0.415 & 0.405 & 0.397 & 0.359 & 0.370 & 0.365 & 0.350 \\ \cmidrule{2-24}
 & Avg & \textcolor{red}{\textbf{0.215}} & \textcolor{red}{\textbf{0.261}} & 0.222 & 0.269 & 0.227 & 0.273 & 0.234 & 0.270 & \textcolor{blue}{\underline{0.219}} & \textcolor{blue}{\underline{0.267}} & 0.265 & 0.305 & 0.292 & 0.313 & 0.288 & 0.309 & 0.256 & 0.289 & 0.265 & 0.307 & 0.260 & 0.275 \\ \midrule
\multirow[c]{5}{*}{\rotatebox{90}{Electricity}} & 96 & 0.206 & 0.294 & 0.170 & 0.269 & 0.161 & 0.261 & \textcolor{red}{\textbf{0.132}} & \textcolor{red}{\textbf{0.229}} & 0.181 & 0.272 & 0.209 & 0.307 & 0.210 & 0.312 & - & - & - & - & 0.166 & 0.263 & \textcolor{blue}{\underline{0.152}} & \textcolor{blue}{\underline{0.242}} \\
 & 192 & 0.247 & 0.338 & 0.198 & 0.298 & 0.190 & 0.292 & \textcolor{red}{\textbf{0.152}} & \textcolor{red}{\textbf{0.250}} & 0.195 & 0.284 & 0.219 & 0.315 & 0.239 & 0.337 & - & - & - & - & 0.191 & 0.286 & \textcolor{blue}{\underline{0.171}} & \textcolor{blue}{\underline{0.259}} \\
 & 336 & 0.310 & 0.398 & 0.230 & 0.331 & 0.227 & 0.326 & \textcolor{red}{\textbf{0.173}} & \textcolor{red}{\textbf{0.271}} & 0.253 & 0.338 & 0.236 & 0.330 & 0.284 & 0.372 & - & - & - & - & 0.237 & 0.336 & \textcolor{blue}{\underline{0.192}} & \textcolor{blue}{\underline{0.278}} \\
 & 720 & 0.383 & 0.455 & 0.288 & 0.379 & 0.289 & 0.372 &\textcolor{red}{\textbf{0.218}} & \textcolor{red}{\textbf{0.311}} & 0.304 & 0.382 & 0.273 & 0.328 & 0.456 & 0.479 & - & - & - & - & 0.292 & 0.384 & \textcolor{blue}{\underline{0.236}} & \textcolor{blue}{\underline{0.313}} \\ \cmidrule{2-24}
 & Avg & 0.287 & 0.371 & 0.222 & 0.319 & 0.217 & 0.313 & \textcolor{red}{\textbf{0.169}} & \textcolor{red}{\textbf{0.265}} & 0.233 & 0.319 & 0.234 & 0.320 & 0.297 & 0.375 & - & - & - & - & 0.222 & 0.317 & \textcolor{blue}{\underline{0.188}} & \textcolor{blue}{\underline{0.273}} \\ \midrule
\multirow[c]{5}{*}{\rotatebox{90}{Traffic}} & 96 & 0.591 & 0.401 & 0.522 & 0.360 & \textcolor{red}{\textbf{0.480}} & \textcolor{red}{\textbf{0.344}} & - & - & 0.525 & 0.356 & 0.572 & 0.407 & 0.526 & 0.368 & 0.562 & 0.378 & - & - & \textcolor{blue}{\underline{0.514}} & \textcolor{blue}{\underline{0.347}} & - & - \\
 & 192 & 0.623 & 0.431 & \textcolor{blue}{\underline{0.538}} & 0.374 & \textcolor{red}{\textbf{0.511}} & \textcolor{blue}{\underline{0.369}} & - & - & 0.547 & \textcolor{red}{\textbf{0.365}} & 0.575 & 0.406 & 0.561 & 0.385 & 0.579 & 0.412 & - & - & 0.543 & 0.373 & - & - \\
 & 336 & 0.697 & 0.485 & \textcolor{blue}{\underline{0.574}} & 0.400 & \textcolor{red}{\textbf{0.553}} & \textcolor{blue}{\underline{0.394}} & - & - & 0.645 & 0.419 & 0.588 & 0.411 & 0.614 & 0.412 & 0.594 & 0.420 & - & - & 0.575 & \textcolor{red}{\textbf{0.389}} & - & - \\
 & 720 & 0.782 & 0.528 & 0.669 & 0.449 & 0.639 & \textcolor{blue}{\underline{0.429}} & - & - & 0.722 & 0.457 & \textcolor{red}{\textbf{0.618}} & \textcolor{red}{\textbf{0.422}} & 0.749 & 0.464 & 0.723 & 0.472 & - & - & \textcolor{blue}{\underline{0.622}} & 0.433 & - & - \\ \cmidrule{2-24}
 & Avg & 0.673 & 0.461 & 0.576 & 0.396 & \textcolor{red}{\textbf{0.546}} & \textcolor{red}{\textbf{0.384}} & - & - & 0.610 & 0.399 & 0.588 & 0.412 & 0.613 & 0.407 & 0.615 & 0.421 & - & - & \textcolor{blue}{\underline{0.564}} & \textcolor{blue}{\underline{0.386}} & - & - \\ \midrule
\multirow[c]{5}{*}{\rotatebox{90}{Solar}} & 96 & 0.185 & 0.263 & \textcolor{blue}{\underline{0.167}} & 0.259 & \textcolor{red}{\textbf{0.153}} & \textcolor{blue}{\underline{0.255}} & 0.204 & \textcolor{red}{\textbf{0.230}} & 0.209 & 0.309 & 0.524 & 0.557 & 0.591 & 0.504 & 0.373 & 0.304 & 0.304 & 0.345 & 0.863 & 0.664 & 0.682 & 0.688 \\
 & 192 & 0.201 & 0.276 & \textcolor{blue}{\underline{0.180}} & 0.273 & \textcolor{red}{\textbf{0.165}} & \textcolor{blue}{\underline{0.269}} & 0.221 & \textcolor{red}{\textbf{0.248}} & 0.220 & 0.315 & 0.507 & 0.550 & 0.689 & 0.567 & 0.363 & 0.303 & 0.309 & 0.342 & 0.823 & 0.695 & 0.694 & 0.695 \\
 & 336 & 0.212 & 0.283 & \textcolor{blue}{\underline{0.190}} & 0.282 & \textcolor{red}{\textbf{0.174}} & \textcolor{blue}{\underline{0.278}} & 0.225 & \textcolor{red}{\textbf{0.260}} & 0.228 & 0.314 & 0.508 & 0.553 & 0.831 & 0.636 & 0.391 & 0.319 & 0.433 & 0.450 & 0.835 & 0.741 & 0.719 & 0.706 \\
 & 720 & 0.213 & 0.291 & \textcolor{blue}{\underline{0.198}} & 0.295 & \textcolor{red}{\textbf{0.179}} & 0.287 & 0.233 & \textcolor{red}{\textbf{0.272}} & 0.218 & \textcolor{blue}{\underline{0.283}} & 0.479 & 0.534 & 0.972 & 0.710 & 0.444 & 0.349 & 0.599 & 0.576 & 0.738 & 0.738 & 0.759 & 0.725 \\ \cmidrule{2-24}
 & Avg & 0.203 & 0.278 & \textcolor{blue}{\underline{0.184}} & 0.277 & \textcolor{red}{\textbf{0.168}} & \textcolor{blue}{\underline{0.272}} & 0.221 & \textcolor{red}{\textbf{0.252}} & 0.219 & 0.305 & 0.505 & 0.549 & 0.771 & 0.604 & 0.393 & 0.319 & 0.411 & 0.428 & 0.815 & 0.710 & 0.714 & 0.704 \\ \midrule
\multirow[c]{5}{*}{\rotatebox{90}{PEMS08}} & 96 & 0.249 & 0.357 & \textcolor{blue}{\underline{0.241}} & \textcolor{blue}{\underline{0.353}} & \textcolor{red}{\textbf{0.224}} & \textcolor{red}{\textbf{0.331}} & - & - & 0.608 & 0.562 & 1.373 & 0.995 & 0.625 & 0.580 & 1.538 & 0.983 & - & - & 1.284 & 0.888 & - & - \\
 & 192 & 0.299 & 0.375 & \textcolor{blue}{\underline{0.288}} & \textcolor{blue}{\underline{0.368}} & \textcolor{red}{\textbf{0.273}} & \textcolor{red}{\textbf{0.342}} & - & - & 0.774 & 0.648 & 1.365 & 0.979 & 0.798 & 0.661 & 1.719 & 0.983 & - & - & 1.638 & 1.039 & - & - \\
 & 336 & 0.330 & 0.376 & \textcolor{blue}{\underline{0.324}} & \textcolor{blue}{\underline{0.367}} & \textcolor{red}{\textbf{0.305}} & \textcolor{red}{\textbf{0.348}} & - & - & 0.892 & 0.693 & 1.338 & 0.960 & 0.910 & 0.716 & 1.768 & 0.998 & - & - & 1.979 & 1.160 & - & - \\
 & 720 & 0.418 & 0.428 & \textcolor{blue}{\underline{0.395}} & \textcolor{blue}{\underline{0.406}} & \textcolor{red}{\textbf{0.368}} & \textcolor{red}{\textbf{0.384}} & - & - & 0.975 & 0.864 & 1.401 & 0.980 & 1.131 & 0.824 & 1.802 & 1.133 & - & - & 2.020 & 1.175 & - & - \\ \cmidrule{2-24}
 & Avg & 0.324 & 0.384 & \textcolor{blue}{\underline{0.312}} & \textcolor{blue}{\underline{0.374}} & \textcolor{red}{\textbf{0.293}} & \textcolor{red}{\textbf{0.351}} & - & - & 0.812 & 0.692 & 1.369 & 0.979 & 0.866 & 0.695 & 1.707 & 1.024 & - & - & 1.730 & 1.066 & - & - \\ \midrule
\multirow[c]{5}{*}{\rotatebox{90}{Wind}} & 96 & 0.948 & 0.663 & \textcolor{blue}{\underline{0.906}} & 0.650 & \textcolor{red}{\textbf{0.867}} & \textcolor{red}{\textbf{0.634}} & 0.931 & \textcolor{blue}{\underline{0.646}} & 0.988 & 0.678 & 1.072 & 0.747 & 0.946 & 0.659 & 1.273 & 0.738 & - & - & 1.077 & 0.701 & 0.992 & 0.656 \\
 & 192 & 1.107 & 0.746 & \textcolor{blue}{\underline{1.062}} & \textcolor{blue}{\underline{0.735}} & \textcolor{red}{\textbf{1.005}} & \textcolor{red}{\textbf{0.714}} & 1.136 & 0.751 & 1.275 & 0.862 & 1.209 & 0.804 & 1.142 & 0.758 & 1.439 & 0.817 & - & - & 1.350 & 0.825 & 1.221 & 0.765 \\
 & 336 & 1.215 & 0.801 & \textcolor{blue}{\underline{1.208}} & \textcolor{blue}{\underline{0.801}} & \textcolor{red}{\textbf{1.097}} & \textcolor{red}{\textbf{0.765}} & 1.285 & 0.820 & 1.332 & 0.897 & 1.318 & 0.848 & 1.300 & 0.830 & 1.550 & 0.869 & - & - & 1.473 & 0.891 & 1.403 & 0.844 \\
 & 720 & 1.307 & 0.849 & \textcolor{blue}{\underline{1.303}} & \textcolor{blue}{\underline{0.847}} & \textcolor{red}{\textbf{1.197}} & \textcolor{red}{\textbf{0.817}} & 1.390 & 0.870 & 1.572 & 0.906 & 1.404 & 0.881 & 1.417 & 0.884 & 1.649 & 0.914 & - & - & 1.447 & 0.899 & 1.581 & 0.915 \\ \cmidrule{2-24}
 & Avg & 1.144 & 0.765 & \textcolor{blue}{\underline{1.120}} & \textcolor{blue}{\underline{0.758}} & \textcolor{red}{\textbf{1.042}} & \textcolor{red}{\textbf{0.733}} & 1.186 & 0.772 & 1.292 & 0.836 & 1.251 & 0.820 & 1.201 & 0.783 & 1.478 & 0.834 & - & - & 1.337 & 0.829 & 1.299 & 0.795 \\ \midrule
\multirow[c]{5}{*}{\rotatebox{90}{NYSE}} & 24 & 0.323 & 0.429 & \textcolor{red}{\textbf{0.240}} & \textcolor{red}{\textbf{0.327}} & \textcolor{blue}{\underline{0.301}} & \textcolor{blue}{\underline{0.385}} & 0.370 & 0.393 & 0.362 & 0.403 & - & - & 0.388 & 0.409 & 0.480 & 0.449 & - & - & - & - & - & - \\
 & 36 & 0.451 & 0.520 & \textcolor{red}{\textbf{0.361}} & \textcolor{red}{\textbf{0.391}} & \textcolor{blue}{\underline{0.437}} & \textcolor{blue}{\underline{0.461}} & 0.681 & 0.566 & 0.574 & 0.592 & - & - & 0.778 & 0.613 & 0.912 & 0.655 & - & - & - & - & - & - \\
 & 48 & \textcolor{blue}{\underline{0.571}} & 0.593 & \textcolor{red}{\textbf{0.487}} & \textcolor{red}{\textbf{0.446}} & 0.589 & \textcolor{blue}{\underline{0.538}} & 1.043 & 0.729 & 0.688 & 0.708 & - & - & 1.200 & 0.792 & 1.371 & 0.818 & - & - & - & - & - & -\\
 & 60 & \textcolor{blue}{\underline{0.762}} & 0.671 & \textcolor{red}{\textbf{0.681}} & \textcolor{red}{\textbf{0.557}} & 0.789 & \textcolor{blue}{\underline{0.644}} & 1.425 & 0.880 & 0.855 & 0.802 & - & - & 1.584 & 1.000 & 1.754 & 0.958 & - & - & - & - & - & - \\ \cmidrule{2-24}
 & Avg & \textcolor{blue}{\underline{0.527}} & 0.553 & \textcolor{red}{\textbf{0.442}} & \textcolor{red}{\textbf{0.430}} & 0.529 & \textcolor{blue}{\underline{0.507}} & 0.880 & 0.642 & 0.620 & 0.626 & - & - & 0.988 & 0.704 & 1.129 & 0.720 & - & - & - & - & - & - \\ \midrule
 \rowc
\multicolumn{2}{c|}{1\textsuperscript{st} Count} & 8 & 8 & 6 &4 & \textcolor{red}{\textbf{15}} & \textcolor{red}{\textbf{9}} & \textcolor{blue}{\underline{9}} & \textcolor{blue}{\underline{9}} & 0 & 3 & 3 & 2 & 0 & 0 & 0 & 0 & 1 & 2 & 2 & 3 & 0 & 4 \\
\rowc
\multicolumn{2}{c|}{2\textsuperscript{nd} Count} & 3 & 1 & \textcolor{red}{\textbf{15}} & \textcolor{blue}{\underline{9}} & 2 & \textcolor{red}{\textbf{10}} & 2 & 7 & \textcolor{blue}{\underline{9}} & 6 & 2 & 3 & 0 & 0 & 0 & 0 & 2 & 1 & 5 & 3 & 4 & 4 \\
\bottomrule
\end{tabular}
}
\end{table*}

We provide all the results of the zero-shot probabilistic forecasting in Table~\ref{tab: main prob forecasting all}. As shown in Table~\ref{tab: main prob forecasting all}, we include 9 representative real-world datasets from ProbTS, demonstrating that FLAME achieves state-of-the-art forecasting performance.
\begin{table*}[!htbp]
\caption{Full results of zero-shot probabilistic forecasting experiments on datasets from ProbTS. Lower CRPS or NMAE values indicate better predictions. (``-'') denotes datasets included in the model’s pretraining and therefore excluded from testing. (``/'') denotes the excessive time consumption. \textcolor{red}{\textbf{Red}}: the best, \textcolor{blue}{\underline{Blue}}: the 2nd best.}
\label{tab: main prob forecasting all}
\resizebox{1\linewidth}{!}{
\begin{tabular}{cc|cc|cc|cc|cc|cc|cc|cc|cc|cc|cc|cc}
\toprule
\multicolumn{2}{c|}{\multirow{2}{*}{Models}} & \multicolumn{2}{c|}{\textbf{FLAME}} & \multicolumn{2}{c|}{\textbf{FLAME}} & \multicolumn{2}{c|}{\textbf{FLAME}} & \multicolumn{2}{c|}{\textbf{Sundial}} & \multicolumn{2}{c|}{\textbf{Chronos}} & \multicolumn{2}{c|}{\textbf{MOIRAI}} & \multicolumn{2}{c|}{\textbf{Lag-Llama}} & \multicolumn{2}{c|}{\textbf{TSDiff}} & \multicolumn{2}{c|}{\textbf{CSDI}} & \multicolumn{2}{c|}{\textbf{TimeGrad}} & \multicolumn{2}{c}{\textbf{GRU NVP}} \\    
\multicolumn{2}{c|}{~} & \multicolumn{2}{c|}{\textbf{Small}} & \multicolumn{2}{c|}{\textbf{Base}} &\multicolumn{2}{c|}{\textbf{Large}} & \multicolumn{2}{c|}{(2025)} & \multicolumn{2}{c|}{(2024)} & \multicolumn{2}{c|}{(2024)} & \multicolumn{2}{c|}{(2023)} & \multicolumn{2}{c|}{(2023)} & \multicolumn{2}{c|}{(2022)} & \multicolumn{2}{c|}{(2022)} & \multicolumn{2}{c}{(2021)}\\
\midrule
\multicolumn{2}{c|}{Metrics} & CRPS & NMAE & CRPS & NMAE & CRPS & NMAE & CRPS & NMAE & CRPS & NMAE & CRPS & NMAE & CRPS & NMAE & CRPS & NMAE & CRPS & NMAE & CRPS & NMAE & CRPS & NMAE \\  
\midrule
 & 96 & 0.294 & 0.368 & 0.326 & 0.417 & 0.315 & 0.407 & \textcolor{blue}{\underline{ 0.253}} & \textcolor{blue}{\underline{ 0.308}} & 0.360 & 0.422 & 0.464 & 0.522 & 0.354 & 0.402 & 0.344 & 0.441 & \textcolor{red}{ \textbf{0.236}} & \textcolor{red}{ \textbf{0.308}} & 0.522 & 0.645 & 0.383 & 0.488 \\
 & 192 & 0.332 & 0.420 & 0.363 & 0.469 & 0.337 & 0.426 & \textcolor{red}{ \textbf{0.279}} & \textcolor{red}{ \textbf{0.337}} & 0.404 & 0.450 & 0.467 & 0.531 & 0.368 & 0.415 & 0.345 & 0.441 & \textcolor{blue}{\underline{ 0.291}} & \textcolor{blue}{\underline{ 0.377}} & 0.603 & 0.748 & 0.396 & 0.514 \\
 & 336 & 0.341 & 0.427 & 0.393 & 0.511 & 0.351 & 0.447 & \textcolor{red}{ \textbf{0.291}} & \textcolor{red}{ \textbf{0.350}} & 0.425 & 0.456 & 0.524 & 0.558 & 0.387 & 0.436 & 0.462 & 0.571 & \textcolor{blue}{\underline{ 0.322}} & \textcolor{blue}{\underline{ 0.419}} & 0.601 & 0.759 & 0.486 & 0.630 \\
 & 720 & \textcolor{blue}{\underline{ 0.364}} & \textcolor{blue}{\underline{ 0.454}} & 0.477 & 0.622 & 0.383 & 0.476 & \textcolor{red}{ \textbf{0.318}} & \textcolor{red}{ \textbf{0.380}} & 0.461 & 0.478 & 0.514 & 0.535 & 0.403 & 0.466 & 0.478 & 0.622 & 0.448 & 0.578 & 0.621 & 0.793 & 0.546 & 0.707 \\
 \cmidrule{2-24}
\multirow{-5}{*}{\rotatebox{90}{ETTm1}} & avg & 0.333 & \textcolor{blue}{\underline{ 0.417}} & 0.390 & 0.505 & 0.347 & 0.439 & \textcolor{red}{ \textbf{0.285}} & \textcolor{red}{ \textbf{0.344}} & 0.413 & 0.451 & 0.492 & 0.536 & 0.378 & 0.430 & 0.407 & 0.519 & \textcolor{blue}{\underline{ 0.324}} & 0.421 & 0.587 & 0.736 & 0.453 & 0.585 \\
\midrule
 & 96 & \textcolor{blue}{\underline{ 0.120}} & \textcolor{blue}{\underline{ 0.153}} & 0.121 & 0.153 & 0.124 & 0.155 & 0.128 & 0.153 & 0.134 & 0.158 & 0.176 & 0.186 & 0.163 & 0.192 & 0.175 & 0.224 & \textcolor{red}{ \textbf{0.115}} & \textcolor{red}{ \textbf{0.146}} & 0.427 & 0.525 & 0.319 & 0.413 \\
 & 192 & \textcolor{blue}{\underline{ 0.146}} & 0.182 & 0.147 & 0.184 & \textcolor{red}{ \textbf{0.143}} & \textcolor{red}{ \textbf{0.175}} & 0.150 & \textcolor{blue}{\underline{ 0.177}} & 0.163 & 0.183 & 0.197 & 0.207 & 0.181 & 0.207 & 0.255 & 0.316 & 0.147 & 0.189 & 0.424 & 0.530 & 0.326 & 0.427 \\
 & 336 & \textcolor{blue}{\underline{ 0.163}} & 0.202 & 0.164 & 0.205 & \textcolor{red}{ \textbf{0.158}} & \textcolor{red}{ \textbf{0.193}} & 0.167 & \textcolor{blue}{\underline{ 0.195}} & 0.190 & 0.204 & 0.229 & 0.227 & 0.206 & 0.229 & 0.328 & 0.397 & 0.190 & 0.248 & 0.469 & 0.566 & 0.449 & 0.580 \\
 & 720 & \textcolor{blue}{\underline{ 0.185}} & 0.223 & 0.186 & 0.231 & \textcolor{red}{ \textbf{0.184}} & \textcolor{blue}{\underline{ 0.219}} & 0.189 & \textcolor{red}{ \textbf{0.217}} & 0.223 & 0.230 & 0.321 & 0.258 & 0.227 & 0.249 & 0.344 & 0.416 & 0.239 & 0.306 & 0.470 & 0.561 & 0.561 & 0.749 \\
 \cmidrule{2-24}
\multirow{-5}{*}{\rotatebox{90}{ETTm2}} & avg & \textcolor{blue}{\underline{ 0.154}} & 0.190 & 0.155 & 0.193 & \textcolor{red}{ \textbf{0.152}} & \textcolor{red}{ \textbf{0.186}} & 0.158 & \textcolor{blue}{\underline{ 0.186}} & 0.178 & 0.194 & 0.231 & 0.220 & 0.194 & 0.219 & 0.276 & 0.338 & 0.173 & 0.222 & 0.448 & 0.546 & 0.414 & 0.542 \\
\midrule
 & 96 & 0.265 & 0.322 & 0.261 & 0.319 & \textcolor{red}{ \textbf{0.259}} & \textcolor{blue}{\underline{ 0.310}} & \textcolor{blue}{\underline{ 0.260}} & \textcolor{red}{ \textbf{0.307}} & 0.293 & 0.341 & 0.469 & 0.488 & 0.297 & 0.340 & 0.395 & 0.510 & 0.437 & 0.557 & 0.455 & 0.585 & 0.379 & 0.481 \\
 & 192 & 0.288 & \textcolor{blue}{\underline{ 0.346}} & \textcolor{blue}{\underline{ 0.282}} & 0.351 & \textcolor{red}{ \textbf{0.278}} & \textcolor{red}{ \textbf{0.342}} & 0.297 & 0.347 & 0.348 & 0.384 & 0.496 & 0.492 & 0.312 & 0.356 & 0.467 & 0.596 & 0.496 & 0.625 & 0.516 & 0.680 & 0.425 & 0.531 \\
 & 336 & 0.300 & 0.371 & \textcolor{blue}{\underline{ 0.297}} & 0.368 & \textcolor{red}{ \textbf{0.290}} & \textcolor{red}{ \textbf{0.362}} & 0.314 & \textcolor{blue}{\underline{ 0.365}} & 0.384 & 0.409 & 0.503 & 0.489 & 0.326 & 0.370 & 0.450 & 0.581 & 0.454 & 0.574 & 0.512 & 0.666 & 0.458 & 0.580 \\
 & 720 & 0.320 & 0.402 & \textcolor{blue}{\underline{ 0.312}} & 0.386 & \textcolor{red}{ \textbf{0.303}} & \textcolor{blue}{\underline{ 0.367}} & 0.312 & \textcolor{red}{ \textbf{0.366}} & 0.413 & 0.425 & 0.526 & 0.508 & 0.340 & 0.405 & 0.516 & 0.657 & 0.528 & 0.657 & 0.523 & 0.672 & 0.502 & 0.643 \\
  \cmidrule{2-24}
\multirow{-5}{*}{\rotatebox{90}{ETTh1}} & avg & 0.293 & 0.360 & \textcolor{blue}{\underline{ 0.288}} & 0.356 & \textcolor{red}{ \textbf{0.283}} & \textcolor{red}{ \textbf{0.345}} & 0.296 & \textcolor{blue}{\underline{ 0.346}} & 0.360 & 0.390 & 0.498 & 0.494 & 0.319 & 0.368 & 0.457 & 0.586 & 0.479 & 0.603 & 0.502 & 0.651 & 0.441 & 0.559 \\
\midrule
 & 96 & \textcolor{red}{ \textbf{0.139}} & \textcolor{red}{ \textbf{0.177}} & 0.145 & \textcolor{blue}{\underline{ 0.185}} & \textcolor{blue}{\underline{ 0.144}} & 0.186 & 0.157 & 0.186 & 0.163 & 0.191 & 0.212 & 0.238 & 0.178 & 0.204 & 0.336 & 0.421 & 0.164 & 0.214 & 0.358 & 0.448 & 0.432 & 0.548 \\
 & 192 & \textcolor{red}{ \textbf{0.152}} & \textcolor{red}{ \textbf{0.195}} & 0.161 & 0.207 & \textcolor{blue}{\underline{ 0.159}} & \textcolor{blue}{\underline{ 0.204}} & 0.183 & 0.213 & 0.195 & 0.218 & 0.229 & 0.250 & 0.193 & 0.217 & 0.265 & 0.339 & 0.226 & 0.294 & 0.457 & 0.575 & 0.625 & 0.766 \\
 & 336 & \textcolor{red}{ \textbf{0.170}} & \textcolor{red}{ \textbf{0.216}} & 0.174 & 0.224 & \textcolor{blue}{\underline{ 0.172}} & \textcolor{blue}{\underline{ 0.217}} & 0.203 & 0.232 & 0.226 & 0.241 & 0.253 & 0.263 & 0.211 & 0.230 & 0.350 & 0.427 & 0.274 & 0.353 & 0.481 & 0.606 & 0.793 & 0.942 \\
 & 720 & \textcolor{blue}{\underline{ 0.183}} & \textcolor{blue}{\underline{ 0.231}} & 0.191 & 0.241 & \textcolor{red}{ \textbf{0.182}} & \textcolor{red}{ \textbf{0.228}} & 0.203 & 0.233 & 0.255 & 0.263 & 0.279 & 0.273 & 0.222 & 0.238 & 0.406 & 0.482 & 0.302 & 0.382 & 0.445 & 0.550 & 0.539 & 0.688 \\
  \cmidrule{2-24}
\multirow{-5}{*}{\rotatebox{90}{ETTh2}} & avg & \textcolor{red}{ \textbf{0.161}} & \textcolor{red}{ \textbf{0.205}} & 0.168 & 0.214 & \textcolor{blue}{\underline{ 0.164}} & \textcolor{blue}{\underline{ 0.209}} & 0.186 & 0.216 & 0.210 & 0.228 & 0.243 & 0.256 & 0.201 & 0.222 & 0.339 & 0.417 & 0.242 & 0.311 & 0.435 & 0.545 & 0.597 & 0.736 \\
\midrule
 & 96 & 0.071 & 0.088 & 0.071 & \textcolor{blue}{\underline{ 0.085}} & \textcolor{red}{ \textbf{0.067}} & \textcolor{red}{ \textbf{0.083}} & 0.078 & 0.091 & 0.136 & 0.155 & 0.161 & 0.141 & 0.085 & 0.094 & 0.104 & 0.113 & \textcolor{blue}{\underline{ 0.068}} & 0.087 & 0.130 & 0.164 & 0.116 & 0.145 \\
 & 192 & 0.089 & 0.113 & 0.081 & 0.100 & \textcolor{red}{ \textbf{0.068}} & \textcolor{red}{ \textbf{0.082}} & 0.083 & 0.097 & 0.145 & 0.165 & 0.173 & 0.139 & 0.094 & 0.102 & 0.134 & 0.144 & \textcolor{blue}{\underline{ 0.068}} & \textcolor{blue}{\underline{ 0.086}} & 0.127 & 0.158 & 0.122 & 0.147 \\
 & 336 & 0.094 & 0.121 & 0.090 & 0.110 & \textcolor{red}{ \textbf{0.079}} & \textcolor{red}{ \textbf{0.094}} & 0.090 & 0.106 & 0.136 & 0.151 & 0.149 & 0.134 & 0.098 & 0.109 & 0.137 & 0.138 & \textcolor{blue}{\underline{ 0.083}} & \textcolor{blue}{\underline{ 0.098}} & 0.130 & 0.162 & 0.128 & 0.160 \\
 & 720 & 0.103 & 0.132 & 0.106 & 0.125 & \textcolor{red}{ \textbf{0.083}} & \textcolor{red}{ \textbf{0.100}} & 0.096 & 0.113 & 0.151 & 0.161 & 0.233 & 0.155 & 0.107 & 0.120 & 0.152 & 0.141 & \textcolor{blue}{\underline{ 0.087}} & \textcolor{blue}{\underline{ 0.102}} & 0.113 & 0.136 & 0.110 & 0.135 \\
  \cmidrule{2-24}
\multirow{-5}{*}{\rotatebox{90}{Weather}} & avg & 0.089 & 0.114 & 0.087 & 0.105 & \textcolor{red}{ \textbf{0.074}} & \textcolor{red}{ \textbf{0.090}} & 0.087 & 0.102 & 0.142 & 0.158 & 0.179 & 0.143 & 0.096 & 0.106 & 0.132 & 0.134 & \textcolor{blue}{\underline{ 0.077}} & \textcolor{blue}{\underline{ 0.093}} & 0.125 & 0.155 & 0.119 & 0.147 \\
\midrule
 & 96 & 0.071 & 0.090 & \textcolor{red}{ \textbf{0.066}} & \textcolor{blue}{\underline{ 0.084}} & 0.069 & 0.089 & \textcolor{blue}{\underline{ 0.068}} & \textcolor{red}{ \textbf{0.083}} & - & - & 0.231 & 0.275 & - & - & 0.344 & 0.441 & 0.153 & 0.203 & 0.096 & 0.119 & 0.094 & 0.118 \\
 & 192 & 0.086 & 0.103 & 0.076 & 0.098 & \textcolor{red}{ \textbf{0.072}} & \textcolor{red}{ \textbf{0.088}} & \textcolor{blue}{\underline{ 0.074}} & \textcolor{blue}{\underline{ 0.090}} & - & - & 0.236 & 0.279 & - & - & 0.345 & 0.441 & 0.200 & 0.264 & 0.100 & 0.124 & 0.097 & 0.121 \\
 & 336 & 0.096 & 0.125 & 0.091 & 0.117 & \textcolor{red}{ \textbf{0.081}} & \textcolor{red}{ \textbf{0.096}} & \textcolor{blue}{\underline{ 0.083}} & \textcolor{blue}{\underline{ 0.100}} & - & - & 0.245 & 0.289 & - & - & 0.462 & 0.571 & / & / & 0.102 & 0.126 & 0.099 & 0.123 \\
 & 720 & 0.117 & 0.154 & 0.110 & 0.141 & \textcolor{blue}{\underline{ 0.100}} & \textcolor{red}{ \textbf{0.116}} & \textcolor{red}{ \textbf{0.098}} & \textcolor{blue}{\underline{ 0.118}} & - & - & 0.274 & 0.318 & - & - & 0.478 & 0.622 & / & / & 0.108 & 0.134 & 0.114 & 0.144 \\
  \cmidrule{2-24}
\multirow{-5}{*}{\rotatebox{90}{Electricity}} & avg & 0.093 & 0.118 & 0.086 & 0.110 & \textcolor{red}{ \textbf{0.081}} & \textcolor{red}{ \textbf{0.097}} & \textcolor{blue}{\underline{ 0.081}} & \textcolor{blue}{\underline{ 0.098}} & - & - & 0.247 & 0.290 & - & - & 0.407 & 0.519 & / & / & 0.102 & 0.126 & 0.101 & 0.127 \\
\midrule
 & 96 & 0.240 & 0.293 & 0.229 & 0.277 & 0.220 & 0.268 & - & - & 0.250 & 0.289 & - & - & 0.255 & 0.297 & 0.294 & 0.342 & / & / & \textcolor{blue}{\underline{ 0.202}} & \textcolor{blue}{\underline{ 0.234}} & \textcolor{red}{ \textbf{0.187}} & \textcolor{red}{ \textbf{0.231}} \\
 & 192 & 0.257 & 0.319 & 0.239 & 0.291 & 0.238 & 0.294 & - & - & 0.249 & 0.278 & - & - & 0.295 & 0.343 & 0.306 & 0.354 & / & / & \textcolor{blue}{\underline{ 0.208}} & \textcolor{blue}{\underline{ 0.239}} & \textcolor{red}{ \textbf{0.192}} & \textcolor{red}{ \textbf{0.236}} \\
 & 336 & 0.271 & 0.341 & 0.250 & 0.311 & 0.254 & 0.319 & - & - & 0.269 & 0.290 & - & - & 0.335 & 0.384 & 0.317 & 0.392 & / & / & \textcolor{blue}{\underline{ 0.213}} & \textcolor{red}{ \textbf{0.246}} & \textcolor{red}{ \textbf{0.201}} & \textcolor{blue}{\underline{ 0.248}} \\
 & 720 & 0.304 & 0.395 & 0.274 & 0.346 & 0.276 & 0.351 & - & - & 0.310 & 0.321 & - & - & 0.434 & 0.517 & 0.391 & 0.478 & / & / & \textcolor{blue}{\underline{ 0.220}} & \textcolor{red}{ \textbf{0.263}} & \textcolor{red}{ \textbf{0.211}} & \textcolor{blue}{\underline{ 0.264}} \\
  \cmidrule{2-24}
\multirow{-5}{*}{\rotatebox{90}{Traffic}} & avg & 0.268 & 0.337 & 0.248 & 0.306 & 0.247 & 0.308 & - & - & 0.269 & 0.295 & - & - & 0.330 & 0.385 & 0.327 & 0.392 & / & / & \textcolor{blue}{\underline{ 0.211}} & \textcolor{blue}{\underline{ 0.246}} & \textcolor{red}{ \textbf{0.198}} & \textcolor{red}{ \textbf{0.245}} \\
\midrule
 & 96 & 0.023 & 0.028 & \textcolor{blue}{\underline{ 0.021}} & 0.027 & \textcolor{red}{ \textbf{0.020}} & \textcolor{blue}{\underline{ 0.026}} & 0.024 & 0.027 & 0.021 & \textcolor{red}{ \textbf{0.025}} & 0.025 & 0.030 & 0.042 & 0.051 & 0.079 & 0.090 & 0.028 & 0.036 & 0.068 & 0.079 & 0.071 & 0.091 \\
 & 192 & 0.032 & 0.037 & \textcolor{red}{ \textbf{0.030}} & 0.036 & \textcolor{blue}{\underline{ 0.031}} & \textcolor{red}{ \textbf{0.035}} & 0.034 & 0.038 & 0.032 & \textcolor{blue}{\underline{ 0.036}} & 0.034 & 0.039 & 0.047 & 0.058 & 0.093 & 0.106 & 0.045 & 0.058 & 0.087 & 0.100 & 0.068 & 0.087 \\
 & 336 & \textcolor{blue}{\underline{ 0.044}} & 0.049 & \textcolor{red}{ \textbf{0.041}} & \textcolor{blue}{\underline{ 0.047}} & 0.044 & \textcolor{red}{ \textbf{0.046}} & 0.046 & 0.050 & 0.045 & 0.048 & 0.047 & 0.052 & 0.061 & 0.073 & 0.081 & 0.106 & 0.060 & 0.076 & 0.074 & 0.086 & 0.072 & 0.091 \\
 & 720 & 0.074 & \textcolor{blue}{\underline{ 0.079}} & \textcolor{red}{ \textbf{0.069}} & \textcolor{red}{ \textbf{0.077}} & \textcolor{blue}{\underline{ 0.073}} & 0.079 & 0.076 & 0.079 & 0.078 & 0.080 & 0.073 & 0.081 & 0.078 & 0.094 & 0.082 & 0.142 & 0.143 & 0.173 & 0.099 & 0.113 & 0.079 & 0.103 \\
  \cmidrule{2-24}
\multirow{-5}{*}{\rotatebox{90}{Exchange}} & avg & 0.043 & 0.048 & \textcolor{red}{ \textbf{0.040}} & \textcolor{blue}{\underline{ 0.047}} & \textcolor{blue}{\underline{ 0.042}} & \textcolor{red}{ \textbf{0.047}} & 0.045 & 0.049 & 0.044 & 0.047 & 0.045 & 0.050 & 0.057 & 0.069 & 0.084 & 0.111 & 0.069 & 0.086 & 0.082 & 0.095 & 0.073 & 0.093 \\
\midrule
 & 24 & 0.135 & 0.166 & \textcolor{blue}{\underline{ 0.111}} & 0.143 & 0.136 & 0.168 & \textcolor{red}{ \textbf{0.108}} & \textcolor{red}{ \textbf{0.123}} & 0.120 & \textcolor{blue}{\underline{ 0.139}} & 0.150 & 0.196 & 0.135 & 0.173 & 0.228 & 0.242 & 0.250 & 0.263 & 0.275 & 0.296 & 0.257 & 0.283 \\
 & 36 & 0.164 & 0.200 & \textcolor{red}{ \textbf{0.119}} & \textcolor{red}{ \textbf{0.153}} & \textcolor{blue}{\underline{ 0.145}} & 0.179 & 0.152 & \textcolor{blue}{\underline{ 0.169}} & 0.179 & 0.205 & 0.171 & 0.222 & 0.163 & 0.227 & 0.235 & 0.246 & 0.285 & 0.298 & 0.272 & 0.298 & 0.281 & 0.307 \\
 & 48 & 0.176 & 0.216 & \textcolor{red}{ \textbf{0.139}} & \textcolor{red}{ \textbf{0.178}} & 0.164 & 0.200 & 0.164 & 0.184 & 0.186 & 0.215 & \textcolor{blue}{\underline{ 0.151}} & \textcolor{blue}{\underline{ 0.184}} & 0.171 & 0.233 & 0.265 & 0.275 & 0.285 & 0.301 & 0.295 & 0.320 & 0.288 & 0.314 \\
 & 60 & 0.171 & 0.211 & \textcolor{red}{ \textbf{0.149}} & \textcolor{blue}{\underline{ 0.189}} & 0.169 & 0.204 & 0.168 & 0.190 & 0.196 & 0.228 & 0.163 & \textcolor{red}{ \textbf{0.188}} & \textcolor{blue}{\underline{ 0.156}} & 0.211 & 0.263 & 0.272 & 0.283 & 0.299 & 0.295 & 0.325 & 0.307 & 0.333 \\
  \cmidrule{2-24}
\multirow{-5}{*}{\rotatebox{90}{ILI}} & avg & 0.162 & 0.198 & \textcolor{red}{ \textbf{0.130}} & \textcolor{red}{ \textbf{0.166}} & 0.154 & 0.188 & \textcolor{blue}{\underline{ 0.148}} & \textcolor{blue}{\underline{ 0.166}} & 0.170 & 0.197 & 0.159 & 0.197 & 0.156 & 0.211 & 0.248 & 0.259 & 0.276 & 0.290 & 0.284 & 0.310 & 0.283 & 0.309 \\
\midrule
\rowc
\multicolumn{2}{c|}{1\textsuperscript{st} Count} &3 &3 &\textcolor{blue}{\underline{7}} &3 &\textcolor{red}{\textbf{15}} &\textcolor{red}{\textbf{14}} &5 &\textcolor{blue}{\underline{8}} &0 &1 &0 &1 &0 &0 &0 &0 &2 &2 &0 &2 &4 &2 \\      \rowc
\multicolumn{2}{c|}{2\textsuperscript{nd} Count} & \textcolor{blue}{\underline{7}} & 5 & 5 & 5 & \textcolor{red}{\textbf{7}} & \textcolor{blue}{\underline{6}} & 5 & \textcolor{red}{\textbf{8}} & 0 & 2 & 1 & 1 & 1 & 0 & 0 & 0 & 6 & 5 & 4 & 2 & 0 & 2\\ 
\bottomrule
        
\end{tabular}}
\end{table*}

\begin{table*}[!htbp]
\centering
\caption{Full results of zero-shot deterministic forecasting experiments on datasets from TSFM-Bench. Lower MSE or MAE values indicate better predictions. Compared with Table~\ref{tab: main point forecasting all}, we exclude the datasets used for pre-training several baseline models. \textcolor{red}{\textbf{Red}}: the best, \textcolor{blue}{\underline{Blue}}: the 2nd best.}
\label{tab: strict comparison point}
\resizebox{\textwidth}{!}{
\begin{tabular}{cc|cc|cc|cc|cc|cc|cc|cc|cc|cc|cc|cc}
\toprule
\multicolumn{2}{c|}{\multirow{2}{*}{Models}} & \multicolumn{2}{c|}{\textbf{FLAME}} & \multicolumn{2}{c|}{\textbf{FLAME}} & \multicolumn{2}{c|}{\textbf{FLAME}} & \multicolumn{2}{c|}{Sundial} & \multicolumn{2}{c|}{LightGTS} & \multicolumn{2}{c|}{ROSE} & \multicolumn{2}{c|}{Timer} & \multicolumn{2}{c|}{Chronos} & \multicolumn{2}{c|}{Time-MoE} & \multicolumn{2}{c|}{TinyTTM} & \multicolumn{2}{c}{MOIRAI} \\
~ & ~ & \multicolumn{2}{c|}{\textbf{Small}} & \multicolumn{2}{c|}{\textbf{Base}} & \multicolumn{2}{c|}{\textbf{Large}} & \multicolumn{2}{c|}{(2025)} & \multicolumn{2}{c|}{(2025)} & \multicolumn{2}{c|}{(2025)} & \multicolumn{2}{c|}{(2024)} & \multicolumn{2}{c|}{(2024)} & \multicolumn{2}{c|}{(2024)} & \multicolumn{2}{c|}{(2024)} & \multicolumn{2}{c}{(2024)} \\ \midrule
\multicolumn{2}{c|}{Metric} & \multicolumn{1}{c}{MSE} & \multicolumn{1}{c|}{MAE} & \multicolumn{1}{c}{MSE} & \multicolumn{1}{c|}{MAE} & \multicolumn{1}{c}{MSE} & \multicolumn{1}{c|}{MAE} & \multicolumn{1}{c}{MSE} & \multicolumn{1}{c|}{MAE} & \multicolumn{1}{c}{MSE} & \multicolumn{1}{c}{MAE} & \multicolumn{1}{c}{MSE} & \multicolumn{1}{c|}{MAE} & \multicolumn{1}{c}{MSE} & \multicolumn{1}{c|}{MAE} & \multicolumn{1}{c}{MSE} & \multicolumn{1}{c|}{MAE} & \multicolumn{1}{c}{MSE} & \multicolumn{1}{c|}{MAE} & \multicolumn{1}{c}{MSE} & \multicolumn{1}{c|}{MAE} & \multicolumn{1}{c}{MSE} & \multicolumn{1}{c}{MAE} \\
\midrule
\multirow[c]{5}{*}{\rotatebox{90}{ETTm1}} & 96 & 0.305 & 0.353 & 0.298 & 0.351 & 0.301 & 0.353 & \textcolor{red}{\textbf{0.280}} & \textcolor{red}{\textbf{0.334}} & 0.307 & 0.359 & 0.512 & 0.460 & 0.817 & 0.611 & 0.402 & 0.373 & \textcolor{blue}{\underline{0.281}} & \textcolor{blue}{\underline{0.341}} & 0.738 & 0.541 & 0.353 & 0.363 \\
 & 192 & 0.328 & 0.370 & 0.324 & 0.369 & 0.328 & 0.373 & \textcolor{blue}{\underline{0.321}} & \textcolor{blue}{\underline{0.366}} & 0.332 & 0.374 & 0.512 & 0.462 & 0.927 & 0.659 & 0.510 & 0.435 & \textcolor{red}{\textbf{0.305}} & \textcolor{red}{\textbf{0.358}} & 0.698 & 0.547 & 0.376 & 0.380 \\
 & 336 & \textcolor{blue}{\underline{0.349}} & \textcolor{red}{\textbf{0.384}} & \textcolor{red}{\textbf{0.346}} & \textcolor{blue}{\underline{0.386}} & 0.351 & 0.389 & 0.350 & 0.389 & 0.353 & 0.393 & 0.523 & 0.470 & 1.043 & 0.704 & 0.590 & 0.477 & 0.369 & 0.395 & 0.670 & 0.533 & 0.399 & 0.395 \\
 & 720 & 0.392 & 0.416 & \textcolor{red}{\textbf{0.385}} & \textcolor{blue}{\underline{0.414}} & 0.392 & 0.419 & 0.394 & 0.418 & \textcolor{blue}{\underline{0.388}} & \textcolor{red}{\textbf{0.386}} & 0.552 & 0.490 & 1.044 & 0.722 & 0.703 & 0.525 & 0.469 & 0.472 & 0.660 & 0.550 & 0.432 & 0.417 \\ \cmidrule{2-24}
 & Avg & 0.344 & 0.381 & \textcolor{blue}{\underline{0.338}} & 0.380 & 0.343 & 0.384 & \textcolor{red}{\textbf{0.336}} & \textcolor{red}{\textbf{0.377}} & 0.345 & \textcolor{blue}{\underline{0.378}} & 0.525 & 0.471 & 0.958 & 0.674 & 0.551 & 0.453 & 0.356 & 0.392 & 0.692 & 0.543 & 0.390 & 0.389 \\ \midrule
\multirow[c]{5}{*}{\rotatebox{90}{ETTm2}} & 96 & \textcolor{red}{\textbf{0.165}} & \textcolor{red}{\textbf{0.255}} & 0.171 & 0.262 & 0.186 & 0.272 & 0.170 & \textcolor{blue}{\underline{0.256}} & \textcolor{blue}{\underline{0.166}} & 0.260 & 0.224 & 0.309 & 0.225 & 0.300 & 0.192 & 0.263 & 0.198 & 0.288 & 0.226 & 0.309 & 0.189 & 0.260 \\
 & 192 & \textcolor{red}{\textbf{0.217}} & \textcolor{red}{\textbf{0.297}} & 0.227 & 0.302 & 0.247 & 0.315 & 0.229 & \textcolor{blue}{\underline{0.300}} & \textcolor{blue}{\underline{0.222}} & 0.300 & 0.266 & 0.333 & 0.286 & 0.339 & 0.256 & 0.308 & 0.235 & 0.312 & 0.311 & 0.360 & 0.247 & 0.300 \\
 & 336 & \textcolor{red}{\textbf{0.262}} & \textcolor{red}{\textbf{0.330}} & 0.280 & 0.340 & 0.300 & 0.352 & 0.281 & 0.337 & \textcolor{blue}{\underline{0.267}} & \textcolor{blue}{\underline{0.333}} & 0.310 & 0.358 & 0.335 & 0.369 & 0.315 & 0.346 & 0.293 & 0.348 & 0.350 & 0.383 & 0.295 & 0.334 \\
 & 720 & \textcolor{red}{\textbf{0.333}} & \textcolor{blue}{\underline{0.379}} & 0.348 & 0.387 & 0.357 & 0.394 & 0.351 & 0.387 & \textcolor{blue}{\underline{0.340}} & \textcolor{red}{\textbf{0.378}} & 0.395 & 0.407 & 0.414 & 0.416 & 0.409 & 0.405 & 0.427 & 0.428 & 0.446 & 0.435 & 0.372 & 0.386 \\ \cmidrule{2-24}
 & Avg & \textcolor{red}{\textbf{0.244}} & \textcolor{red}{\textbf{0.315}} & 0.257 & 0.323 & 0.273 & 0.333 & 0.258 & 0.320 & \textcolor{blue}{\underline{0.249}} & \textcolor{blue}{\underline{0.318}} & 0.299 & 0.352 & 0.315 & 0.356 & 0.293 & 0.331 & 0.288 & 0.344 & 0.333 & 0.372 & 0.276 & 0.320 \\ \midrule
\multirow[c]{5}{*}{\rotatebox{90}{ETTh1}} & 96 & 0.362 & 0.395 & 0.361 & 0.392 & 0.358 & 0.389 & \textcolor{red}{\textbf{0.348}} & \textcolor{blue}{\underline{0.385}} & 0.359 & 0.390 & 0.382 & 0.408 & 0.454 & 0.434 & 0.444 & 0.409 & \textcolor{blue}{\underline{0.349}} & \textcolor{red}{\textbf{0.379}} & 0.364 & 0.389 & 0.380 & 0.398 \\
 & 192 & 0.406 & 0.427 & 0.400 & 0.422 & 0.392 & 0.416 & 0.393 & 0.418 & \textcolor{blue}{\underline{0.390}} & \textcolor{blue}{\underline{0.409}} & 0.400 & 0.420 & 0.522 & 0.465 & 0.502 & 0.443 & 0.395 & 0.413 & \textcolor{red}{\textbf{0.386}} & \textcolor{red}{\textbf{0.407}} & 0.440 & 0.434 \\
 & 336 & 0.436 & 0.449 & 0.417 & 0.436 & 0.411 & 0.432 & 0.422 & 0.440 & 0.420 & 0.436 & \textcolor{blue}{\underline{0.404}} & \textcolor{blue}{\underline{0.426}} & 0.559 & 0.484 & 0.580 & 0.460 & 0.447 & 0.453 & \textcolor{red}{\textbf{0.404}} & \textcolor{red}{\textbf{0.422}} & 0.514 & 0.474 \\
 & 720 & 0.471 & 0.482 & 0.433 & 0.459 & 0.444 & 0.463 & 0.481 & 0.493 & 0.437 & 0.460 & \textcolor{red}{\textbf{0.420}} & \textcolor{red}{\textbf{0.447}} & 0.714 & 0.549 & 0.605 & 0.495 & 0.457 & 0.462 & \textcolor{blue}{\underline{0.424}} & \textcolor{blue}{\underline{0.448}} & 0.705 & 0.568 \\ \cmidrule{2-24}
 & Avg & 0.419 & 0.438 & 0.403 & 0.427 & \textcolor{blue}{\underline{0.401}} & 0.425 & 0.411 & 0.434 & 0.402 & \textcolor{blue}{\underline{0.424}} & 0.402 & 0.425 & 0.562 & 0.483 & 0.533 & 0.452 & 0.412 & 0.427 & \textcolor{red}{\textbf{0.395}} & \textcolor{red}{\textbf{0.417}} & 0.510 & 0.469 \\ \midrule
\multirow[c]{5}{*}{\rotatebox{90}{ETTh2}} & 96 & 0.278 & 0.339 & 0.295 & 0.345 & 0.300 & 0.350 & \textcolor{red}{\textbf{0.271}} & \textcolor{blue}{\underline{0.333}} & 0.278 & 0.342 & 0.298 & 0.362 & 0.316 & 0.359 & 0.306 & 0.338 & 0.292 & 0.352 & \textcolor{blue}{\underline{0.277}} & 0.335 & 0.287 & \textcolor{red}{\textbf{0.325}} \\
 & 192 & 0.336 & 0.381 & 0.353 & 0.388 & 0.351 & 0.390 & \textcolor{red}{\textbf{0.327}} & 0.376 & 0.345 & 0.383 & 0.336 & 0.385 & 0.374 & 0.398 & 0.396 & 0.394 & 0.347 & 0.379 & \textcolor{blue}{\underline{0.334}} & \textcolor{blue}{\underline{0.373}} & 0.347 & \textcolor{red}{\textbf{0.367}} \\
 & 336 & 0.373 & 0.410 & 0.373 & 0.409 & 0.374 & 0.411 & \textcolor{blue}{\underline{0.354}} & 0.402 & 0.394 & 0.416 & \textcolor{red}{\textbf{0.353}} & \textcolor{blue}{\underline{0.399}} & 0.381 & 0.410 & 0.423 & 0.417 & 0.406 & 0.419 & 0.362 & 0.402 & 0.377 & \textcolor{red}{\textbf{0.393}} \\
 & 720 & 0.411 & 0.441 & 0.413 & 0.440 & 0.416 & 0.444 & \textcolor{red}{\textbf{0.381}} & 0.435 & 0.430 & 0.445 & \textcolor{blue}{\underline{0.395}} & \textcolor{blue}{\underline{0.432}} & 0.408 & 0.434 & 0.442 & 0.439 & 0.439 & 0.447 & 0.408 & 0.444 & 0.404 & \textcolor{red}{\textbf{0.421}} \\ \cmidrule{2-24}
 & Avg & 0.350 & 0.393 & 0.359 & 0.396 & 0.360 & 0.399 & \textcolor{red}{\textbf{0.333}} & \textcolor{blue}{\underline{0.387}} & 0.362 & 0.397 & 0.346 & 0.395 & 0.370 & 0.400 & 0.392 & 0.397 & 0.371 & 0.399 & \textcolor{blue}{\underline{0.345}} & 0.389 & 0.354 & \textcolor{red}{\textbf{0.377}} \\ \midrule
\multirow[c]{5}{*}{\rotatebox{90}{Weather}} & 96 & \textcolor{red}{\textbf{0.148}} & \textcolor{red}{\textbf{0.204}} & \textcolor{blue}{\underline{0.151}} & 0.209 & 0.152 & 0.211 & 0.157 & \textcolor{blue}{\underline{0.205}} & 0.151 & 0.211 & 0.200 & 0.260 & 0.190 & 0.236 & 0.186 & 0.208 & 0.157 & 0.211 & 0.183 & 0.242 & 0.177 & 0.208 \\
 & 192 & \textcolor{red}{\textbf{0.189}} & \textcolor{red}{\textbf{0.243}} & 0.194 & 0.250 & 0.199 & 0.254 & 0.205 & 0.251 & \textcolor{blue}{\underline{0.191}} & \textcolor{blue}{\underline{0.248}} & 0.239 & 0.288 & 0.261 & 0.293 & 0.238 & 0.258 & 0.208 & 0.256 & 0.229 & 0.285 & 0.219 & 0.249 \\
 & 336 & \textcolor{red}{\textbf{0.233}} & \textcolor{red}{\textbf{0.278}} & 0.241 & 0.286 & 0.250 & 0.293 & 0.253 & 0.289 & \textcolor{blue}{\underline{0.236}} & \textcolor{blue}{\underline{0.284}} & 0.279 & 0.315 & 0.332 & 0.340 & 0.313 & 0.353 & 0.255 & 0.290 & 0.289 & 0.330 & 0.277 & 0.292 \\
 & 720 & \textcolor{red}{\textbf{0.289}} & \textcolor{red}{\textbf{0.320}} & 0.301 & 0.330 & 0.307 & 0.333 & 0.320 & 0.336 & \textcolor{blue}{\underline{0.299}} & \textcolor{blue}{\underline{0.323}} & 0.340 & 0.357 & 0.385 & 0.381 & 0.416 & 0.415 & 0.405 & 0.397 & 0.359 & 0.370 & 0.365 & 0.350 \\ \cmidrule{2-24}
 & Avg & \textcolor{red}{\textbf{0.215}} & \textcolor{red}{\textbf{0.261}} & 0.222 & 0.269 & 0.227 & 0.273 & 0.234 & 0.270 & \textcolor{blue}{\underline{0.219}} & \textcolor{blue}{\underline{0.267}} & 0.265 & 0.305 & 0.292 & 0.313 & 0.288 & 0.309 & 0.256 & 0.289 & 0.265 & 0.307 & 0.260 & 0.275 \\ \midrule
\multirow[c]{5}{*}{\rotatebox{90}{Solar}} & 96 & 0.185 & 0.263 & \textcolor{blue}{\underline{0.167}} & 0.259 & \textcolor{red}{\textbf{0.153}} & \textcolor{blue}{\underline{0.255}} & 0.204 & \textcolor{red}{\textbf{0.230}} & 0.209 & 0.309 & 0.524 & 0.557 & 0.591 & 0.504 & 0.373 & 0.304 & 0.304 & 0.345 & 0.863 & 0.664 & 0.682 & 0.688 \\
 & 192 & 0.201 & 0.276 & \textcolor{blue}{\underline{0.180}} & 0.273 & \textcolor{red}{\textbf{0.165}} & \textcolor{blue}{\underline{0.269}} & 0.221 & \textcolor{red}{\textbf{0.248}} & 0.220 & 0.315 & 0.507 & 0.550 & 0.689 & 0.567 & 0.363 & 0.303 & 0.309 & 0.342 & 0.823 & 0.695 & 0.694 & 0.695 \\
 & 336 & 0.212 & 0.283 & \textcolor{blue}{\underline{0.190}} & 0.282 & \textcolor{red}{\textbf{0.174}} & \textcolor{blue}{\underline{0.278}} & 0.225 & \textcolor{red}{\textbf{0.260}} & 0.228 & 0.314 & 0.508 & 0.553 & 0.831 & 0.636 & 0.391 & 0.319 & 0.433 & 0.450 & 0.835 & 0.741 & 0.719 & 0.706 \\
 & 720 & 0.213 & 0.291 & \textcolor{blue}{\underline{0.198}} & 0.295 & \textcolor{red}{\textbf{0.179}} & 0.287 & 0.233 & \textcolor{red}{\textbf{0.272}} & 0.218 & \textcolor{blue}{\underline{0.283}} & 0.479 & 0.534 & 0.972 & 0.710 & 0.444 & 0.349 & 0.599 & 0.576 & 0.738 & 0.738 & 0.759 & 0.725 \\ \cmidrule{2-24}
 & Avg & 0.203 & 0.278 & \textcolor{blue}{\underline{0.184}} & 0.277 & \textcolor{red}{\textbf{0.168}} & \textcolor{blue}{\underline{0.272}} & 0.221 & \textcolor{red}{\textbf{0.252}} & 0.219 & 0.305 & 0.505 & 0.549 & 0.771 & 0.604 & 0.393 & 0.319 & 0.411 & 0.428 & 0.815 & 0.710 & 0.714 & 0.704 \\ \midrule

 \rowc
\multicolumn{2}{c|}{1\textsuperscript{st} Count} & \textcolor{red}{\textbf{8}} & \textcolor{red}{\textbf{8}} & 2 &0 & 4 & 0 & \textcolor{blue}{\underline{5}} & \textcolor{blue}{\underline{5}} & 0 & 2 & 2 & 1 & 0 & 0 & 0 & 0 & 1 & 2 & 2 & 2 & 0 & 4 \\
\rowc
\multicolumn{2}{c|}{2\textsuperscript{nd} Count} & 1 & 1 & \textcolor{blue}{\underline{5}} & 2 & 0 & 3 & 2 & \textcolor{red}{\textbf{6}} & \textcolor{red}{\textbf{9}} & \textcolor{blue}{\underline{6}} & 2 & 3 & 0 & 0 & 0 & 0 & 2 & 1 & 4 & 2 & 0 & 0 \\
\bottomrule
\end{tabular}
}
\end{table*}

\begin{table*}[!htbp]
\caption{Full results of zero-shot probabilistic forecasting experiments on datasets from ProbTS. Lower CRPS or NMAE values indicate better predictions. Compared with Table~\ref{tab: main prob forecasting all}, we exclude the datasets used for pre-training several baseline models. \textcolor{red}{\textbf{Red}}: the best, \textcolor{blue}{\underline{Blue}}: the 2nd best.}
\label{tab: strict comparison prob}
\resizebox{1\linewidth}{!}{
\begin{tabular}{cc|cc|cc|cc|cc|cc|cc|cc|cc|cc|cc|cc}
\toprule
\multicolumn{2}{c|}{\multirow{2}{*}{Models}} & \multicolumn{2}{c|}{\textbf{FLAME}} & \multicolumn{2}{c|}{\textbf{FLAME}} & \multicolumn{2}{c|}{\textbf{FLAME}} & \multicolumn{2}{c|}{\textbf{Sundial}} & \multicolumn{2}{c|}{\textbf{Chronos}} & \multicolumn{2}{c|}{\textbf{MOIRAI}} & \multicolumn{2}{c|}{\textbf{Lag-Llama}} & \multicolumn{2}{c|}{\textbf{TSDiff}} & \multicolumn{2}{c|}{\textbf{CSDI}} & \multicolumn{2}{c|}{\textbf{TimeGrad}} & \multicolumn{2}{c}{\textbf{GRU NVP}} \\    
\multicolumn{2}{c|}{~} & \multicolumn{2}{c|}{\textbf{Small}} & \multicolumn{2}{c|}{\textbf{Base}} &\multicolumn{2}{c|}{\textbf{Large}} & \multicolumn{2}{c|}{(2025)} & \multicolumn{2}{c|}{(2024)} & \multicolumn{2}{c|}{(2024)} & \multicolumn{2}{c|}{(2023)} & \multicolumn{2}{c|}{(2023)} & \multicolumn{2}{c|}{(2022)} & \multicolumn{2}{c|}{(2022)} & \multicolumn{2}{c}{(2021)}\\
\midrule
\multicolumn{2}{c|}{Metrics} & CRPS & NMAE & CRPS & NMAE & CRPS & NMAE & CRPS & NMAE & CRPS & NMAE & CRPS & NMAE & CRPS & NMAE & CRPS & NMAE & CRPS & NMAE & CRPS & NMAE & CRPS & NMAE \\  
\midrule
 & 96 & 0.294 & 0.368 & 0.326 & 0.417 & 0.315 & 0.407 & \textcolor{blue}{\underline{ 0.253}} & \textcolor{blue}{\underline{ 0.308}} & 0.360 & 0.422 & 0.464 & 0.522 & 0.354 & 0.402 & 0.344 & 0.441 & \textcolor{red}{ \textbf{0.236}} & \textcolor{red}{ \textbf{0.308}} & 0.522 & 0.645 & 0.383 & 0.488 \\
 & 192 & 0.332 & 0.420 & 0.363 & 0.469 & 0.337 & 0.426 & \textcolor{red}{ \textbf{0.279}} & \textcolor{red}{ \textbf{0.337}} & 0.404 & 0.450 & 0.467 & 0.531 & 0.368 & 0.415 & 0.345 & 0.441 & \textcolor{blue}{\underline{ 0.291}} & \textcolor{blue}{\underline{ 0.377}} & 0.603 & 0.748 & 0.396 & 0.514 \\
 & 336 & 0.341 & 0.427 & 0.393 & 0.511 & 0.351 & 0.447 & \textcolor{red}{ \textbf{0.291}} & \textcolor{red}{ \textbf{0.350}} & 0.425 & 0.456 & 0.524 & 0.558 & 0.387 & 0.436 & 0.462 & 0.571 & \textcolor{blue}{\underline{ 0.322}} & \textcolor{blue}{\underline{ 0.419}} & 0.601 & 0.759 & 0.486 & 0.630 \\
 & 720 & \textcolor{blue}{\underline{ 0.364}} & \textcolor{blue}{\underline{ 0.454}} & 0.477 & 0.622 & 0.383 & 0.476 & \textcolor{red}{ \textbf{0.318}} & \textcolor{red}{ \textbf{0.380}} & 0.461 & 0.478 & 0.514 & 0.535 & 0.403 & 0.466 & 0.478 & 0.622 & 0.448 & 0.578 & 0.621 & 0.793 & 0.546 & 0.707 \\
 \cmidrule{2-24}
\multirow{-5}{*}{\rotatebox{90}{ETTm1}} & avg & 0.333 & \textcolor{blue}{\underline{ 0.417}} & 0.390 & 0.505 & 0.347 & 0.439 & \textcolor{red}{ \textbf{0.285}} & \textcolor{red}{ \textbf{0.344}} & 0.413 & 0.451 & 0.492 & 0.536 & 0.378 & 0.430 & 0.407 & 0.519 & \textcolor{blue}{\underline{ 0.324}} & 0.421 & 0.587 & 0.736 & 0.453 & 0.585 \\
\midrule
 & 96 & \textcolor{blue}{\underline{ 0.120}} & \textcolor{blue}{\underline{ 0.153}} & 0.121 & 0.153 & 0.124 & 0.155 & 0.128 & 0.153 & 0.134 & 0.158 & 0.176 & 0.186 & 0.163 & 0.192 & 0.175 & 0.224 & \textcolor{red}{ \textbf{0.115}} & \textcolor{red}{ \textbf{0.146}} & 0.427 & 0.525 & 0.319 & 0.413 \\
 & 192 & \textcolor{blue}{\underline{ 0.146}} & 0.182 & 0.147 & 0.184 & \textcolor{red}{ \textbf{0.143}} & \textcolor{red}{ \textbf{0.175}} & 0.150 & \textcolor{blue}{\underline{ 0.177}} & 0.163 & 0.183 & 0.197 & 0.207 & 0.181 & 0.207 & 0.255 & 0.316 & 0.147 & 0.189 & 0.424 & 0.530 & 0.326 & 0.427 \\
 & 336 & \textcolor{blue}{\underline{ 0.163}} & 0.202 & 0.164 & 0.205 & \textcolor{red}{ \textbf{0.158}} & \textcolor{red}{ \textbf{0.193}} & 0.167 & \textcolor{blue}{\underline{ 0.195}} & 0.190 & 0.204 & 0.229 & 0.227 & 0.206 & 0.229 & 0.328 & 0.397 & 0.190 & 0.248 & 0.469 & 0.566 & 0.449 & 0.580 \\
 & 720 & \textcolor{blue}{\underline{ 0.185}} & 0.223 & 0.186 & 0.231 & \textcolor{red}{ \textbf{0.184}} & \textcolor{blue}{\underline{ 0.219}} & 0.189 & \textcolor{red}{ \textbf{0.217}} & 0.223 & 0.230 & 0.321 & 0.258 & 0.227 & 0.249 & 0.344 & 0.416 & 0.239 & 0.306 & 0.470 & 0.561 & 0.561 & 0.749 \\
 \cmidrule{2-24}
\multirow{-5}{*}{\rotatebox{90}{ETTm2}} & avg & \textcolor{blue}{\underline{ 0.154}} & 0.190 & 0.155 & 0.193 & \textcolor{red}{ \textbf{0.152}} & \textcolor{red}{ \textbf{0.186}} & 0.158 & \textcolor{blue}{\underline{ 0.186}} & 0.178 & 0.194 & 0.231 & 0.220 & 0.194 & 0.219 & 0.276 & 0.338 & 0.173 & 0.222 & 0.448 & 0.546 & 0.414 & 0.542 \\
\midrule
 & 96 & 0.265 & 0.322 & 0.261 & 0.319 & \textcolor{red}{ \textbf{0.259}} & \textcolor{blue}{\underline{ 0.310}} & \textcolor{blue}{\underline{ 0.260}} & \textcolor{red}{ \textbf{0.307}} & 0.293 & 0.341 & 0.469 & 0.488 & 0.297 & 0.340 & 0.395 & 0.510 & 0.437 & 0.557 & 0.455 & 0.585 & 0.379 & 0.481 \\
 & 192 & 0.288 & \textcolor{blue}{\underline{ 0.346}} & \textcolor{blue}{\underline{ 0.282}} & 0.351 & \textcolor{red}{ \textbf{0.278}} & \textcolor{red}{ \textbf{0.342}} & 0.297 & 0.347 & 0.348 & 0.384 & 0.496 & 0.492 & 0.312 & 0.356 & 0.467 & 0.596 & 0.496 & 0.625 & 0.516 & 0.680 & 0.425 & 0.531 \\
 & 336 & 0.300 & 0.371 & \textcolor{blue}{\underline{ 0.297}} & 0.368 & \textcolor{red}{ \textbf{0.290}} & \textcolor{red}{ \textbf{0.362}} & 0.314 & \textcolor{blue}{\underline{ 0.365}} & 0.384 & 0.409 & 0.503 & 0.489 & 0.326 & 0.370 & 0.450 & 0.581 & 0.454 & 0.574 & 0.512 & 0.666 & 0.458 & 0.580 \\
 & 720 & 0.320 & 0.402 & \textcolor{blue}{\underline{ 0.312}} & 0.386 & \textcolor{red}{ \textbf{0.303}} & \textcolor{blue}{\underline{ 0.367}} & 0.312 & \textcolor{red}{ \textbf{0.366}} & 0.413 & 0.425 & 0.526 & 0.508 & 0.340 & 0.405 & 0.516 & 0.657 & 0.528 & 0.657 & 0.523 & 0.672 & 0.502 & 0.643 \\
  \cmidrule{2-24}
\multirow{-5}{*}{\rotatebox{90}{ETTh1}} & avg & 0.293 & 0.360 & \textcolor{blue}{\underline{ 0.288}} & 0.356 & \textcolor{red}{ \textbf{0.283}} & \textcolor{red}{ \textbf{0.345}} & 0.296 & \textcolor{blue}{\underline{ 0.346}} & 0.360 & 0.390 & 0.498 & 0.494 & 0.319 & 0.368 & 0.457 & 0.586 & 0.479 & 0.603 & 0.502 & 0.651 & 0.441 & 0.559 \\
\midrule
 & 96 & \textcolor{red}{ \textbf{0.139}} & \textcolor{red}{ \textbf{0.177}} & 0.145 & \textcolor{blue}{\underline{ 0.185}} & \textcolor{blue}{\underline{ 0.144}} & 0.186 & 0.157 & 0.186 & 0.163 & 0.191 & 0.212 & 0.238 & 0.178 & 0.204 & 0.336 & 0.421 & 0.164 & 0.214 & 0.358 & 0.448 & 0.432 & 0.548 \\
 & 192 & \textcolor{red}{ \textbf{0.152}} & \textcolor{red}{ \textbf{0.195}} & 0.161 & 0.207 & \textcolor{blue}{\underline{ 0.159}} & \textcolor{blue}{\underline{ 0.204}} & 0.183 & 0.213 & 0.195 & 0.218 & 0.229 & 0.250 & 0.193 & 0.217 & 0.265 & 0.339 & 0.226 & 0.294 & 0.457 & 0.575 & 0.625 & 0.766 \\
 & 336 & \textcolor{red}{ \textbf{0.170}} & \textcolor{red}{ \textbf{0.216}} & 0.174 & 0.224 & \textcolor{blue}{\underline{ 0.172}} & \textcolor{blue}{\underline{ 0.217}} & 0.203 & 0.232 & 0.226 & 0.241 & 0.253 & 0.263 & 0.211 & 0.230 & 0.350 & 0.427 & 0.274 & 0.353 & 0.481 & 0.606 & 0.793 & 0.942 \\
 & 720 & \textcolor{blue}{\underline{ 0.183}} & \textcolor{blue}{\underline{ 0.231}} & 0.191 & 0.241 & \textcolor{red}{ \textbf{0.182}} & \textcolor{red}{ \textbf{0.228}} & 0.203 & 0.233 & 0.255 & 0.263 & 0.279 & 0.273 & 0.222 & 0.238 & 0.406 & 0.482 & 0.302 & 0.382 & 0.445 & 0.550 & 0.539 & 0.688 \\
  \cmidrule{2-24}
\multirow{-5}{*}{\rotatebox{90}{ETTh2}} & avg & \textcolor{red}{ \textbf{0.161}} & \textcolor{red}{ \textbf{0.205}} & 0.168 & 0.214 & \textcolor{blue}{\underline{ 0.164}} & \textcolor{blue}{\underline{ 0.209}} & 0.186 & 0.216 & 0.210 & 0.228 & 0.243 & 0.256 & 0.201 & 0.222 & 0.339 & 0.417 & 0.242 & 0.311 & 0.435 & 0.545 & 0.597 & 0.736 \\
\midrule
 & 96 & 0.071 & 0.088 & 0.071 & \textcolor{blue}{\underline{ 0.085}} & \textcolor{red}{ \textbf{0.067}} & \textcolor{red}{ \textbf{0.083}} & 0.078 & 0.091 & 0.136 & 0.155 & 0.161 & 0.141 & 0.085 & 0.094 & 0.104 & 0.113 & \textcolor{blue}{\underline{ 0.068}} & 0.087 & 0.130 & 0.164 & 0.116 & 0.145 \\
 & 192 & 0.089 & 0.113 & 0.081 & 0.100 & \textcolor{red}{ \textbf{0.068}} & \textcolor{red}{ \textbf{0.082}} & 0.083 & 0.097 & 0.145 & 0.165 & 0.173 & 0.139 & 0.094 & 0.102 & 0.134 & 0.144 & \textcolor{blue}{\underline{ 0.068}} & \textcolor{blue}{\underline{ 0.086}} & 0.127 & 0.158 & 0.122 & 0.147 \\
 & 336 & 0.094 & 0.121 & 0.090 & 0.110 & \textcolor{red}{ \textbf{0.079}} & \textcolor{red}{ \textbf{0.094}} & 0.090 & 0.106 & 0.136 & 0.151 & 0.149 & 0.134 & 0.098 & 0.109 & 0.137 & 0.138 & \textcolor{blue}{\underline{ 0.083}} & \textcolor{blue}{\underline{ 0.098}} & 0.130 & 0.162 & 0.128 & 0.160 \\
 & 720 & 0.103 & 0.132 & 0.106 & 0.125 & \textcolor{red}{ \textbf{0.083}} & \textcolor{red}{ \textbf{0.100}} & 0.096 & 0.113 & 0.151 & 0.161 & 0.233 & 0.155 & 0.107 & 0.120 & 0.152 & 0.141 & \textcolor{blue}{\underline{ 0.087}} & \textcolor{blue}{\underline{ 0.102}} & 0.113 & 0.136 & 0.110 & 0.135 \\
  \cmidrule{2-24}
\multirow{-5}{*}{\rotatebox{90}{Weather}} & avg & 0.089 & 0.114 & 0.087 & 0.105 & \textcolor{red}{ \textbf{0.074}} & \textcolor{red}{ \textbf{0.090}} & 0.087 & 0.102 & 0.142 & 0.158 & 0.179 & 0.143 & 0.096 & 0.106 & 0.132 & 0.134 & \textcolor{blue}{\underline{ 0.077}} & \textcolor{blue}{\underline{ 0.093}} & 0.125 & 0.155 & 0.119 & 0.147 \\
\midrule
 & 96 & 0.023 & 0.028 & \textcolor{blue}{\underline{ 0.021}} & 0.027 & \textcolor{red}{ \textbf{0.020}} & \textcolor{blue}{\underline{ 0.026}} & 0.024 & 0.027 & 0.021 & \textcolor{red}{ \textbf{0.025}} & 0.025 & 0.030 & 0.042 & 0.051 & 0.079 & 0.090 & 0.028 & 0.036 & 0.068 & 0.079 & 0.071 & 0.091 \\
 & 192 & 0.032 & 0.037 & \textcolor{red}{ \textbf{0.030}} & 0.036 & \textcolor{blue}{\underline{ 0.031}} & \textcolor{red}{ \textbf{0.035}} & 0.034 & 0.038 & 0.032 & \textcolor{blue}{\underline{ 0.036}} & 0.034 & 0.039 & 0.047 & 0.058 & 0.093 & 0.106 & 0.045 & 0.058 & 0.087 & 0.100 & 0.068 & 0.087 \\
 & 336 & \textcolor{blue}{\underline{ 0.044}} & 0.049 & \textcolor{red}{ \textbf{0.041}} & \textcolor{blue}{\underline{ 0.047}} & 0.044 & \textcolor{red}{ \textbf{0.046}} & 0.046 & 0.050 & 0.045 & 0.048 & 0.047 & 0.052 & 0.061 & 0.073 & 0.081 & 0.106 & 0.060 & 0.076 & 0.074 & 0.086 & 0.072 & 0.091 \\
 & 720 & 0.074 & \textcolor{blue}{\underline{ 0.079}} & \textcolor{red}{ \textbf{0.069}} & \textcolor{red}{ \textbf{0.077}} & \textcolor{blue}{\underline{ 0.073}} & 0.079 & 0.076 & 0.079 & 0.078 & 0.080 & 0.073 & 0.081 & 0.078 & 0.094 & 0.082 & 0.142 & 0.143 & 0.173 & 0.099 & 0.113 & 0.079 & 0.103 \\
  \cmidrule{2-24}
\multirow{-5}{*}{\rotatebox{90}{Exchange}} & avg & 0.043 & 0.048 & \textcolor{red}{ \textbf{0.040}} & \textcolor{blue}{\underline{ 0.047}} & \textcolor{blue}{\underline{ 0.042}} & \textcolor{red}{ \textbf{0.047}} & 0.045 & 0.049 & 0.044 & 0.047 & 0.045 & 0.050 & 0.057 & 0.069 & 0.084 & 0.111 & 0.069 & 0.086 & 0.082 & 0.095 & 0.073 & 0.093 \\
\midrule
 & 24 & 0.135 & 0.166 & \textcolor{blue}{\underline{ 0.111}} & 0.143 & 0.136 & 0.168 & \textcolor{red}{ \textbf{0.108}} & \textcolor{red}{ \textbf{0.123}} & 0.120 & \textcolor{blue}{\underline{ 0.139}} & 0.150 & 0.196 & 0.135 & 0.173 & 0.228 & 0.242 & 0.250 & 0.263 & 0.275 & 0.296 & 0.257 & 0.283 \\
 & 36 & 0.164 & 0.200 & \textcolor{red}{ \textbf{0.119}} & \textcolor{red}{ \textbf{0.153}} & \textcolor{blue}{\underline{ 0.145}} & 0.179 & 0.152 & \textcolor{blue}{\underline{ 0.169}} & 0.179 & 0.205 & 0.171 & 0.222 & 0.163 & 0.227 & 0.235 & 0.246 & 0.285 & 0.298 & 0.272 & 0.298 & 0.281 & 0.307 \\
 & 48 & 0.176 & 0.216 & \textcolor{red}{ \textbf{0.139}} & \textcolor{red}{ \textbf{0.178}} & 0.164 & 0.200 & 0.164 & 0.184 & 0.186 & 0.215 & \textcolor{blue}{\underline{ 0.151}} & \textcolor{blue}{\underline{ 0.184}} & 0.171 & 0.233 & 0.265 & 0.275 & 0.285 & 0.301 & 0.295 & 0.320 & 0.288 & 0.314 \\
 & 60 & 0.171 & 0.211 & \textcolor{red}{ \textbf{0.149}} & \textcolor{blue}{\underline{ 0.189}} & 0.169 & 0.204 & 0.168 & 0.190 & 0.196 & 0.228 & 0.163 & \textcolor{red}{ \textbf{0.188}} & \textcolor{blue}{\underline{ 0.156}} & 0.211 & 0.263 & 0.272 & 0.283 & 0.299 & 0.295 & 0.325 & 0.307 & 0.333 \\
  \cmidrule{2-24}
\multirow{-5}{*}{\rotatebox{90}{ILI}} & avg & 0.162 & 0.198 & \textcolor{red}{ \textbf{0.130}} & \textcolor{red}{ \textbf{0.166}} & 0.154 & 0.188 & \textcolor{blue}{\underline{ 0.148}} & \textcolor{blue}{\underline{ 0.166}} & 0.170 & 0.197 & 0.159 & 0.197 & 0.156 & 0.211 & 0.248 & 0.259 & 0.276 & 0.290 & 0.284 & 0.310 & 0.283 & 0.309 \\
\midrule
\rowc
\multicolumn{2}{c|}{1\textsuperscript{st} Count} &3 &3 &\textcolor{blue}{\underline{6}} &3 &\textcolor{red}{\textbf{13}} &\textcolor{red}{\textbf{11}} &4 &\textcolor{blue}{\underline{7}} &0 &1 &0 &1 &0 &0 &0 &0 &2 &2 &0 &0 &0 &0 \\      \rowc
\multicolumn{2}{c|}{2\textsuperscript{nd} Count} & \textcolor{red}{\textbf{7}} & 5 & 5 & 4 & \textcolor{blue}{\underline{6}} & \textcolor{red}{\textbf{6}} & 2 & \textcolor{blue}{\underline{5}} & 0 & 2 & 1 & 1 & 1 & 0 & 0 & 0 & 6 & 5 & 0 & 0 & 0 & 0\\ 
\bottomrule
        
\end{tabular}}
\end{table*}

\subsection{Full-shot and few-shot results}
We further compare FLAME Small with end-to-end supervised models. We provide all the results of deterministic forecasting in Table~\ref{tab: Full-shot result 1}, comparing FLAME Small of different training budgets with most advanced end-to-end deterministic forecasting models.

We also provide all the results of probabilistic forecasting in Table~\ref{tab: Full-shot result 2}, comparing FLAME Small of different training budgets with most advanced end-to-end probabilistic forecasting models.

\renewcommand{\arraystretch}{1.2}
\begin{table*}[!htbp]
\centering
\caption{Probabilistic forecasting results. Full-shot, 10\% few-shot and zero-shot results of FLAME Small vs. the most advanced end-to-end supervised models on commonly-used datasets in ProbTS. Lower CRPS or NMAE values indicate better predictions. \textcolor{red}{\textbf{Red}}: the best, \textcolor{blue}{\underline{Blue}}: the second best.}
\label{tab: Full-shot result 2}
\resizebox{\textwidth}{!}{
\begin{tabular}{cc|cc|cc|cc|cc|cc|cc|cc|cc|cc}
\toprule
\multicolumn{2}{c|}{\multirow{2}{*}{Models}} & \multicolumn{2}{c|}{FLAME Small} & \multicolumn{2}{c|}{FLAME Small} & \multicolumn{2}{c|}{FLAME Small} & \multicolumn{2}{c|}{NSDIFF} & \multicolumn{2}{c|}{$K^2$VAE} & \multicolumn{2}{c|}{TSDiff} & \multicolumn{2}{c|}{CSDI} & \multicolumn{2}{c|}{TimeGrad} & \multicolumn{2}{c}{GRU NVP} \\
~ & ~ & \multicolumn{2}{c|}{(Zero-shot)} & \multicolumn{2}{c|}{(10\% few-shot)} & \multicolumn{2}{c|}{(Full-shot)} & \multicolumn{2}{c|}{(2025)} & \multicolumn{2}{c|}{(2025)} & \multicolumn{2}{c|}{(2023)} & \multicolumn{2}{c|}{(2022)} & \multicolumn{2}{c|}{(2022)} & \multicolumn{2}{c}{(2021)} \\\midrule
\multicolumn{2}{c|}{Metric} & \multicolumn{1}{c}{CRPS} & \multicolumn{1}{c|}{NMAE} & \multicolumn{1}{c}{CRPS} & \multicolumn{1}{c|}{NMAE} & \multicolumn{1}{c}{CRPS} & \multicolumn{1}{c|}{NMAE} & \multicolumn{1}{c}{CRPS} & \multicolumn{1}{c|}{NMAE} & \multicolumn{1}{c}{CRPS} & \multicolumn{1}{c|}{NMAE} & \multicolumn{1}{c}{CRPS} & \multicolumn{1}{c|}{NMAE} & \multicolumn{1}{c}{CRPS} & \multicolumn{1}{c|}{NMAE} & \multicolumn{1}{c}{CRPS} & \multicolumn{1}{c|}{NMAE} & \multicolumn{1}{c}{CRPS} & \multicolumn{1}{c}{NMAE} \\
\midrule
\multirow[c]{5}{*}{\rotatebox{90}{ETTm1}} & 96 & 0.294 & 0.368 & \textcolor{blue}{\underline{0.233}} & \textcolor{blue}{\underline{0.287}} & \textcolor{red}{\textbf{0.211}} & \textcolor{red}{\textbf{0.275}} & 0.295 & 0.364 & 0.235 & 0.294 & 0.344 & 0.441 & 0.236 & 0.308 & 0.522 & 0.645 & 0.383 & 0.488 \\
 & 192 & 0.332 & 0.420 & \textcolor{blue}{\underline{0.259}} & \textcolor{blue}{\underline{0.317}} & \textcolor{red}{\textbf{0.232}} & \textcolor{red}{\textbf{0.298}} & 0.285 & 0.350 & 0.261 & 0.330 & 0.345 & 0.441 & 0.291 & 0.377 & 0.603 & 0.748 & 0.396 & 0.514 \\
 & 336 & 0.341 & 0.427 & 0.292 & \textcolor{red}{\textbf{0.351}} & \textcolor{red}{\textbf{0.279}} & \textcolor{blue}{\underline{0.355}} & \textcolor{blue}{\underline{0.289}} & 0.357 & 0.304 & 0.386 & 0.462 & 0.571 & 0.322 & 0.419 & 0.601 & 0.759 & 0.486 & 0.630 \\
 & 720 & 0.364 & 0.454 & 0.304 & 0.392 & \textcolor{red}{\textbf{0.282}} & \textcolor{red}{\textbf{0.358}} & 0.411 & 0.500 & \textcolor{blue}{\underline{0.302}} & \textcolor{blue}{\underline{0.387}} & 0.478 & 0.622 & 0.448 & 0.578 & 0.621 & 0.793 & 0.546 & 0.707 \\ \cmidrule{2-20}
 & Avg & 0.333 & 0.417 & \textcolor{blue}{\underline{0.272}} & \textcolor{blue}{\underline{0.337}} & \textcolor{red}{\textbf{0.251}} & \textcolor{red}{\textbf{0.322}} & 0.320 & 0.393 & 0.276 & 0.349 & 0.407 & 0.519 & 0.324 & 0.421 & 0.587 & 0.736 & 0.453 & 0.585 \\ \midrule
\multirow[c]{5}{*}{\rotatebox{90}{ETTm2}} & 96 & 0.120 & 0.153 & 0.117 & \textcolor{blue}{\underline{0.144}} & \textcolor{red}{\textbf{0.114}} & \textcolor{red}{\textbf{0.141}} & 0.145 & 0.176 & 0.128 & 0.148 & 0.175 & 0.224 & \textcolor{blue}{\underline{0.115}} & 0.146 & 0.427 & 0.525 & 0.319 & 0.413 \\
 & 192 & 0.146 & 0.182 & \textcolor{blue}{\underline{0.145}} & 0.178 & \textcolor{red}{\textbf{0.138}} & \textcolor{blue}{\underline{0.166}} & 0.162 & 0.196 & 0.161 & \textcolor{red}{\textbf{0.163}} & 0.255 & 0.316 & 0.147 & 0.189 & 0.424 & 0.530 & 0.326 & 0.427 \\
 & 336 & 0.163 & 0.202 & \textcolor{blue}{\underline{0.160}} & 0.199 & \textcolor{red}{\textbf{0.151}} & \textcolor{red}{\textbf{0.182}} & 0.172 & 0.209 & 0.180 & \textcolor{blue}{\underline{0.183}} & 0.328 & 0.397 & 0.190 & 0.248 & 0.469 & 0.566 & 0.449 & 0.580 \\
 & 720 & 0.185 & \textcolor{blue}{\underline{0.223}} & \textcolor{blue}{\underline{0.181}} & 0.231 & \textcolor{red}{\textbf{0.170}} & \textcolor{red}{\textbf{0.211}} & 0.364 & 0.428 & 0.224 & 0.224 & 0.344 & 0.416 & 0.239 & 0.306 & 0.470 & 0.561 & 0.561 & 0.749 \\ \cmidrule{2-20}
 & Avg & 0.154 & 0.190 & \textcolor{blue}{\underline{0.151}} & 0.188 & \textcolor{red}{\textbf{0.143}} & \textcolor{red}{\textbf{0.175}} & 0.211 & 0.252 & 0.173 & \textcolor{blue}{\underline{0.180}} & 0.276 & 0.338 & 0.173 & 0.222 & 0.448 & 0.546 & 0.414 & 0.542 \\ \midrule
\multirow[c]{5}{*}{\rotatebox{90}{ETTh1}} & 96 & 0.265 & 0.322 & \textcolor{blue}{\underline{0.254}} & \textcolor{blue}{\underline{0.311}} & \textcolor{red}{\textbf{0.247}} & \textcolor{red}{\textbf{0.301}} & 0.301 & 0.380 & 0.280 & 0.358 & 0.395 & 0.510 & 0.437 & 0.557 & 0.455 & 0.585 & 0.379 & 0.481 \\
 & 192 & 0.288 & 0.346 & \textcolor{blue}{\underline{0.272}} & \textcolor{blue}{\underline{0.336}} & \textcolor{red}{\textbf{0.268}} & \textcolor{red}{\textbf{0.322}} & 0.309 & 0.386 & 0.304 & 0.392 & 0.467 & 0.596 & 0.496 & 0.625 & 0.516 & 0.680 & 0.425 & 0.531 \\
 & 336 & 0.300 & 0.371 & \textcolor{blue}{\underline{0.297}} & \textcolor{blue}{\underline{0.369}} & \textcolor{red}{\textbf{0.288}} & \textcolor{red}{\textbf{0.363}} & 0.357 & 0.448 & 0.326 & 0.413 & 0.450 & 0.581 & 0.454 & 0.574 & 0.512 & 0.666 & 0.458 & 0.580 \\
 & 720 & 0.320 & 0.402 & \textcolor{blue}{\underline{0.308}} & \textcolor{blue}{\underline{0.395}} & \textcolor{red}{\textbf{0.300}} & \textcolor{red}{\textbf{0.392}} & 0.384 & 0.513 & 0.331 & 0.428 & 0.516 & 0.657 & 0.528 & 0.657 & 0.523 & 0.672 & 0.502 & 0.643 \\  \cmidrule{2-20}
 & Avg & 0.293 & 0.360 & \textcolor{blue}{\underline{0.283}} & \textcolor{blue}{\underline{0.353}} & \textcolor{red}{\textbf{0.276}} & \textcolor{red}{\textbf{0.345}} & 0.337 & 0.432 & 0.310 & 0.398 & 0.457 & 0.586 & 0.479 & 0.603 & 0.502 & 0.651 & 0.441 & 0.559 \\ \midrule
\multirow[c]{5}{*}{\rotatebox{90}{ETTh2}} & 96 & 0.139 & 0.177 & \textcolor{blue}{\underline{0.133}} & \textcolor{blue}{\underline{0.175}} & \textcolor{red}{\textbf{0.122}} & \textcolor{red}{\textbf{0.162}} & 0.175 & 0.220 & 0.182 & 0.206 & 0.336 & 0.421 & 0.164 & 0.214 & 0.358 & 0.448 & 0.432 & 0.548 \\
 & 192 & 0.152 & 0.195 & \textcolor{blue}{\underline{0.148}} & \textcolor{blue}{\underline{0.193}} & \textcolor{red}{\textbf{0.137}} & \textcolor{red}{\textbf{0.188}} & 0.200 & 0.245 & 0.185 & 0.220 & 0.265 & 0.339 & 0.226 & 0.294 & 0.457 & 0.575 & 0.625 & 0.766 \\
 & 336 & 0.170 & 0.216 & \textcolor{blue}{\underline{0.167}} & \textcolor{blue}{\underline{0.209}} & \textcolor{red}{\textbf{0.155}} & \textcolor{red}{\textbf{0.194}} & 0.207 & 0.257 & 0.245 & 0.277 & 0.350 & 0.427 & 0.274 & 0.353 & 0.481 & 0.606 & 0.793 & 0.942 \\
 & 720 & 0.183 & 0.231 & \textcolor{blue}{\underline{0.177}} & \textcolor{blue}{\underline{0.228}} & \textcolor{red}{\textbf{0.171}} & \textcolor{red}{\textbf{0.222}} & 0.209 & 0.260 & 0.237 & 0.272 & 0.406 & 0.482 & 0.302 & 0.382 & 0.445 & 0.550 & 0.539 & 0.688 \\  \cmidrule{2-20}
 & Avg & 0.161 & 0.205 & \textcolor{blue}{\underline{0.156}} & \textcolor{blue}{\underline{0.201}} & \textcolor{red}{\textbf{0.146}} & \textcolor{red}{\textbf{0.192}} & 0.198 & 0.246 & 0.212 & 0.244 & 0.339 & 0.417 & 0.242 & 0.311 & 0.435 & 0.545 & 0.597 & 0.736 \\ \midrule
\multirow[c]{5}{*}{\rotatebox{90}{Weather}} & 96 & 0.071 & 0.088 & 0.069 & \textcolor{blue}{\underline{0.085}} & \textcolor{red}{\textbf{0.065}} & \textcolor{red}{\textbf{0.083}} & 0.123 & 0.129 & 0.083 & 0.087 & 0.104 & 0.113 & \textcolor{blue}{\underline{0.068}} & 0.087 & 0.130 & 0.164 & 0.116 & 0.145 \\
 & 192 & 0.089 & 0.113 & 0.078 & 0.101 & \textcolor{blue}{\underline{0.073}} & \textcolor{blue}{\underline{0.088}} & 0.132 & 0.132 & 0.085 & 0.090 & 0.134 & 0.144 & \textcolor{red}{\textbf{0.068}} & \textcolor{red}{\textbf{0.086}} & 0.127 & 0.158 & 0.122 & 0.147 \\
 & 336 & 0.094 & 0.121 & \textcolor{blue}{\underline{0.082}} & 0.114 & \textcolor{red}{\textbf{0.075}} & \textcolor{red}{\textbf{0.089}} & 0.130 & 0.128 & 0.087 & \textcolor{blue}{\underline{0.091}} & 0.137 & 0.138 & 0.083 & 0.098 & 0.130 & 0.162 & 0.128 & 0.160 \\
 & 720 & 0.103 & 0.132 & 0.098 & 0.119 & 0.091 & \textcolor{blue}{\underline{0.096}} & 0.142 & 0.147 & \textcolor{blue}{\underline{0.090}} & \textcolor{red}{\textbf{0.094}} & 0.152 & 0.141 & \textcolor{red}{\textbf{0.087}} & 0.102 & 0.113 & 0.136 & 0.110 & 0.135 \\  \cmidrule{2-20}
 & Avg & 0.089 & 0.114 & 0.082 & 0.105 & \textcolor{red}{\textbf{0.076}} & \textcolor{red}{\textbf{0.089}} & 0.132 & 0.134 & 0.086 & \textcolor{blue}{\underline{0.091}} & 0.132 & 0.134 & \textcolor{blue}{\underline{0.077}} & 0.093 & 0.125 & 0.155 & 0.119 & 0.147 \\ \midrule
\multirow[c]{5}{*}{\rotatebox{90}{Electricity}} & 96 & 0.071 & 0.090 & \textcolor{blue}{\underline{0.069}} & \textcolor{blue}{\underline{0.088}} & \textcolor{red}{\textbf{0.069}} & \textcolor{red}{\textbf{0.088}} & 0.160 & 0.202 & 0.089 & 0.093 & 0.344 & 0.441 & 0.153 & 0.203 & 0.096 & 0.119 & 0.094 & 0.118 \\
 & 192 & 0.086 & 0.103 & \textcolor{red}{\textbf{0.083}} & \textcolor{blue}{\underline{0.097}} & \textcolor{blue}{\underline{0.084}} & \textcolor{red}{\textbf{0.095}} & 0.248 & 0.358 & 0.097 & 0.102 & 0.345 & 0.441 & 0.200 & 0.264 & 0.100 & 0.124 & 0.097 & 0.121 \\
 & 336 &\textcolor{blue}{\underline{0.096}} & 0.125 & 0.097 & 0.116 & \textcolor{red}{\textbf{0.094}} & \textcolor{blue}{\underline{0.114}} & 0.252 & 0.362 & 0.101 & \textcolor{red}{\textbf{0.105}} & 0.462 & 0.571 & / & / & 0.102 & 0.126 & 0.099 & 0.123 \\
 & 720 & 0.117 & 0.154 & 0.108 & 0.123 & \textcolor{red}{\textbf{0.104}} & \textcolor{blue}{\underline{0.117}} & 0.259 & 0.368 & \textcolor{blue}{\underline{0.105}} & \textcolor{red}{\textbf{0.110}} & 0.478 & 0.622 & / & / & 0.108 & 0.134 & 0.114 & 0.144 \\  \cmidrule{2-20}
 & Avg & 0.093 & 0.118 &\textcolor{blue}{\underline{ 0.089}} & 0.106 & \textcolor{red}{\textbf{0.088}} & \textcolor{blue}{\underline{0.104}} & 0.230 & 0.322 & 0.098 & \textcolor{red}{\textbf{0.103}} & 0.407 & 0.519 & / & / & 0.102 & 0.126 & 0.101 & 0.127 \\ \midrule
\multirow[c]{5}{*}{\rotatebox{90}{Traffic}} & 96 & 0.240 & 0.293 & 0.227 & 0.262 & 0.203 & \textcolor{red}{\textbf{0.217}} & 0.393 & 0.500 & 0.216 & \textcolor{blue}{\underline{0.222}} & 0.294 & 0.342 & / & / & \textcolor{blue}{\underline{0.202}} & 0.234 & \textcolor{red}{\textbf{0.187}} & 0.231 \\
 & 192 & 0.257 & 0.319 & 0.243 & 0.289 & 0.212 & \textcolor{red}{\textbf{0.223}} & 0.448 & 0.585 & 0.220 & \textcolor{blue}{\underline{0.227}} & 0.306 & 0.354 & / & / & \textcolor{blue}{\underline{0.208}} & 0.239 & \textcolor{red}{\textbf{0.192 }}& 0.236 \\
 & 336 & 0.271 & 0.341 & 0.266 & 0.317 & 0.220 & \textcolor{red}{\textbf{0.228}} & 0.455 & 0.592 & 0.228 & \textcolor{blue}{\underline{0.235}} & 0.317 & 0.392 & / & / & \textcolor{blue}{\underline{0.213}} & 0.246 & \textcolor{red}{\textbf{0.201}} & 0.248 \\
 & 720 & 0.304 & 0.395 & 0.288 & 0.322 & 0.231 & \textcolor{red}{\textbf{0.237}} & 0.452 & 0.588 & 0.235 & \textcolor{blue}{\underline{0.242}} & 0.391 & 0.478 & / & / & \textcolor{blue}{\underline{0.220}} & 0.263 & \textcolor{red}{\textbf{0.211}} & 0.264 \\  \cmidrule{2-20}
 & Avg & 0.268 & 0.337 & 0.256 & 0.298 & 0.217 & \textcolor{red}{\textbf{0.226}} & 0.437 & 0.566 & 0.225 & \textcolor{blue}{\underline{0.232}} & 0.327 & 0.392 & / & / & \textcolor{blue}{\underline{0.211}} & 0.246 & \textcolor{red}{\textbf{0.198}} & 0.245 \\ \midrule
\multirow[c]{5}{*}{\rotatebox{90}{Exchange}} & 96 & \textcolor{blue}{\underline{0.023}} & \textcolor{blue}{\underline{0.028}} & 0.024 & 0.031 & \textcolor{red}{\textbf{0.022}} & \textcolor{red}{\textbf{0.027}} & 0.023 & 0.029 & 0.027 & 0.028 & 0.079 & 0.090 & 0.028 & 0.036 & 0.068 & 0.079 & 0.071 & 0.091 \\
 & 192 & 0.032 & 0.037 & \textcolor{blue}{\underline{0.028}} & \textcolor{blue}{\underline{0.036}} & \textcolor{red}{\textbf{0.025}} & \textcolor{red}{\textbf{0.031}} & 0.029 & 0.037 & 0.035 & 0.036 & 0.093 & 0.106 & 0.045 & 0.058 & 0.087 & 0.100 & 0.068 & 0.087 \\
 & 336 & 0.044 & 0.049 & 0.035 & 0.037 & \textcolor{blue}{\underline{0.031}} & \textcolor{blue}{\underline{0.033}} & 0.043 & 0.053 & \textcolor{red}{\textbf{0.027}} & \textcolor{red}{\textbf{0.028}} & 0.081 & 0.106 & 0.060 & 0.076 & 0.074 & 0.086 & 0.072 & 0.091 \\
 & 720 & 0.074 & 0.079 & 0.067 & 0.074 & \textcolor{red}{\textbf{0.053}} & \textcolor{red}{\textbf{0.070}} & \textcolor{blue}{\underline{0.057}} & \textcolor{blue}{\underline{0.071}} & 0.081 & 0.083 & 0.082 & 0.142 & 0.143 & 0.173 & 0.099 & 0.113 & 0.079 & 0.103 \\  \cmidrule{2-20}
 & Avg & 0.043 & 0.048 & 0.039 & 0.045 & \textcolor{red}{\textbf{0.033}} & \textcolor{red}{\textbf{0.040}} & \textcolor{blue}{\underline{0.038}} & 0.047 & 0.043 & \textcolor{blue}{\underline{0.044}} & 0.084 & 0.111 & 0.069 & 0.086 & 0.082 & 0.095 & 0.073 & 0.093 \\ \midrule
\multirow[c]{5}{*}{\rotatebox{90}{ILI}} & 24 & 0.135 & 0.166 & 0.112 & 0.133 & \textcolor{red}{\textbf{0.088}} & \textcolor{red}{\textbf{0.103}} & \textcolor{blue}{\underline{0.092}} & \textcolor{blue}{\underline{0.108}} & 0.110 & 0.113 & 0.228 & 0.242 & 0.250 & 0.263 & 0.275 & 0.296 & 0.257 & 0.283 \\
 & 36 & 0.164 & 0.200 & 0.148 & 0.185 & \textcolor{red}{\textbf{0.137}} & \textcolor{blue}{\underline{0.168}} & 0.151 & 0.178 & \textcolor{blue}{\underline{0.140}} & \textcolor{red}{\textbf{0.144}} & 0.235 & 0.246 & 0.285 & 0.298 & 0.272 & 0.298 & 0.281 & 0.307 \\
 & 48 & 0.176 & 0.216 & 0.169 & 0.205 & \textcolor{red}{\textbf{0.142}} & \textcolor{blue}{\underline{0.171}} & \textcolor{blue}{\underline{0.151}} & 0.178 & 0.153 & \textcolor{red}{\textbf{0.157}} & 0.265 & 0.275 & 0.285 & 0.301 & 0.295 & 0.320 & 0.288 & 0.314 \\
 & 60 & 0.171 & 0.211 & 0.165 & 0.198 & \textcolor{red}{\textbf{0.152}} & \textcolor{blue}{\underline{0.184}} & 0.168 & 0.198 & \textcolor{blue}{\underline{0.164}} & \textcolor{red}{\textbf{0.167}} & 0.263 & 0.272 & 0.283 & 0.299 & 0.295 & 0.325 & 0.307 & 0.333 \\  \cmidrule{2-20}
 & Avg & 0.162 & 0.198 & 0.149 & 0.180 & \textcolor{red}{\textbf{0.130}} & \textcolor{blue}{\underline{0.157}} & \textcolor{blue}{\underline{0.140}} & 0.165 & 0.142 & \textcolor{red}{\textbf{0.145}} & 0.248 & 0.259 & 0.276 & 0.290 & 0.284 & 0.310 & 0.283 & 0.309 \\ \midrule
 \rowc
\multicolumn{2}{c|}{1\textsuperscript{st} Count} & 0 & 0 & 1 & 1 & \textcolor{red}{\textbf{28}} & \textcolor{red}{\textbf{26}} & 0 & 0 & 1 & \textcolor{blue}{\underline{8}} & 0 & 0 & 2 & 1 & 0 & 0 & \textcolor{blue}{\underline{4}} & 0 \\
\bottomrule
\end{tabular}
}
\end{table*}

\renewcommand{\arraystretch}{1.2}
\begin{table*}[t]
\centering
\caption{Deterministic forecasting results. Full-shot, 10\% few-shot and zero-shot results of FLAME Small vs. the most advanced end-to-end supervised models on commonly-used datasets in TSFM-Bench. Lower MSE or MAE values indicate better predictions. \textcolor{red}{\textbf{Red}}: the best, \textcolor{blue}{\underline{Blue}}: the second best.}
\label{tab: Full-shot result 1}
\resizebox{\textwidth}{!}{
\begin{tabular}{cc|cc|cc|cc|cc|cc|cc|cc|cc|cc|cc|cc}
\toprule
\multicolumn{2}{c|}{\multirow{2}{*}{Models}} & \multicolumn{2}{c|}{FLAME Small} & \multicolumn{2}{c|}{FLAME Small} & \multicolumn{2}{c|}{FLAME Small} & \multicolumn{2}{c|}{TimeKAN} & \multicolumn{2}{c|}{AMD} & \multicolumn{2}{c|}{TimePro} & \multicolumn{2}{c|}{TimeXer} & \multicolumn{2}{c|}{Fredformer} & \multicolumn{2}{c|}{iTransformer} & \multicolumn{2}{c|}{PatchTST}  & \multicolumn{2}{c}{TimesNet} \\
~ & ~& \multicolumn{2}{c|}{(Zero-shot)} & \multicolumn{2}{c|}{(10\% few-shot))} & \multicolumn{2}{c|}{(Full-shot)} & \multicolumn{2}{c|}{(2025)} & \multicolumn{2}{c|}{(2025)} & \multicolumn{2}{c|}{(2025)} & \multicolumn{2}{c|}{(2024)} & \multicolumn{2}{c|}{(2024)} & \multicolumn{2}{c|}{(2024)} & \multicolumn{2}{c|}{(2023)}  & \multicolumn{2}{c}{(2023)} \\ \midrule
\multicolumn{2}{c|}{Metric} & \multicolumn{1}{c}{MSE} & \multicolumn{1}{c|}{MAE} & \multicolumn{1}{c}{MSE} & \multicolumn{1}{c|}{MAE} & \multicolumn{1}{c}{MSE} & \multicolumn{1}{c|}{MAE} & \multicolumn{1}{c}{MSE} & \multicolumn{1}{c|}{MAE} & \multicolumn{1}{c}{MSE} & \multicolumn{1}{c|}{MAE} & \multicolumn{1}{c}{MSE} & \multicolumn{1}{c|}{MAE} & \multicolumn{1}{c}{MSE} & \multicolumn{1}{c|}{MAE} & \multicolumn{1}{c}{MSE} & \multicolumn{1}{c|}{MAE} & \multicolumn{1}{c}{MSE} & \multicolumn{1}{c|}{MAE} & \multicolumn{1}{c}{MSE} & \multicolumn{1}{c|}{MAE} & \multicolumn{1}{c}{MSE} & \multicolumn{1}{c}{MAE} \\
\midrule
\multirow[c]{5}{*}{\rotatebox{90}{ETTm1}} & 96 & 0.305 & 0.353 & \textcolor{blue}{\underline{0.278}} & \textcolor{blue}{\underline{0.328}} & \textcolor{red}{\textbf{0.264}} & \textcolor{red}{\textbf{0.318}} & 0.327 & 0.365 & 0.327 & 0.361 & 0.326 & 0.364 & 0.309 & 0.352 & 0.326 & 0.361 & 0.334 & 0.368 & 0.329 & 0.367 & 0.338 & 0.375 \\
 & 192 & 0.328 & 0.370 & \textcolor{blue}{\underline{0.320}} & \textcolor{blue}{\underline{0.359}} & \textcolor{red}{\textbf{0.311}} & \textcolor{red}{\textbf{0.352}} & 0.363 & 0.387 & 0.366 & 0.383 & 0.367 & 0.383 & 0.355 & 0.378 & 0.363 & 0.384 & 0.377 & 0.391 & 0.367 & 0.385 & 0.374 & 0.387 \\
 & 336 & 0.349 & 0.384 & \textcolor{blue}{\underline{0.338}} & \textcolor{blue}{\underline{0.371}} & \textcolor{red}{\textbf{0.332}} & \textcolor{red}{\textbf{0.368}} & 0.389 & 0.407 & 0.398 & 0.404 & 0.402 & 0.409 & 0.387 & 0.399 & 0.395 & 0.406 & 0.426 & 0.420 & 0.399 & 0.410 & 0.410 & 0.411 \\
 & 720 & 0.392 & 0.416 & \textcolor{blue}{\underline{0.383}} & \textcolor{blue}{\underline{0.404}} & \textcolor{red}{\textbf{0.372}} & \textcolor{red}{\textbf{0.399}} & 0.457 & 0.445 & 0.464 & 0.437 & 0.469 & 0.446 & 0.448 & 0.435 & 0.456 & 0.441 & 0.491 & 0.459 & 0.454 & 0.439 & 0.478 & 0.450 \\
\cmidrule(lr){2-24}
 & Avg & 0.344 & 0.381 & \textcolor{blue}{\underline{0.330}} & \textcolor{blue}{\underline{0.366}} & \textcolor{red}{\textbf{0.320}} & \textcolor{red}{\textbf{0.359}} & 0.384 & 0.401 & 0.389 & 0.396 & 0.391 & 0.401 & 0.375 & 0.391 & 0.385 & 0.398 & 0.407 & 0.410 & 0.387 & 0.400 & 0.400 & 0.406 \\
\midrule
\multirow[c]{5}{*}{\rotatebox{90}{ETTm2}} & 96 & 0.165 & 0.255 & \textcolor{blue}{\underline{0.155}} & \textcolor{blue}{\underline{0.250}} & \textcolor{red}{\textbf{0.150}} & \textcolor{red}{\textbf{0.243}} & 0.178 & 0.262 & 0.176 & 0.259 & 0.178 & 0.260 & 0.171 & 0.255 & 0.177 & 0.258 & 0.180 & 0.264 & 0.175 & 0.259 & 0.187 & 0.267 \\
 & 192 & 0.217 & 0.297 & \textcolor{blue}{\underline{0.215}} & \textcolor{blue}{\underline{0.294}} & \textcolor{red}{\textbf{0.214}} & \textcolor{red}{\textbf{0.292}} & 0.244 & 0.308 & 0.242 & 0.302 & 0.242 & 0.303 & 0.238 & 0.300 & 0.243 & 0.301 & 0.250 & 0.309 & 0.241 & 0.302 & 0.249 & 0.309 \\
 & 336 & 0.262 & 0.330 & \textcolor{blue}{\underline{0.258}} & \textcolor{blue}{\underline{0.318}} & \textcolor{red}{\textbf{0.253}} & \textcolor{red}{\textbf{0.312}} & 0.305 & 0.346 & 0.298 & 0.337 & 0.303 & 0.342 & 0.301 & 0.340 & 0.302 & 0.340 & 0.311 & 0.348 & 0.305 & 0.343 & 0.321 & 0.351 \\
 & 720 & 0.333 & 0.379 & \textcolor{blue}{\underline{0.329}} & \textcolor{blue}{\underline{0.368}} & \textcolor{red}{\textbf{0.328}} & \textcolor{red}{\textbf{0.364}} & 0.402 & 0.404 & 0.396 & 0.394 & 0.400 & 0.399 & 0.401 & 0.397 & 0.404 & 0.398 & 0.412 & 0.407 & 0.402 & 0.400 & 0.408 & 0.403 \\
\cmidrule(lr){2-24}
 & Avg & 0.244 & 0.315 & \textcolor{blue}{\underline{0.239}} & \textcolor{blue}{\underline{0.308}} & \textcolor{red}{\textbf{0.236}} & \textcolor{red}{\textbf{0.303}} & 0.282 & 0.330 & 0.278 & 0.323 & 0.281 & 0.326 & 0.278 & 0.323 & 0.281 & 0.324 & 0.288 & 0.332 & 0.281 & 0.326 & 0.291 & 0.333 \\
\midrule
\multirow[c]{5}{*}{\rotatebox{90}{ETTh1}} & 96 & 0.362 & 0.395 & \textcolor{blue}{\underline{0.357}} & \textcolor{blue}{\underline{0.386}} & \textcolor{red}{\textbf{0.342}} & \textcolor{red}{\textbf{0.375}} & 0.374 & 0.391 & 0.375 & 0.392 & 0.375 & 0.398 & 0.377 & 0.397 & 0.378 & 0.395 & 0.386 & 0.405 & 0.414 & 0.419 & 0.384 & 0.402 \\
 & 192 & 0.406 & 0.427 & \textcolor{blue}{\underline{0.392}} & \textcolor{blue}{\underline{0.413}} & \textcolor{red}{\textbf{0.384}} & \textcolor{red}{\textbf{0.404}} & 0.421 & 0.421 & 0.430 & 0.422 & 0.427 & 0.429 & 0.425 & 0.425 & 0.435 & 0.424 & 0.441 & 0.436 & 0.460 & 0.445 & 0.436 & 0.429 \\
 & 336 & 0.436 & 0.449 & \textcolor{blue}{\underline{0.420}} & \textcolor{blue}{\underline{0.435}} & \textcolor{red}{\textbf{0.407}} & \textcolor{red}{\textbf{0.428}} & 0.464 & 0.440 & 0.471 & 0.443 & 0.472 & 0.450 & 0.457 & 0.441 & 0.485 & 0.447 & 0.487 & 0.458 & 0.501 & 0.466 & 0.491 & 0.469 \\
 & 720 & 0.471 & 0.482 & \textcolor{blue}{\underline{0.436}} & \textcolor{blue}{\underline{0.450}} & \textcolor{red}{\textbf{0.418}} & \textcolor{red}{\textbf{0.436}} & 0.466 & 0.462 & 0.478 & 0.464 & 0.476 & 0.474 & 0.464 & 0.463 & 0.496 & 0.472 & 0.503 & 0.491 & 0.500 & 0.488 & 0.521 & 0.500 \\
\cmidrule(lr){2-24}
 & Avg & 0.419 & 0.438 & \textcolor{blue}{\underline{0.401}} & \textcolor{blue}{\underline{0.421}} & \textcolor{red}{\textbf{0.388}} & \textcolor{red}{\textbf{0.411}} & 0.431 & 0.429 & 0.438 & 0.430 & 0.438 & 0.438 & 0.431 & 0.432 & 0.448 & 0.435 & 0.454 & 0.448 & 0.469 & 0.455 & 0.458 & 0.450 \\
\midrule
\multirow[c]{5}{*}{\rotatebox{90}{ETTh2}} & 96 & 0.278 & 0.339 & \textcolor{blue}{\underline{0.271}} & \textcolor{blue}{\underline{0.327}} & \textcolor{red}{\textbf{0.264}} & \textcolor{red}{\textbf{0.320}} & 0.293 & 0.343 & 0.287 & 0.338 & 0.293 & 0.345 & 0.289 & 0.340 & 0.291 & 0.342 & 0.297 & 0.349 & 0.302 & 0.348 & 0.340 & 0.374 \\
 & 192 & 0.336 & 0.381 & \textcolor{blue}{\underline{0.335}} & \textcolor{blue}{\underline{0.376}} & \textcolor{red}{\textbf{0.332}} & \textcolor{red}{\textbf{0.364}} & 0.375 & 0.396 & 0.367 & 0.388 & 0.367 & 0.394 & 0.370 & 0.391 & 0.372 & 0.390 & 0.380 & 0.400 & 0.388 & 0.400 & 0.402 & 0.414 \\
 & 336 & 0.373 & 0.410 & \textcolor{blue}{\underline{0.364}} & \textcolor{blue}{\underline{0.402}} & \textcolor{red}{\textbf{0.360}} & \textcolor{red}{\textbf{0.399}} & 0.429 & 0.441 & 0.410 & 0.424 & 0.419 & 0.431 & 0.422 & 0.434 & 0.419 & 0.431 & 0.428 & 0.432 & 0.426 & 0.433 & 0.452 & 0.452 \\
 & 720 & 0.411 & 0.441 & \textcolor{blue}{\underline{0.391}} & \textcolor{blue}{\underline{0.428}} & \textcolor{red}{\textbf{0.382}} & \textcolor{red}{\textbf{0.423}} & 0.466 & 0.468 & 0.421 & 0.440 & 0.427 & 0.445 & 0.429 & 0.445 & 0.431 & 0.450 & 0.427 & 0.445 & 0.431 & 0.446 & 0.462 & 0.468 \\
\cmidrule(lr){2-24}
 & Avg & 0.350 & 0.393 & \textcolor{blue}{\underline{0.340}} & \textcolor{blue}{\underline{0.383}} & \textcolor{red}{\textbf{0.335}} & \textcolor{red}{\textbf{0.377}} & 0.391 & 0.412 & 0.371 & 0.397 & 0.377 & 0.404 & 0.378 & 0.403 & 0.378 & 0.403 & 0.383 & 0.407 & 0.387 & 0.407 & 0.414 & 0.427 \\
\midrule
\multirow[c]{5}{*}{\rotatebox{90}{Weather}} & 96 & 0.148 & 0.204 & \textcolor{blue}{\underline{0.144}} & \textcolor{blue}{\underline{0.197}} & \textcolor{red}{\textbf{0.141}} & \textcolor{red}{\textbf{0.184}} & 0.164 & 0.210 & 0.174 & 0.221 & 0.166 & 0.207 & 0.168 & 0.209 & 0.163 & 0.207 & 0.174 & 0.214 & 0.177 & 0.218 & 0.172 & 0.220 \\
 & 192 & 0.189 & 0.243 & \textcolor{blue}{\underline{0.184}} & \textcolor{blue}{\underline{0.241}} & \textcolor{red}{\textbf{0.182}} & \textcolor{red}{\textbf{0.240}} & 0.209 & 0.250 & 0.219 & 0.259 & 0.216 & 0.254 & 0.220 & 0.254 & 0.224 & 0.258 & 0.221 & 0.254 & 0.255 & 0.259 & 0.219 & 0.261 \\
 & 336 & 0.233 & 0.278 & \textcolor{blue}{\underline{0.230}} & \textcolor{blue}{\underline{0.276}} & \textcolor{red}{\textbf{0.224}} & \textcolor{red}{\textbf{0.272}} & 0.264 & 0.290 & 0.273 & 0.296 & 0.273 & 0.296 & 0.276 & 0.296 & 0.278 & 0.298 & 0.278 & 0.296 & 0.278 & 0.297 & 0.280 & 0.306 \\
 & 720 & 0.289 & 0.320 & \textcolor{blue}{\underline{0.287}} & \textcolor{blue}{\underline{0.318}} & \textcolor{red}{\textbf{0.280}} & \textcolor{red}{\textbf{0.306}} & 0.343 & 0.342 & 0.349 & 0.345 & 0.351 & 0.346 & 0.353 & 0.347 & 0.357 & 0.350 & 0.358 & 0.349 & 0.354 & 0.348 & 0.365 & 0.359 \\
\cmidrule(lr){2-24}
 & Avg & 0.215 & 0.261 & \textcolor{blue}{\underline{0.211}} & \textcolor{blue}{\underline{0.258}} & \textcolor{red}{\textbf{0.207}} & \textcolor{red}{\textbf{0.251}} & 0.245 & 0.273 & 0.254 & 0.280 & 0.252 & 0.276 & 0.254 & 0.277 & 0.256 & 0.278 & 0.258 & 0.278 & 0.266 & 0.281 & 0.259 & 0.287 \\
\midrule
\multirow[c]{5}{*}{\rotatebox{90}{Electricity}} & 96 & 0.206 & 0.294 & 0.142 & 0.235 & \textcolor{red}{\textbf{0.124}} & \textcolor{red}{\textbf{0.218}} & 0.174 & 0.266 & 0.147 & 0.251 & \textcolor{blue}{\underline{0.139}} & \textcolor{blue}{\underline{0.234}} & 0.151 & 0.247 & 0.148 & 0.242 & 0.148 & 0.240 & 0.195 & 0.285 & 0.168 & 0.272 \\
 & 192 & 0.247 & 0.338 & 0.168 & 0.255 & \textcolor{red}{\textbf{0.149}} & \textcolor{red}{\textbf{0.236}} & 0.182 & 0.273 & 0.176 & 0.262 & \textcolor{blue}{\underline{0.156}} & \textcolor{blue}{\underline{0.249}} & 0.165 & 0.261 & 0.165 & 0.257 & 0.162 & 0.253 & 0.199 & 0.289 & 0.184 & 0.289 \\
 & 336 & 0.310 & 0.398 & 0.188 & 0.271 & \textcolor{red}{\textbf{0.164}} & \textcolor{red}{\textbf{0.252}} & 0.197 & 0.286 & 0.193 & 0.281 & \textcolor{blue}{\underline{0.172}} & \textcolor{blue}{\underline{0.267}} & 0.183 & 0.280 & 0.180 & 0.274 & 0.178 & 0.269 & 0.215 & 0.305 & 0.198 & 0.300 \\
 & 720 & 0.383 & 0.455 & \textcolor{red}{\textbf{0.192}} & \textcolor{red}{\textbf{0.277}} & \textcolor{blue}{\underline{0.205}} & \textcolor{blue}{\underline{0.288}} & 0.236 & 0.320 & 0.232 & 0.329 & 0.209 & 0.299 & 0.220 & 0.309 & 0.218 & 0.305 & 0.225 & 0.317 & 0.256 & 0.337 & 0.220 & 0.320 \\
\cmidrule(lr){2-24}
 & Avg & 0.287 & 0.371 & 0.173 & \textcolor{blue}{\underline{0.260}} & \textcolor{red}{\textbf{0.161}} & \textcolor{red}{\textbf{0.249}} & 0.197 & 0.286 & 0.187 & 0.281 & \textcolor{blue}{\underline{0.169}} & 0.262 & 0.180 & 0.274 & 0.178 & 0.270 & 0.178 & 0.270 & 0.216 & 0.304 & 0.193 & 0.295 \\
\midrule
\multirow[c]{5}{*}{\rotatebox{90}{Traffic}} & 96 & 0.591 & 0.401 & \textcolor{blue}{\underline{0.375}} & \textcolor{blue}{\underline{0.262}} & \textcolor{red}{\textbf{0.358}} & \textcolor{red}{\textbf{0.246}} & 0.423 & 0.286 & 0.443 & 0.298 & 0.426 & 0.292 & 0.416 & 0.280 & 0.403 & 0.274 & 0.395 & 0.268 & 0.544 & 0.359 & 0.593 & 0.321 \\
 & 192 & 0.623 & 0.431 & \textcolor{blue}{\underline{0.388}} & \textcolor{blue}{\underline{0.265}} & \textcolor{red}{\textbf{0.379}} & \textcolor{red}{\textbf{0.248}} & 0.442 & 0.295 & 0.496 & 0.323 & 0.439 & 0.298 & 0.435 & 0.288 & 0.429 & 0.289 & 0.417 & 0.276 & 0.540 & 0.354 & 0.617 & 0.336 \\
 & 336 & 0.697 & 0.485 & \textcolor{blue}{\underline{0.397}} & 0.285 & \textcolor{red}{\textbf{0.386}} & \textcolor{red}{\textbf{0.262}} & 0.473 & 0.335 & 0.520 & 0.330 & 0.449 & 0.307 & 0.451 & 0.296 & 0.441 & 0.295 & 0.433 & \textcolor{blue}{\underline{0.283}} & 0.551 & 0.358 & 0.629 & 0.336 \\
 & 720 & 0.782 & 0.528 & \textcolor{blue}{\underline{0.451}} & 0.332 & \textcolor{red}{\textbf{0.427}} & \textcolor{red}{\textbf{0.282}} & 0.481 & 0.357 & 0.540 & 0.344 & 0.475 & 0.309 & 0.484 & 0.314 & 0.463 & \textcolor{blue}{\underline{0.300}} & 0.467 & 0.302 & 0.586 & 0.375 & 0.640 & 0.350 \\
\cmidrule(lr){2-24}
 & Avg & 0.673 & 0.461 & \textcolor{blue}{\underline{0.403}} & 0.286 & \textcolor{red}{\textbf{0.388}} & \textcolor{red}{\textbf{0.260}} & 0.455 & 0.318 & 0.500 & 0.324 & 0.447 & 0.302 & 0.447 & 0.295 & 0.434 & 0.289 & 0.428 & \textcolor{blue}{\underline{0.282}} & 0.555 & 0.362 & 0.620 & 0.336 \\
\midrule
\rowc
\multicolumn{2}{c|}{1\textsuperscript{st} Count} & 0 & 0 & \textcolor{blue}{\underline{1}} & \textcolor{blue}{\underline{1}} & \textcolor{red}{\textbf{34}} & \textcolor{red}{\textbf{34}} & 0 & 0 & 0 & 0 & 0 & 0 & 0 & 0 & 0 & 0 & 0 & 0 & 0 & 0 & 0 & 0 \\
\bottomrule
\end{tabular}
}
\end{table*}

\clearpage
\section{Showcases}
\begin{figure*}[!htbp]
    \centering
    \begin{subfigure}[b]{0.49\textwidth}
        \centering
        \includegraphics[width=\textwidth]{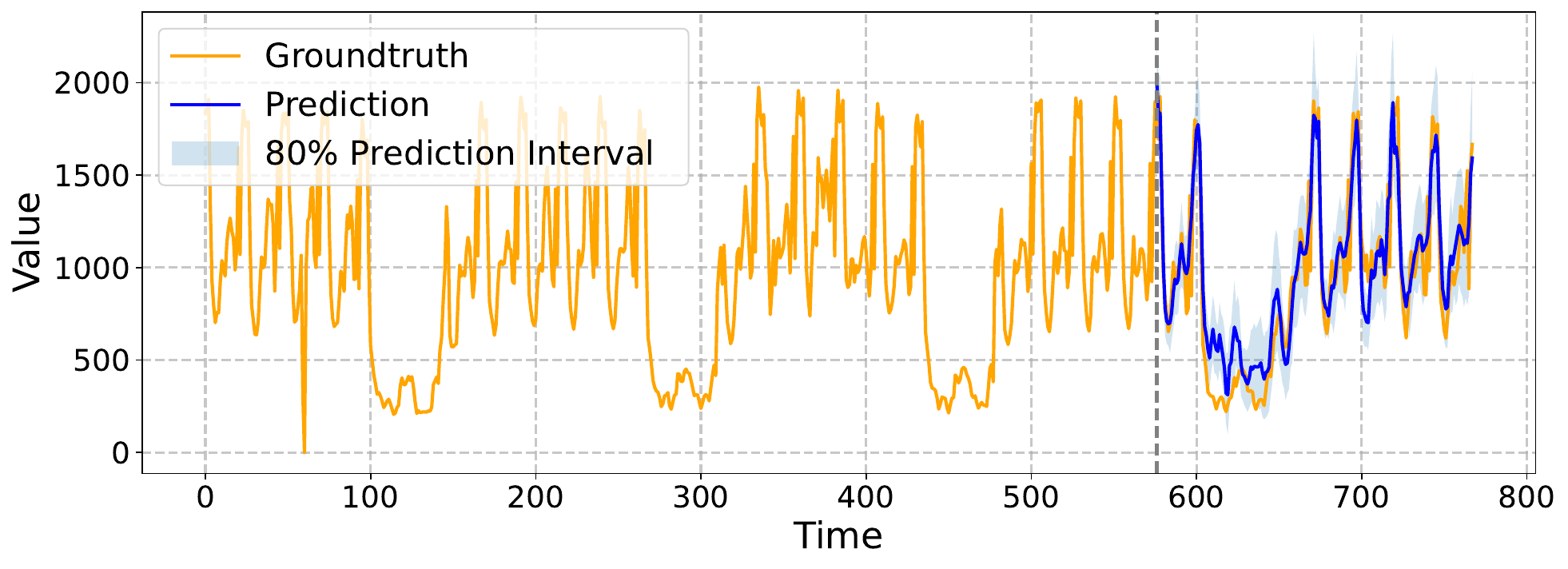} 
        \caption{Electricity}
    \end{subfigure}
    \hfill 
    \begin{subfigure}[b]{0.49\textwidth}
        \centering
        \includegraphics[width=\textwidth]{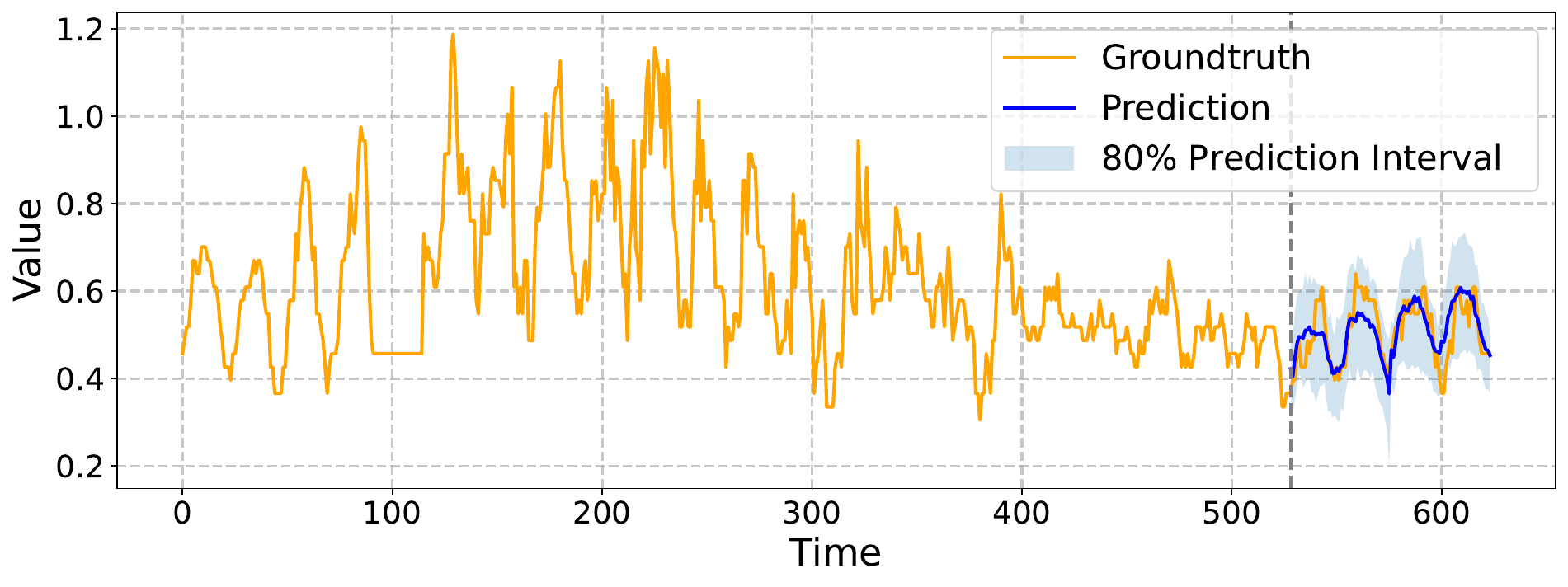} 
        \caption{ETTh1}
    \end{subfigure}
    
    \begin{subfigure}[b]{0.49\textwidth}
        \centering
        \includegraphics[width=\textwidth]{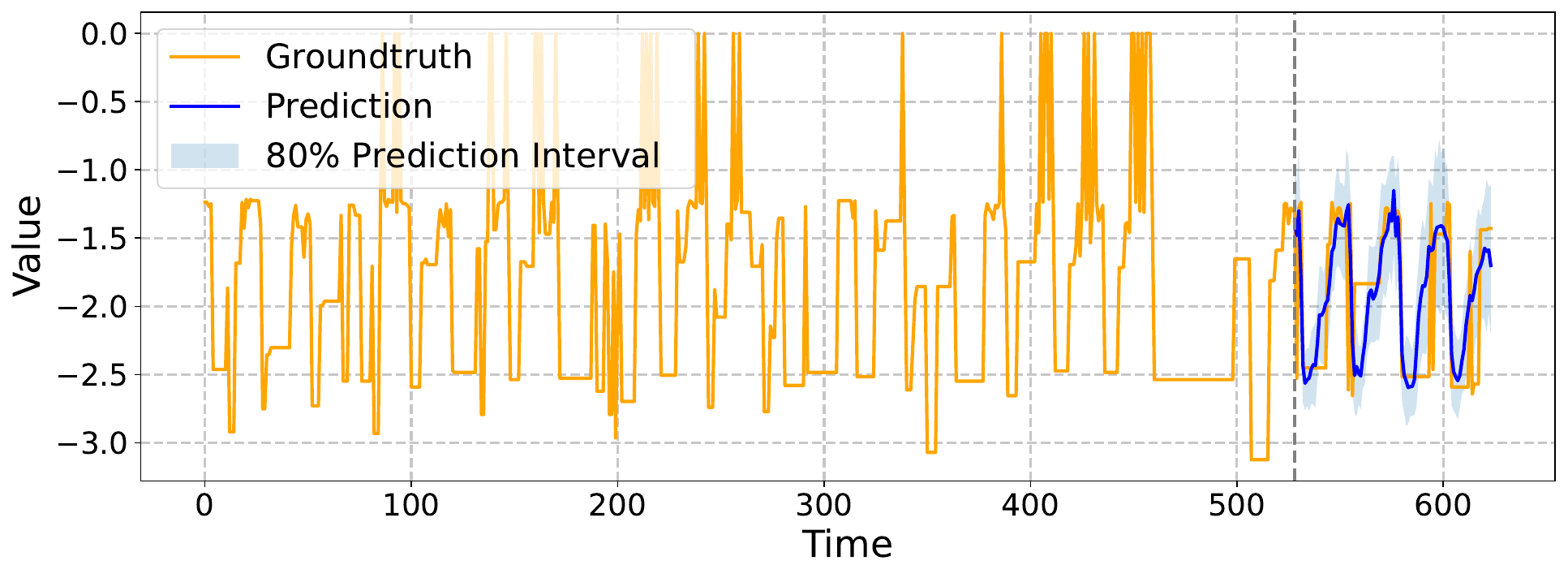}
        \caption{ETTh2}
    \end{subfigure}
    \hfill
    \begin{subfigure}[b]{0.49\textwidth}
        \centering
        \includegraphics[width=\textwidth]{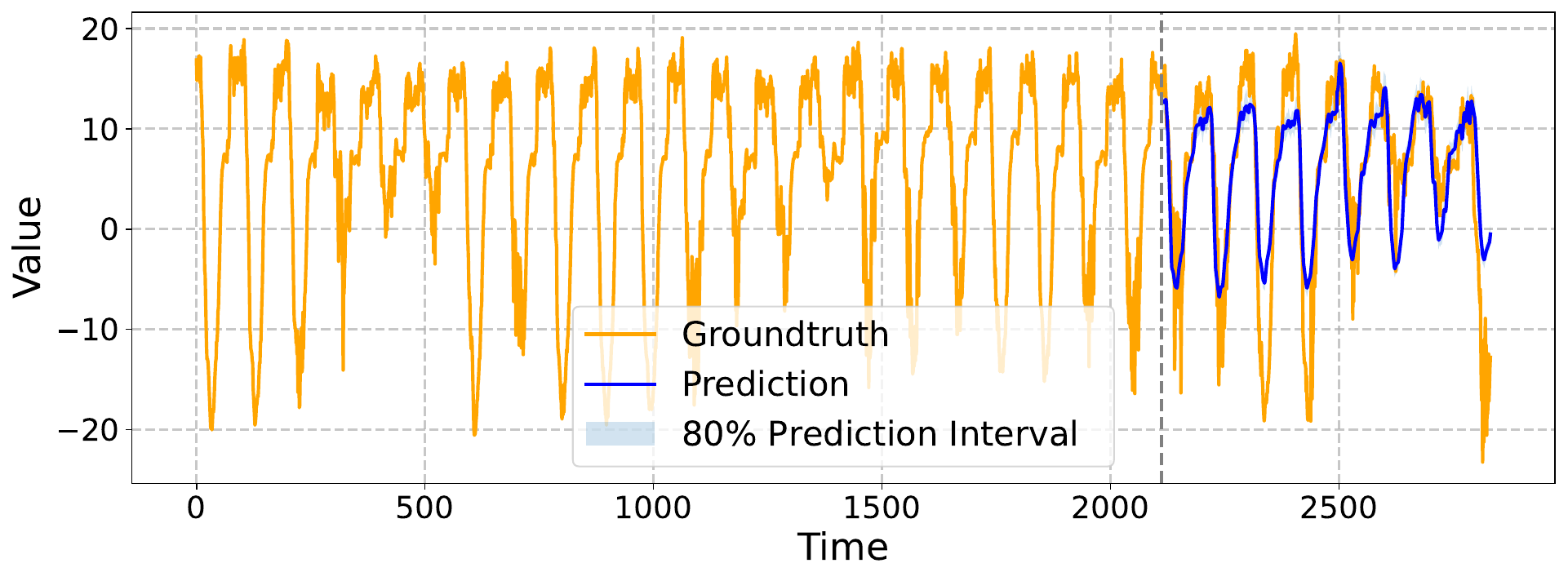}
        \caption{ETTm1}
    \end{subfigure}
    
    \begin{subfigure}[b]{0.49\textwidth}
        \centering
        \includegraphics[width=\textwidth]{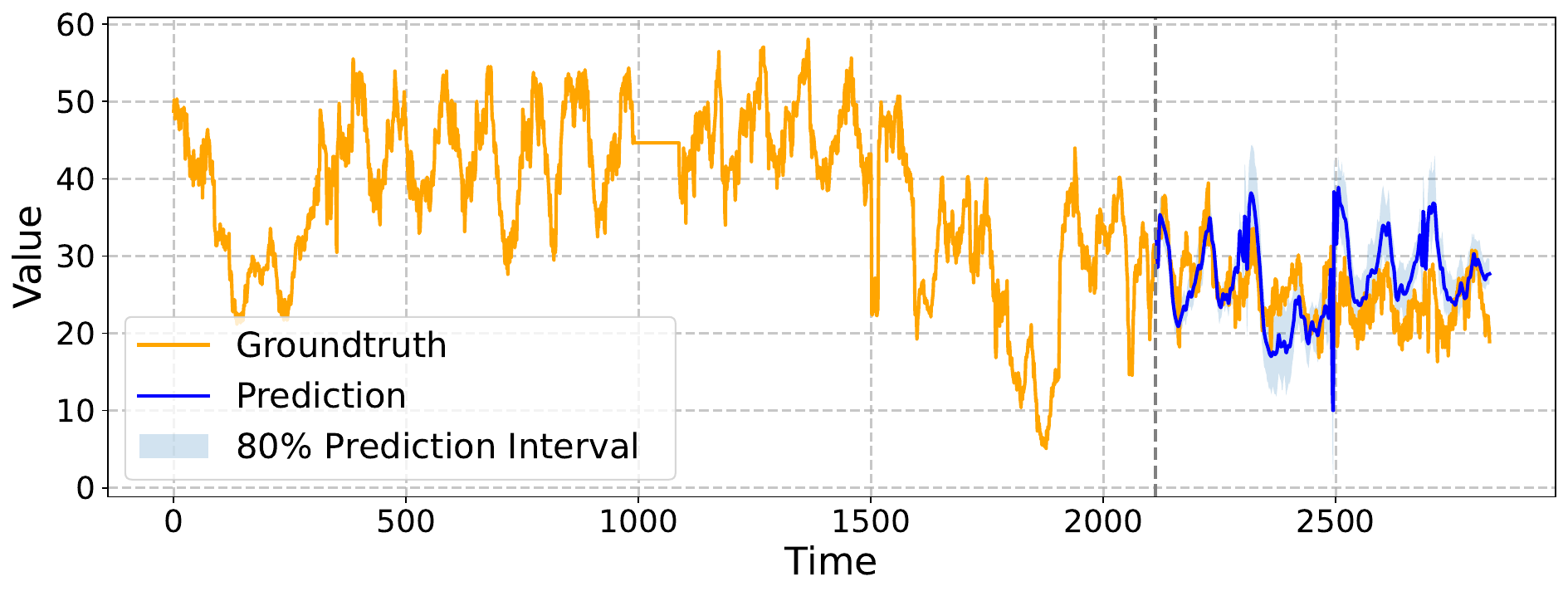}
        \caption{ETTm2}
    \end{subfigure}
    \hfill
    \begin{subfigure}[b]{0.49\textwidth}
        \centering
        \includegraphics[width=\textwidth]{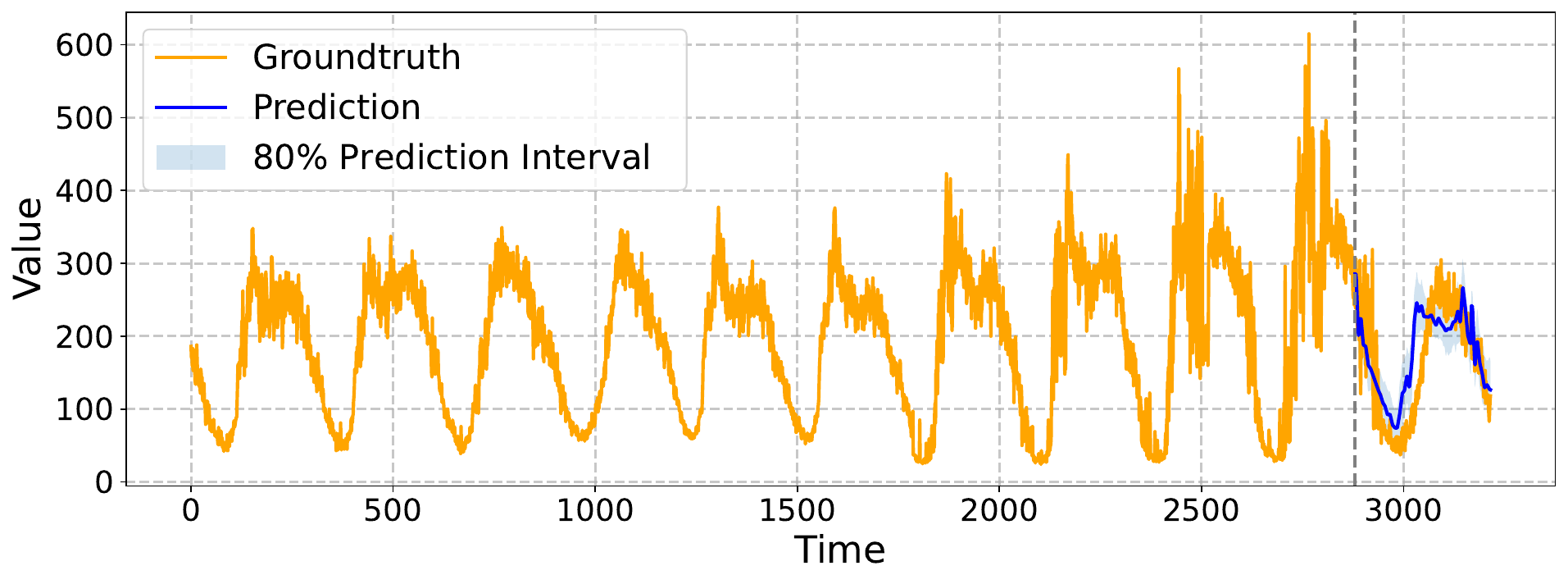}
        \caption{PEMS08}
    \end{subfigure}
    
    \begin{subfigure}[b]{0.49\textwidth}
        \centering
        \includegraphics[width=\textwidth]{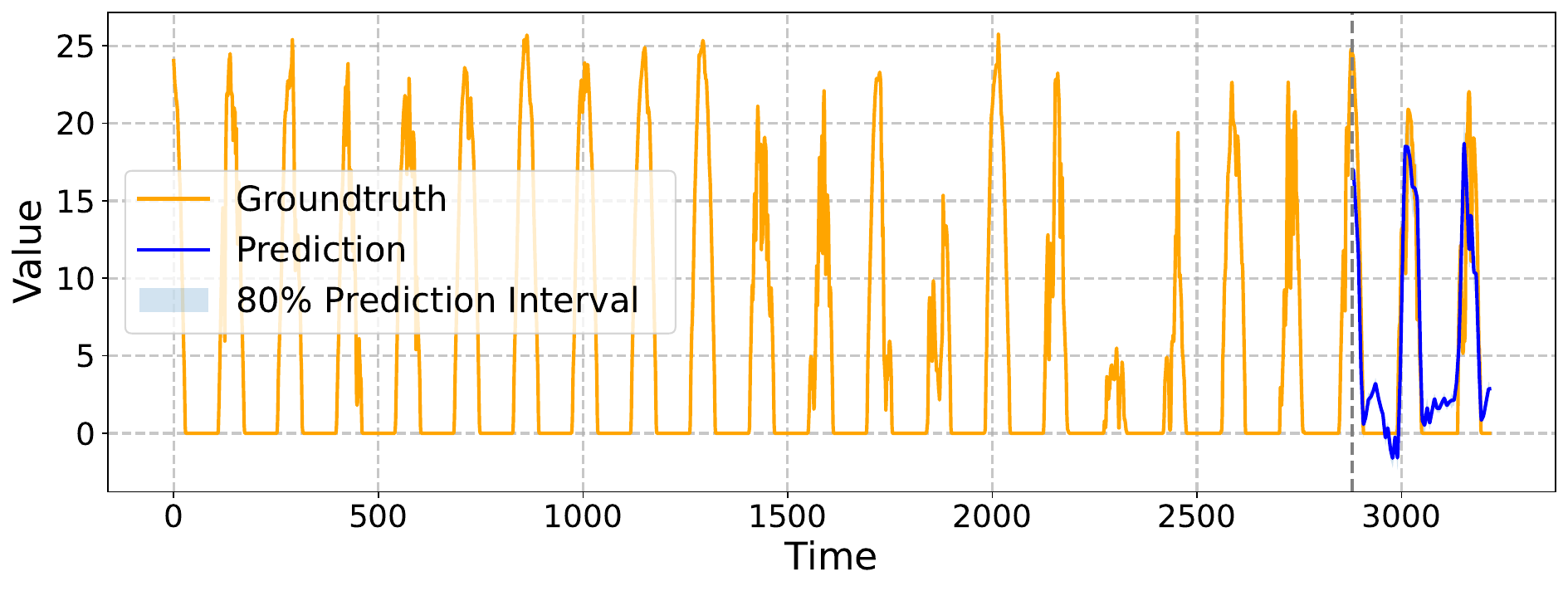}
        \caption{Solar}
    \end{subfigure}
    \hfill
    \begin{subfigure}[b]{0.49\textwidth}
        \centering
        \includegraphics[width=\textwidth]{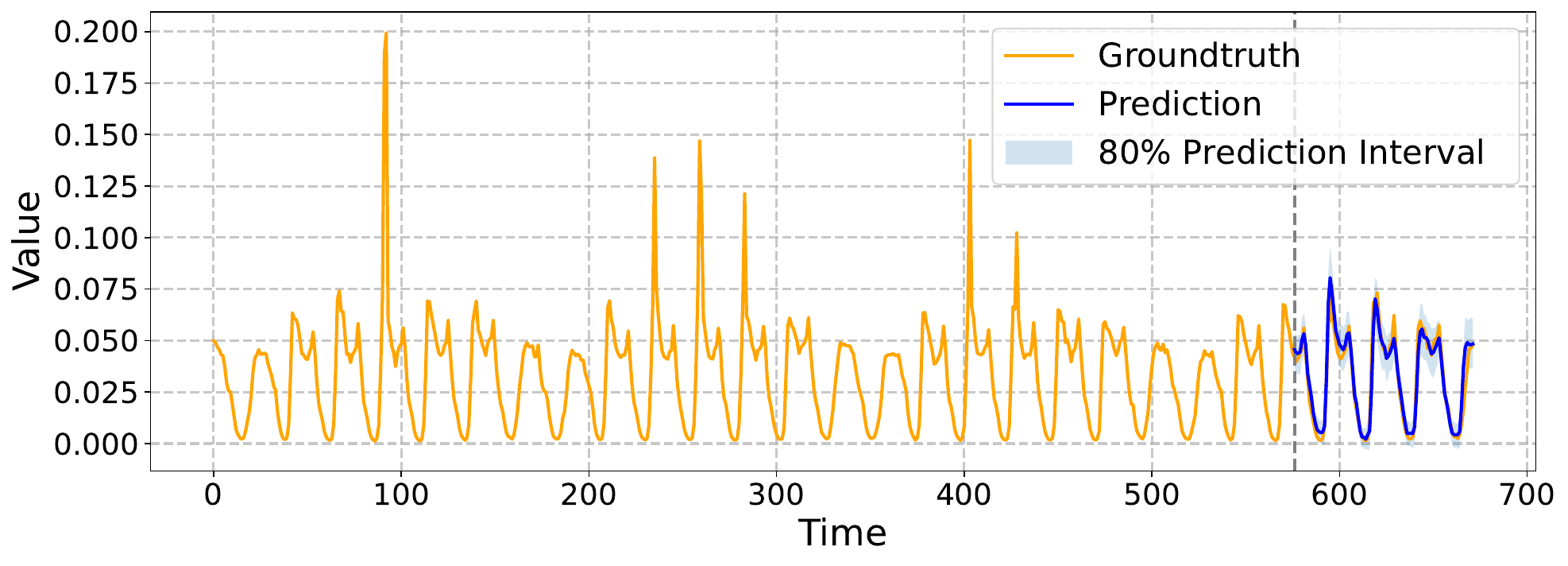}
        \caption{Traffic}
    \end{subfigure}

    \begin{subfigure}[b]{0.49\textwidth}
        \centering
        \includegraphics[width=\textwidth]{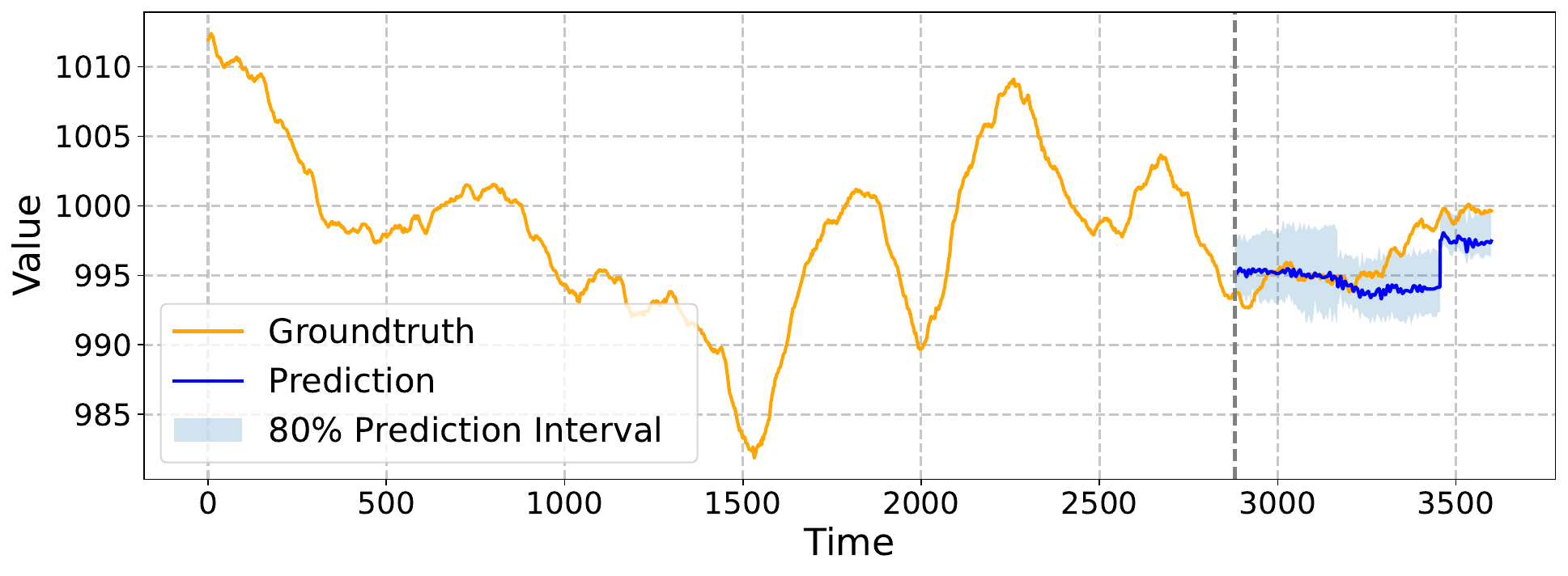}
        \caption{Weather}
    \end{subfigure}
    \hfill
    \begin{subfigure}[b]{0.49\textwidth}
        \centering
        \includegraphics[width=\textwidth]{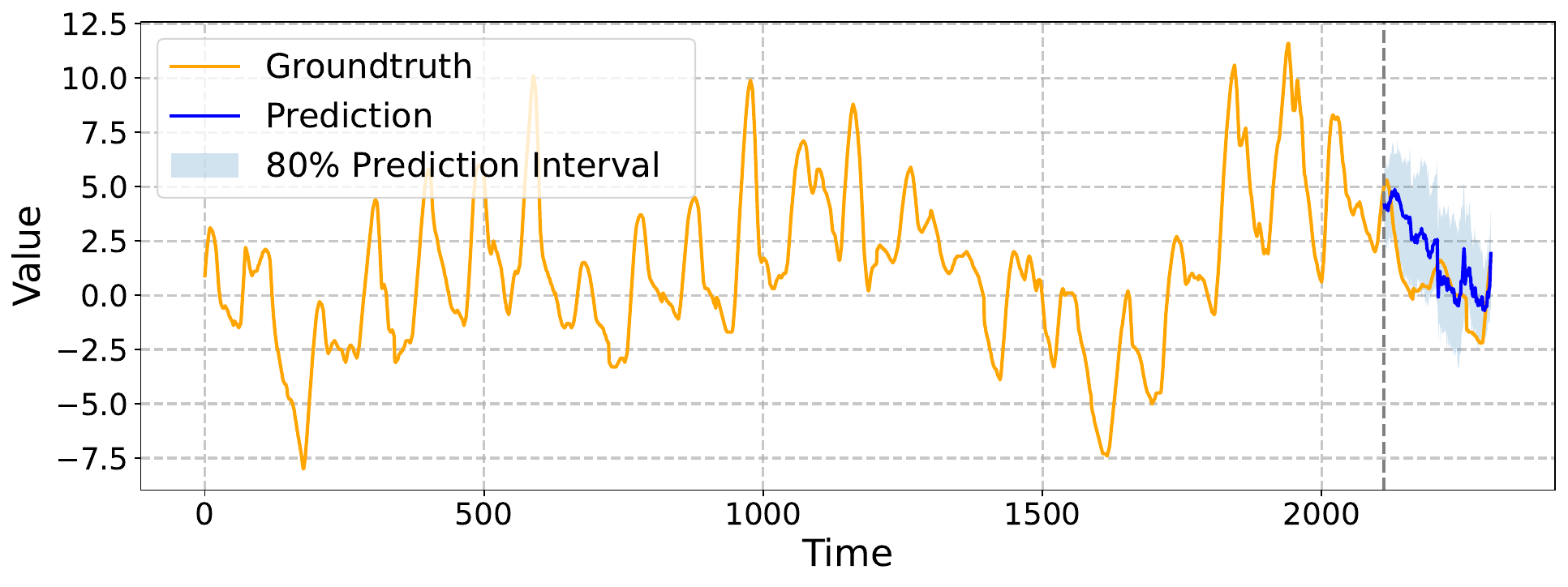}
        \caption{Wind}
    \end{subfigure}
    
    \caption{Visualization of TSFM-Bench} 
    \label{fig: visualization of TSFM-Bench} 
\end{figure*}

\begin{figure*}[!htbp]
    \centering
    \begin{subfigure}[b]{0.49\textwidth}
        \centering
        \includegraphics[width=\textwidth]{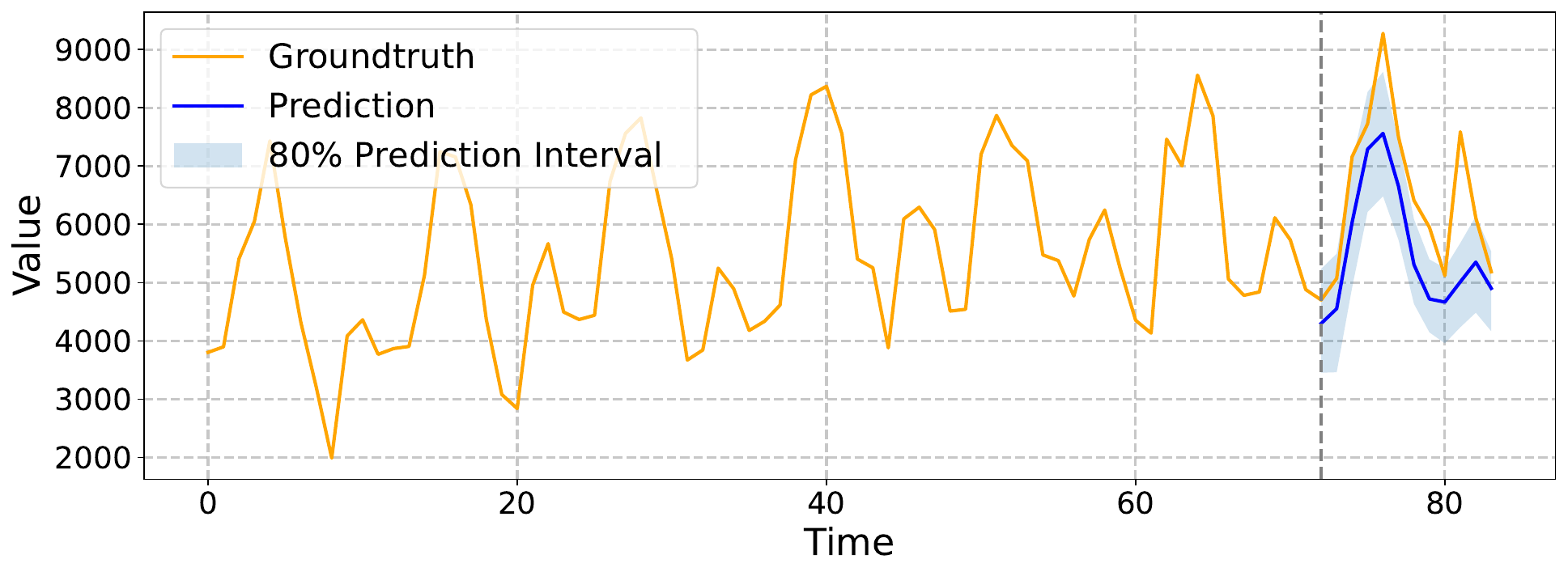} 
        \caption{economics\_36}
    \end{subfigure}
    \hfill 
    \begin{subfigure}[b]{0.49\textwidth}
        \centering
        \includegraphics[width=\textwidth]{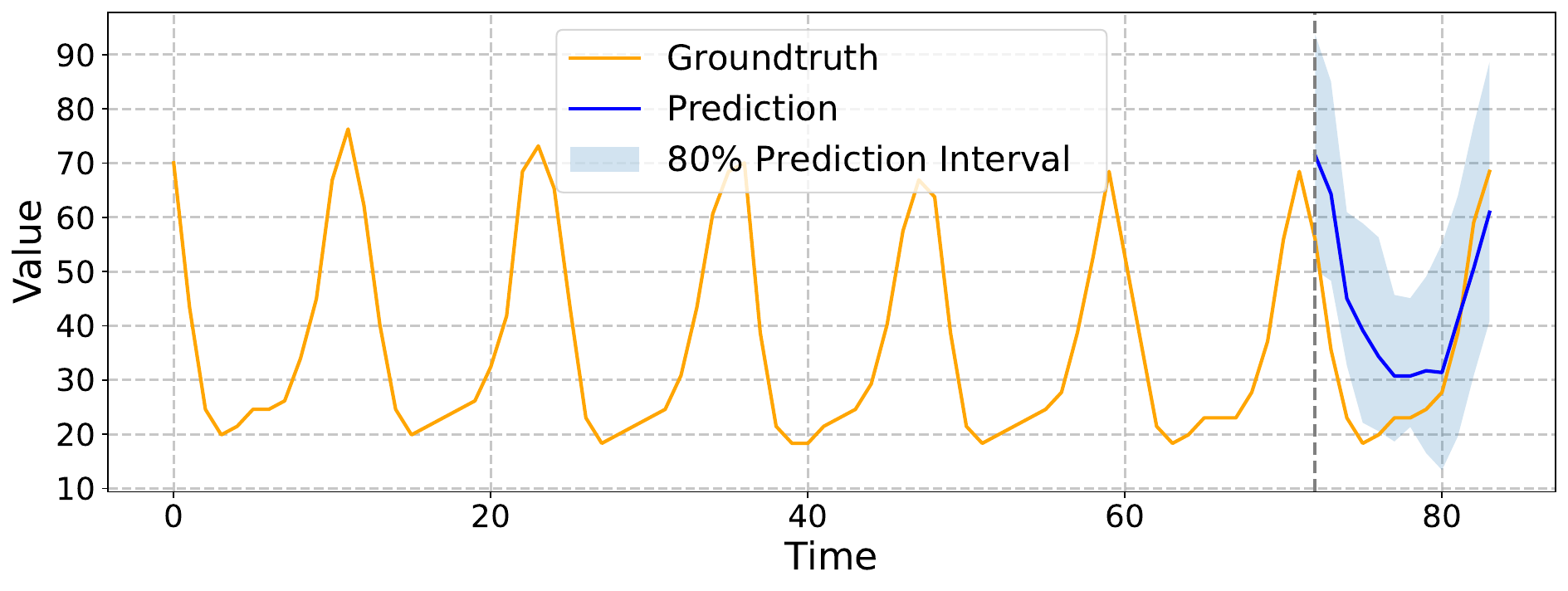} 
        \caption{human\_58}
    \end{subfigure}
    
    \begin{subfigure}[b]{0.49\textwidth}
        \centering
        \includegraphics[width=\textwidth]{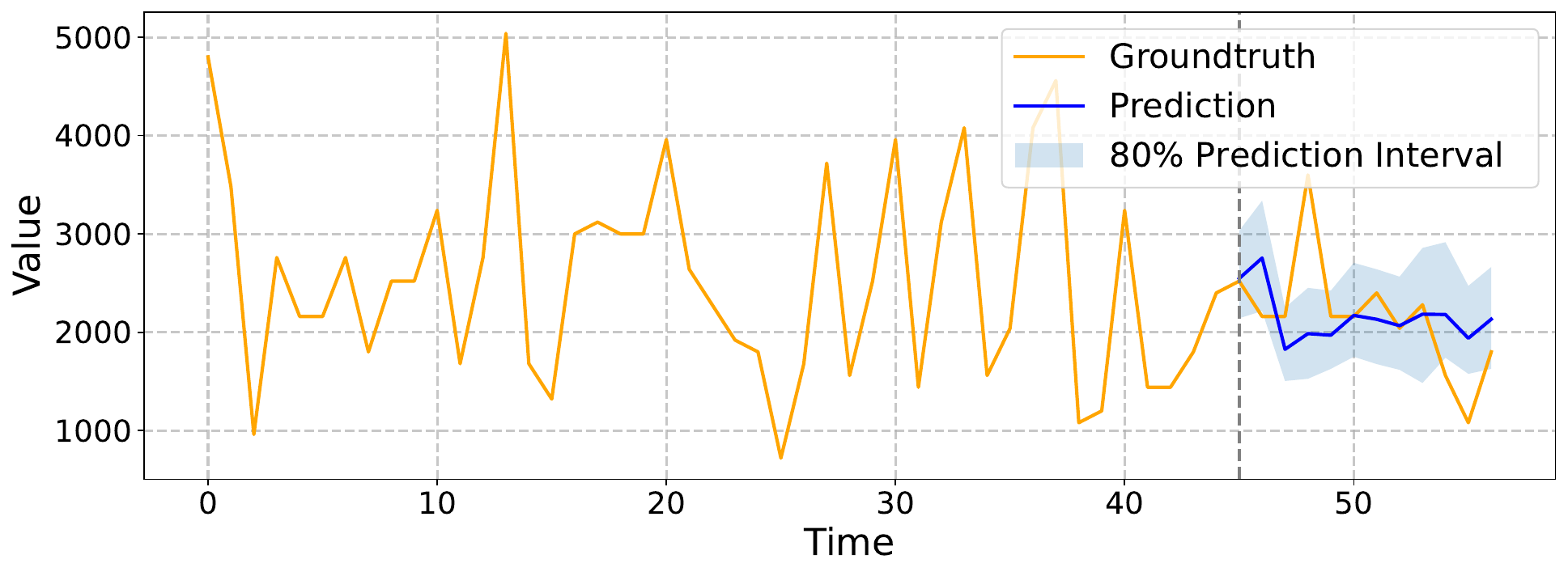}
        \caption{m3\_monthly\_dataset\_223}
    \end{subfigure}
    \hfill
    \begin{subfigure}[b]{0.49\textwidth}
        \centering
        \includegraphics[width=\textwidth]{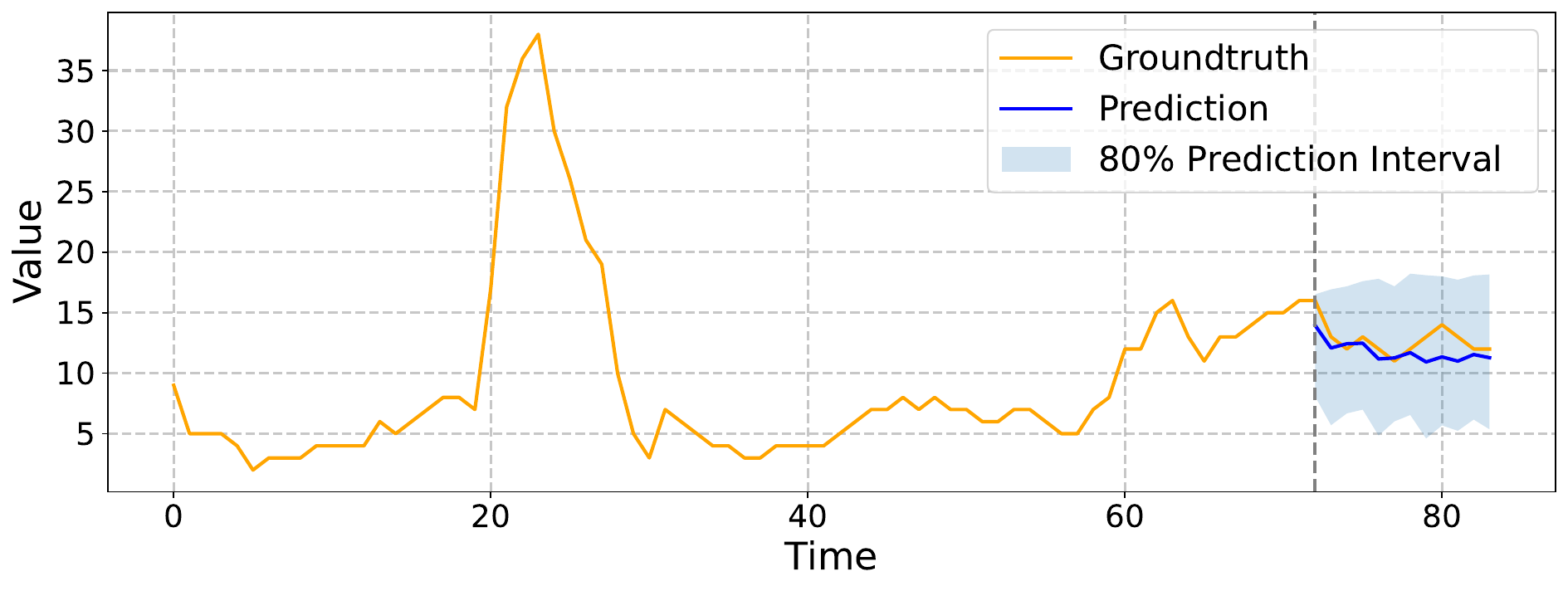}
        \caption{KDD Cup 2018, series 60}
    \end{subfigure}
    
    \begin{subfigure}[b]{0.49\textwidth}
        \centering
        \includegraphics[width=\textwidth]{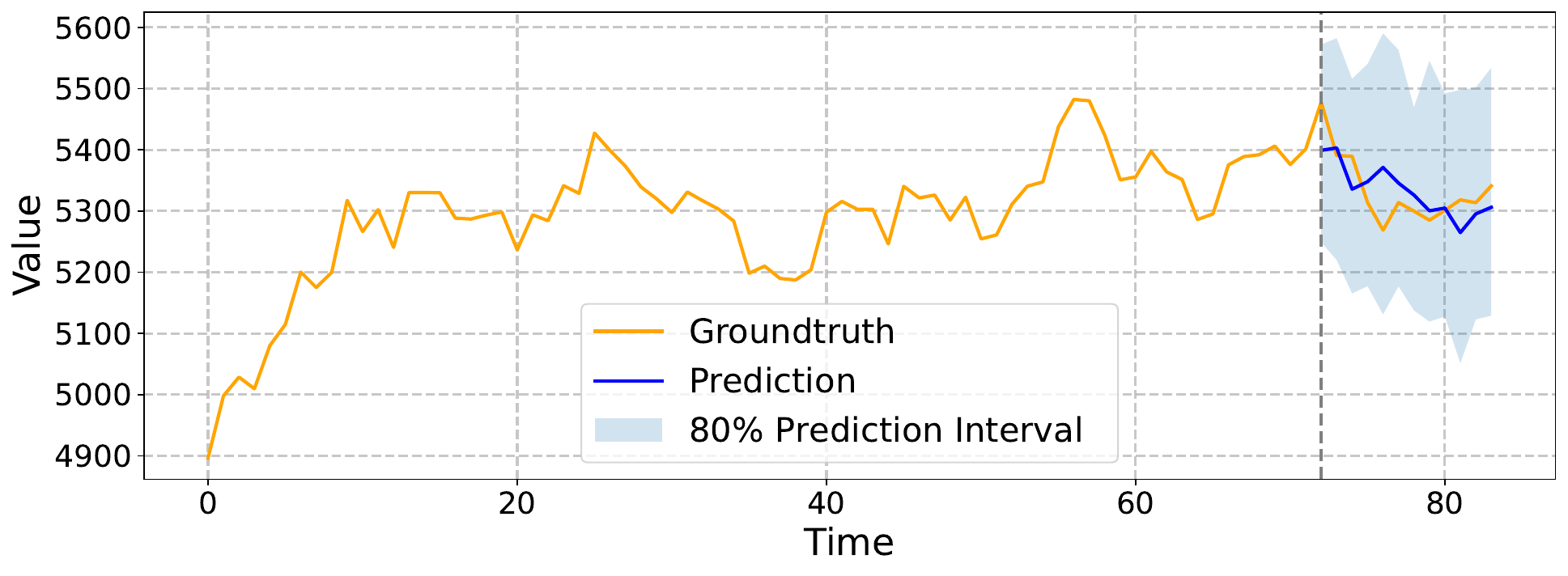}
        \caption{m3\_other\_dataset\_7}
    \end{subfigure}
    \hfill
    \begin{subfigure}[b]{0.49\textwidth}
        \centering
        \includegraphics[width=\textwidth]{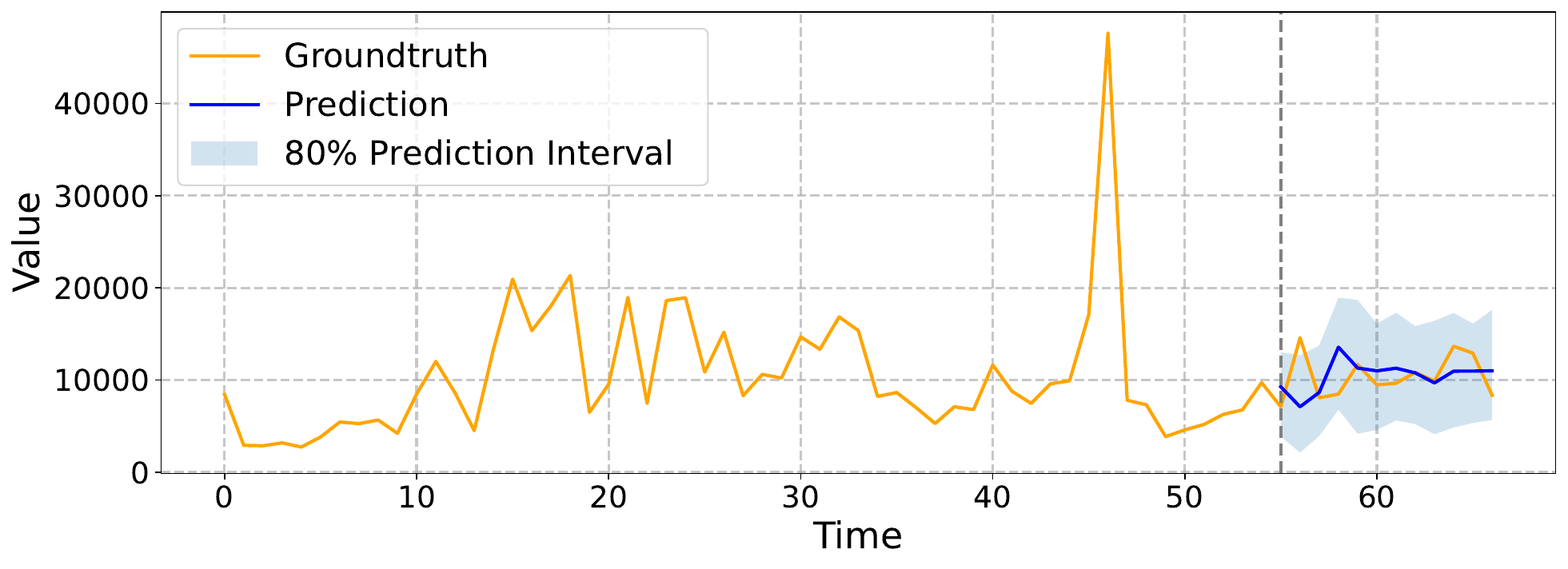}
        \caption{m4\_quarterly\_dataset\_18009}
    \end{subfigure}
    
    \begin{subfigure}[b]{0.49\textwidth}
        \centering
        \includegraphics[width=\textwidth]{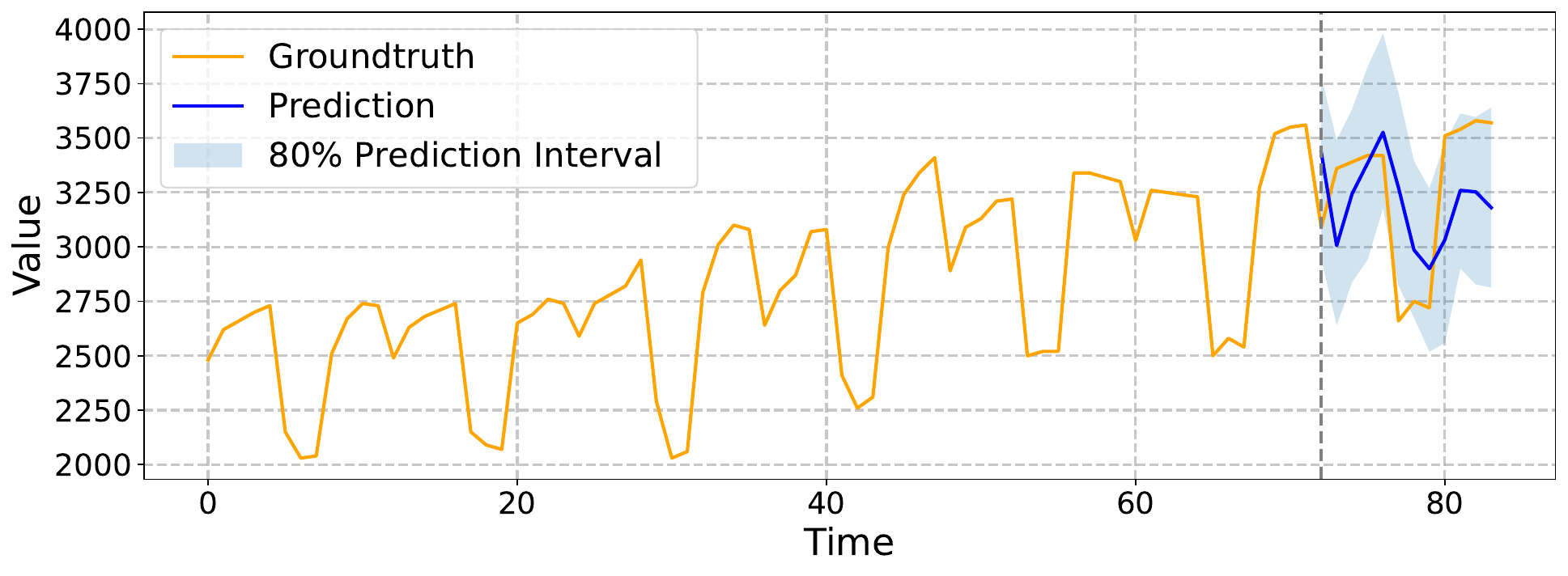}
        \caption{m4\_monthly\_dataset\_2019}
    \end{subfigure}
    \hfill
    \begin{subfigure}[b]{0.49\textwidth}
        \centering
        \includegraphics[width=\textwidth]{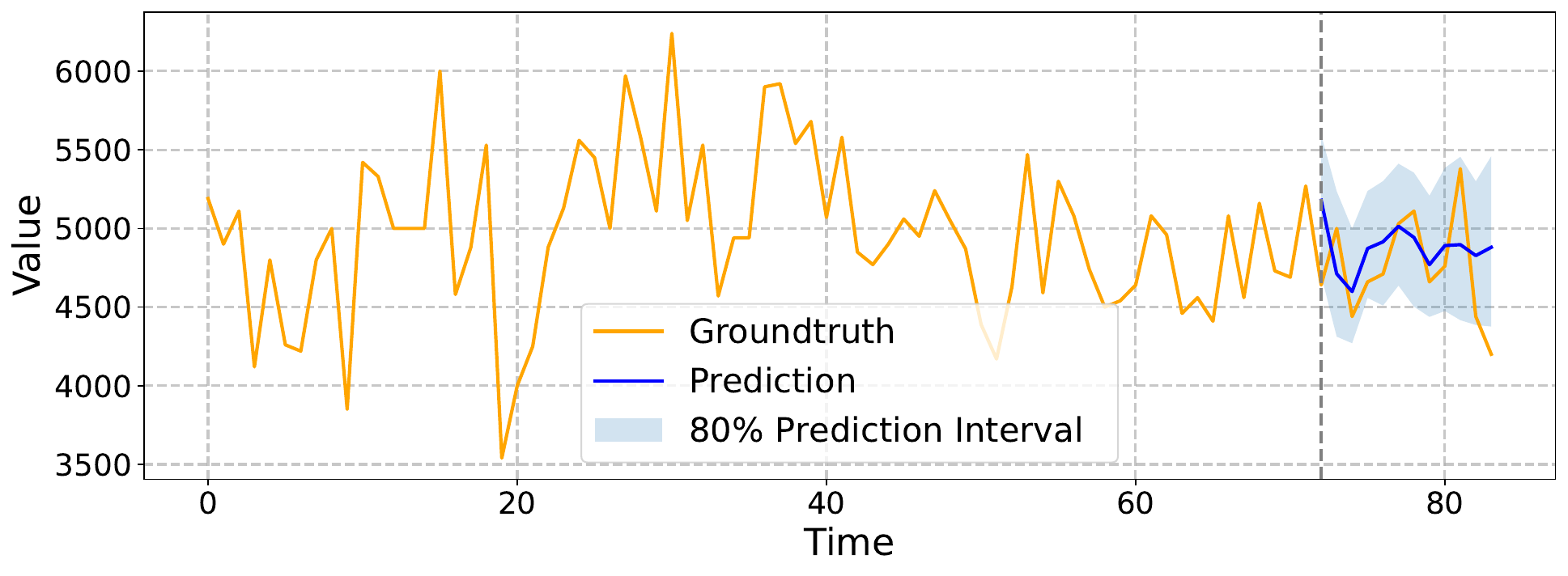}
        \caption{m4\_monthly\_dataset\_84}
    \end{subfigure}

    \begin{subfigure}[b]{0.49\textwidth}
        \centering
        \includegraphics[width=\textwidth]{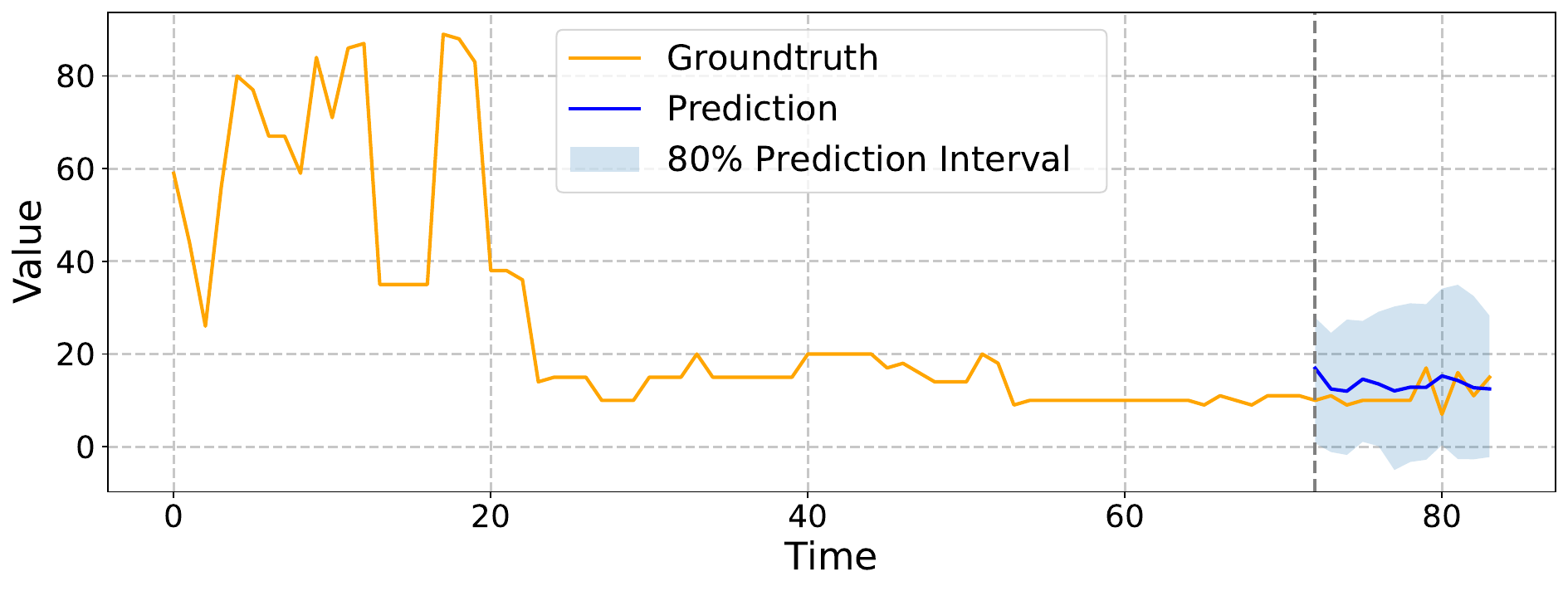}
        \caption{Vehicle trips, series 82}
    \end{subfigure}
    \hfill
    \begin{subfigure}[b]{0.49\textwidth}
        \centering
        \includegraphics[width=\textwidth]{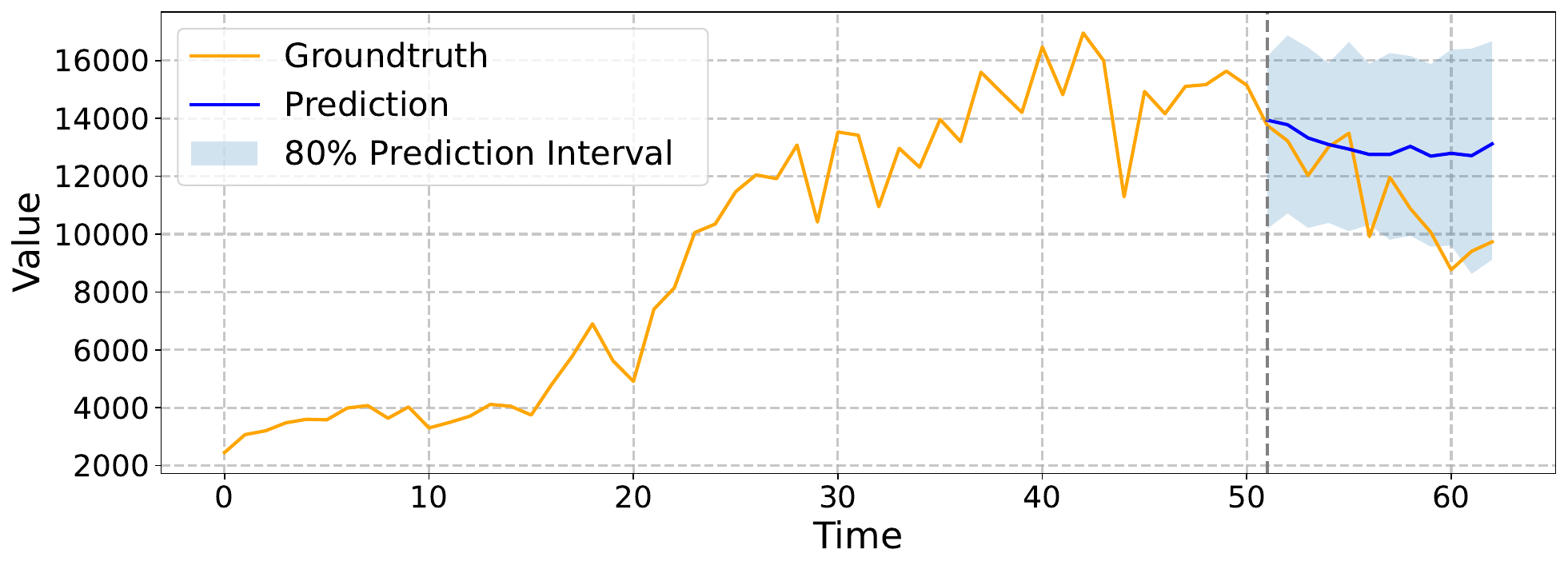}
        \caption{cif\_2016\_dataset\_65}
    \end{subfigure}
    
    \caption{Visualization of Univariate Datasets in TFB} 
    \label{fig: visualization of univariate datasets} 
\end{figure*}

We visualize the forecasting results of FLAME Large on TSFM-Bench and TFB in Figures~\ref{fig: visualization of TSFM-Bench} and \ref{fig: visualization of univariate datasets}. The figures show that FLAME can accurately model the future patterns by providing prediction intervals.

\section{Limitations}
\label{app:limitations}
Despite the strong performance of FLAME, this work still has some limitations. FLAME's local perception mechanism is particularly designed for time series with strong periodicity. For non-stationary scenarios, the mechanism may degrade, but it does not introduce adverse effects. Experiments on datasets like ILI, Exchange, NYSE, and multiple very short time series in TFB, demonstrate that FLAME also achieves state-of-the-art performance. On the other hand, FLAME is proposed for regular time series, and does not adapt to irregular time series.

\section{Impact Statement}
\label{app:impact}
This paper presents a family of time series foundation models for both deterministic and probabilistic forecasting. The release of code and experimental protocols also promotes reproducible research and further exploration of time series models.

\section{The Use of Large Language Models}
\label{app:llm_use}
LLMs were used for language editing and assistance with drafting and checking the mathematical exposition and proofs. Their use did not involve designing the model or generating experimental results.

\section{Theoretical Proof}
\label{app: theory}
\subsection{Legendre Approximation Bound}
\label{app:legt_proof}
\textbf{Theorem~\ref{theory: 1} (restated).}
Let $\theta>0$, $t\geq\theta$, and $d\geq1$. For a differentiable function $F:\mathbb{R}_+\to\mathbb{R}$, let $f_{[t-\theta,t]}$ be its optimal polynomial approximation of degree at most $d-1$ under the uniform probability measure $\mu^{(t)}$ on $[t-\theta,t]$. If $F$ is $L$-Lipschitz on this window, then
\[
\Vert F-f_{[t-\theta,t]}\Vert_{\mu^{(t)}}
    =\mathcal{O}(\theta L/\sqrt{d}).
\]
For a fixed integer $k\geq1$, if $F\in C^k([t-\theta,t])$,
$M_k=\sup_{x\in[t-\theta,t]}|F^{(k)}(x)|<\infty$, and $d\geq k$, then
\[
\Vert F-f_{[t-\theta,t]}\Vert_{\mu^{(t)}}
    =\mathcal{O}(M_k\theta^k d^{-k+1/2}),
\]
where the implied constant depends only on $k$. These bounds concern the exact projection; they do not include the finite-state recurrence or discretization errors.

\vspace{0.5em}
\noindent\textbf{Proof.} Fix a time $t$. The HiPPO-LegT framework utilizes the uniform measure $\mu^{(t)}$ supported on the sliding window $[t-\theta, t]$, with the density function $\omega(t,x) = \frac{1}{\theta}\mathbb{I}_{[t-\theta, t]}(x)$. The corresponding orthonormal polynomial basis is given by:
\[
p_n^{(t)}(x) = \sqrt{2n+1} P_n\left(\frac{2(x-t)}{\theta} + 1\right)
\]
where $P_n$ denotes the canonical Legendre polynomial of degree $n$.

Let $c_n(t) = \langle F_{[t-\theta, t]}, p_n^{(t)} \rangle_{\mu^{(t)}}$ be the projection coefficients for $n = 0, 1, \dots$. The projection $f_{[t-\theta, t]}$ is constructed by a linear combination of the basis functions up to degree $d-1$:
\[
f_{[t-\theta, t]}(x) = \sum_{n=0}^{d-1} c_n(t) p_n^{(t)}(x)
\]

Since $p_n^{(t)}(x)$ forms an orthonormal basis in the Hilbert space defined by the inner product $\langle \cdot, \cdot \rangle_{\mu^{(t)}}$, we apply Parseval's identity to express the approximation error as the sum of squared high-order coefficients:
\[
\| F_{[t-\theta, t]} - f_{[t-\theta, t]} \|_{\mu^{(t)}}^2 = \sum_{n=d}^{\infty} c_n^2(t)
\]
To bound the error, it suffices to bound $c_n(t)$ for $n \ge d$. By definition:
\[
c_n(t) = \frac{\sqrt{2n+1}}{\theta} \int_{t-\theta}^t F(x) P_n\left(\frac{2(x-t)}{\theta} + 1\right) dx
\]

Applying the change of variable $y = \frac{2(x-t)}{\theta} + 1$, we have $dx = \frac{\theta}{2} dy$. The integral domain transforms to $[-1, 1]$:
\[
c_n(t) = \frac{\sqrt{2n+1}}{2} \int_{-1}^1 F\left(t + \frac{\theta}{2}(y-1)\right) P_n(y) dy
\]

Recall the property of Legendre polynomials: $P_n(y) = \frac{1}{2n+1} \frac{d}{dy}(P_{n+1}(y) - P_{n-1}(y))$. Applying integration by parts yields:
\begin{equation*}
\begin{aligned}
c_n(t) &= \frac{\sqrt{2n+1}}{2(2n+1)} \left[ F\left(t + \frac{\theta}{2}(y-1)\right) (P_{n+1}(y) - P_{n-1}(y)) \right]_{-1}^1 \\
&\quad - \frac{\sqrt{2n+1}}{2(2n+1)} \int_{-1}^1 \frac{\theta}{2} F'\left(t + \frac{\theta}{2}(y-1)\right) (P_{n+1}(y) - P_{n-1}(y)) dy
\end{aligned}
\end{equation*}

Since $P_m(1) = 1$ and $P_m(-1) = (-1)^m$, the boundary terms evaluate to zero. The expression simplifies to:
\[
c_n(t) = - \frac{\theta}{4\sqrt{2n+1}} \int_{-1}^1 F'\left(t + \frac{\theta}{2}(y-1)\right) (P_{n+1}(y) - P_{n-1}(y)) dy
\]

\vspace{0.5em}
\noindent\textbf{Case 1: $F(x)$ is $L$-Lipschitz}\\
If $F$ is $L$-Lipschitz, its first derivative is bounded by $L$ (i.e., $|F'| \le L$). By applying the Cauchy-Schwarz inequality, we bound the squared coefficient:
\begin{equation*}
\begin{aligned}
c_n^2(t) &\le \frac{\theta^2 L^2}{16(2n+1)} \left[ \int_{-1}^1 |P_{n+1}(y) - P_{n-1}(y)| dy \right]^2 \\
&\le \frac{\theta^2 L^2}{16(2n+1)} \cdot 2 \int_{-1}^1 (P_{n+1}(y) - P_{n-1}(y))^2 dy
\end{aligned}
\end{equation*}

Due to the orthogonality of Legendre polynomials:
\[
\int_{-1}^1 (P_{n+1}(y) - P_{n-1}(y))^2 dy = \int_{-1}^1 P_{n+1}^2(y) dy + \int_{-1}^1 P_{n-1}^2(y) dy = \frac{2}{2n+3} + \frac{2}{2n-1}
\]
Thus, $c_n^2(t) \le \mathcal{O}(1) \theta^2 L^2 \frac{1}{n^2}$.
Summing over all $n \ge d$ yields:
\[
\| F_{[t-\theta, t]} - f_{[t-\theta, t]} \|_{\mu^{(t)}}^2 = \sum_{n=d}^{\infty} c_n^2(t) = \mathcal{O}(1) \theta^2 L^2 \sum_{n=d}^{\infty} \frac{1}{n^2} = \mathcal{O}\left( \frac{\theta^2 L^2}{d} \right)
\]
Taking the square root gives $\| F_{[t-\theta, t]} - f_{[t-\theta, t]} \|_{\mu^{(t)}} \le \mathcal{O}(\theta L / \sqrt{d})$, which concludes the first part of the theorem.

\vspace{0.5em}
\noindent\textbf{Case 2: $F$ has a bounded $k$-th derivative}\\
Let $F\in C^k([t-\theta,t])$ and $\|F^{(k)}\|_\infty\leq M_k$.
For $n\geq d\geq k$, define the $k$-fold antiderivative
\[
q_{n,k}(y)=\frac{1}{(k-1)!}\int_{-1}^{y}(y-u)^{k-1}P_n(u)\,du.
\]
Then $q_{n,k}^{(k)}=P_n$ and $q_{n,k}^{(j)}(-1)=0$ for $0\leq j<k$.
At $y=1$, these derivatives are integrals of $P_n$ against polynomials of degree at most $k-1<n$, so orthogonality also gives $q_{n,k}^{(j)}(1)=0$.
Consequently, all boundary terms vanish when integrating by parts $k$ times:
\[
c_n(t)=\frac{(-1)^k\sqrt{2n+1}}{2}\left(\frac{\theta}{2}\right)^k
\int_{-1}^{1}F^{(k)}\left(t+\frac{\theta}{2}(y-1)\right)q_{n,k}(y)\,dy.
\]
Repeatedly applying
$\int_{-1}^{y}P_m(u)\,du=(P_{m+1}(y)-P_{m-1}(y))/(2m+1)$ for $m\geq1$
expresses $q_{n,k}$ as a linear combination of at most $k+1$ Legendre polynomials, with indices between $n-k$ and $n+k$.
For fixed $k$, their coefficients are $\mathcal{O}_k(n^{-k})$.
Using $\|P_m\|_{L^2([-1,1])}^2=2/(2m+1)$ and absorbing the finitely many small values of $n$ into the constant gives
\[
\int_{-1}^{1}q_{n,k}^2(y)\,dy=\mathcal{O}_k(n^{-2k-1}).
\]
The Cauchy--Schwarz inequality therefore yields
\[
c_n^2(t)\leq
\frac{(2n+1)M_k^2\theta^{2k}}{2^{2k+1}}
\int_{-1}^{1}q_{n,k}^2(y)\,dy
=\mathcal{O}_k(M_k^2\theta^{2k}n^{-2k}).
\]
Summing the squared coefficients and taking the square root gives
\[
\begin{aligned}
\|F-f_{[t-\theta,t]}\|_{\mu^{(t)}}^2
  &=\sum_{n=d}^{\infty}c_n^2(t)
    =\mathcal{O}_k\left(M_k^2\theta^{2k}d^{-2k+1}\right),\\
\|F-f_{[t-\theta,t]}\|_{\mu^{(t)}}
  &=\mathcal{O}_k\left(M_k\theta^k d^{-k+1/2}\right).
\end{aligned}
\]
This completes the proof. \hfill $\blacksquare$

\input{Sections/MAF-Properties}

\clearpage
\input{Sections/Additional-Related-Work}

%% file: Sections/MAF-Properties.tex
\subsection{Properties of the Conditional MAF}
\label{app:maf_properties}

We establish the properties of the conditional affine MAF used in FLAME, following the normalizing-flow construction of \citet{maf} and the data-to-latent convention in Eq.~\eqref{eq:maf_affine}. The proofs below concern the transformation within one target patch; they do not require the conditioning network itself to be invertible.

\paragraph{Notation and assumptions.}
Fix a patch index $i$ and a conditioning token $q=q_i$. Let $p,K\geq1$ be integers. To simplify notation, write $y=\mathrm{Y}_i$, $z^{(0)}=y$, and $z^{(j)}=T_j(z^{(j-1)};q)$. For $x\in\mathbb{R}^p$, the $j$-th layer is
\begin{equation}
    [T_j(x;q)]_k
    =e^{a_{j,k}(x_{<k},q)}x_k+b_{j,k}(x_{<k},q),
    \qquad k=1,\ldots,p,
    \label{eq:maf_layer_appendix}
\end{equation}
where $x_{<k}=(x_1,\ldots,x_{k-1})$. The functions $a_{j,k}$ and $b_{j,k}$ are the masked MLP outputs, whose evaluations at $z^{(j-1)}_{<k}$ are $\alpha_{i,k}^{(j)}$ and $\beta_{i,k}^{(j)}$ in the main text. For $k=1$, the prefix is empty and both functions depend only on $q$.

We assume that these functions are finite and real-valued for every finite prefix, and that the effective mask enforces dependence only on $x_{<k}$ and $q$. All layers use the same coordinate order as in Eq.~\eqref{eq:maf_affine}. For the classical Jacobian and change-of-variables arguments, we additionally assume that the functions are continuously differentiable in the prefix. The token $q$ is held fixed throughout all derivatives and inversions; it is not recomputed from the target patch being transformed. Bijectivity itself only requires the finite-valued, strictly autoregressive parameterization.

\begin{proposition}[Global invertibility]
\label{prop:maf_inverse}
For each fixed $q$, every layer $T_j(\cdot;q)$ in Eq.~\eqref{eq:maf_layer_appendix} is a bijection from $\mathbb{R}^p$ to $\mathbb{R}^p$. Its inverse is uniquely determined by the recursion
\begin{equation}
    x_k=e^{-a_{j,k}(x_{<k},q)}
        \bigl(z_k-b_{j,k}(x_{<k},q)\bigr),
    \qquad k=1,\ldots,p,
    \label{eq:maf_inverse_appendix}
\end{equation}
where $z=T_j(x;q)$. Consequently, the full map $T=T_K\circ\cdots\circ T_1$ is bijective, with
\begin{equation}
    T^{-1}=T_1^{-1}\circ\cdots\circ T_K^{-1}.
    \label{eq:maf_composite_inverse}
\end{equation}
If the parameter functions are continuous, both $T$ and $T^{-1}$ are continuous. Under the continuous differentiability assumption, they are continuously differentiable as well.
\end{proposition}

\begin{proof}
Fix a layer $j$ and an arbitrary desired output $z\in\mathbb{R}^p$. For the first coordinate, $a_{j,1}(q)$ and $b_{j,1}(q)$ are known constants. Since $e^{a_{j,1}(q)}>0$, the equation $z_1=e^{a_{j,1}(q)}x_1+b_{j,1}(q)$ has exactly one real solution, namely Eq.~\eqref{eq:maf_inverse_appendix} for $k=1$.

Suppose that $x_1,\ldots,x_{k-1}$ have already been recovered uniquely. The strict autoregressive dependence makes $a_{j,k}(x_{<k},q)$ and $b_{j,k}(x_{<k},q)$ known at this step. They are finite, so the coefficient $e^{a_{j,k}(x_{<k},q)}$ is nonzero. The equation for $z_k$ therefore has the unique finite solution in Eq.~\eqref{eq:maf_inverse_appendix}. Induction recovers all $p$ coordinates.

This construction provides a preimage for every $z$, proving surjectivity. Any two preimages must agree in the first coordinate, then the second, and so on by the same recursion, proving injectivity. Substitution into Eq.~\eqref{eq:maf_layer_appendix} gives $T_j(T_j^{-1}(z;q);q)=z$; uniqueness also gives $T_j^{-1}(T_j(x;q);q)=x$.

A composition of these bijections is bijective. Cancelling adjacent inverse pairs yields Eq.~\eqref{eq:maf_composite_inverse}: generation starts from $z^{(K)}=\epsilon$ and applies $T_K^{-1},T_{K-1}^{-1},\ldots,T_1^{-1}$ in that execution order. Finally, continuity, or continuous differentiability, of the inverse follows inductively from Eq.~\eqref{eq:maf_inverse_appendix}, since exponentiation and the finite algebraic operations preserve these properties. Composition gives the corresponding regularity of $T$ and $T^{-1}$.
\end{proof}

\begin{proposition}[Triangular Jacobian and exact conditional likelihood]
\label{prop:maf_density}
Under the continuous differentiability assumption, each $T_j(\cdot;q)$ has a lower triangular Jacobian with a strictly positive diagonal. At $z^{(0)}=y$ and $\epsilon=T(y;q)$, the full transformation satisfies
\begin{equation}
\begin{aligned}
    \log\left|\det D_yT(y;q)\right|
      &=\sum_{j=1}^{K}\sum_{k=1}^{p}
          a_{j,k}(z^{(j-1)}_{<k},q),\\
    p_\theta(y\mid q)
      &=p_\epsilon(T(y;q))\left|\det D_yT(y;q)\right|.
\end{aligned}
\label{eq:maf_density_appendix}
\end{equation}
For any normalized base density $p_\epsilon$, this defines a normalized conditional density. For the standard Gaussian base used in FLAME,
\begin{equation}
    \log p_\theta(y\mid q)
      =-\frac{p}{2}\log(2\pi)-\frac{1}{2}\|\epsilon\|_2^2
        +\sum_{j=1}^{K}\sum_{k=1}^{p}
          a_{j,k}(z^{(j-1)}_{<k},q).
    \label{eq:maf_gaussian_likelihood}
\end{equation}
\end{proposition}

\begin{proof}
Suppress the layer index and write $a_k=a_{j,k}(x_{<k},q)$ and $b_k=b_{j,k}(x_{<k},q)$. Differentiating Eq.~\eqref{eq:maf_layer_appendix} while holding $q$ fixed gives
\begin{equation}
    \frac{\partial[T_j(x;q)]_k}{\partial x_r}
      =\begin{cases}
          e^{a_k}x_k\,\dfrac{\partial a_k}{\partial x_r}
            +\dfrac{\partial b_k}{\partial x_r}, & r<k,\\[4pt]
          e^{a_k}, & r=k,\\
          0, & r>k.
        \end{cases}
    \label{eq:maf_jacobian_entries}
\end{equation}
The diagonal formula uses the strict mask: neither $a_k$ nor $b_k$ depends on $x_k$. Thus the parameter dependency mask is strictly lower triangular, whereas the transformation Jacobian has a nonzero diagonal. Its determinant is
\begin{equation}
    \det D_xT_j(x;q)
      =\prod_{k=1}^{p}e^{a_{j,k}(x_{<k},q)}
      =\exp\!\left(\sum_{k=1}^{p}a_{j,k}(x_{<k},q)\right)>0.
    \label{eq:maf_layer_determinant}
\end{equation}
By the chain rule, $D_yT$ is the product of the layer Jacobians, evaluated at $z^{(0)},\ldots,z^{(K-1)}$. Determinants multiply, and taking logarithms proves the first identity in Eq.~\eqref{eq:maf_density_appendix}. This also shows that the composition remains nonsingular and orientation-preserving.

Proposition~\ref{prop:maf_inverse} establishes a global continuously differentiable inverse. The change-of-variables formula therefore gives the second identity in Eq.~\eqref{eq:maf_density_appendix}. In particular,
\begin{equation}
\begin{aligned}
    \int_{\mathbb{R}^p}p_\theta(y\mid q)\,dy
      &=\int_{\mathbb{R}^p}p_\epsilon(T(y;q))
           \left|\det D_yT(y;q)\right|\,dy\\
      &=\int_{\mathbb{R}^p}p_\epsilon(\epsilon)\,d\epsilon=1.
\end{aligned}
\label{eq:maf_normalization}
\end{equation}
Substituting the standard Gaussian log density yields Eq.~\eqref{eq:maf_gaussian_likelihood}. The Jacobian term has a plus sign because $T$ maps data to the base variable. Restoring the patch index $i$ and averaging the negative log densities over $i=1,\ldots,l$ gives exactly the objective in Eq.~\eqref{eq:flow_nll}.
\end{proof}

\begin{corollary}[Autoregressive order and exact sampling]
\label{cor:maf_order}
For fixed $q$, the $k$-th coordinate of $T(y;q)$ depends only on $y_1,\ldots,y_k$, and the $k$-th coordinate of $T^{-1}(\epsilon;q)$ depends only on $\epsilon_1,\ldots,\epsilon_k$. Each coordinate map is strictly increasing in its own coordinate when the preceding coordinates are fixed. With a standard Gaussian base, the conditional density admits the autoregressive factorization
\begin{equation}
\begin{aligned}
    p_\theta(y\mid q)
      &=\prod_{k=1}^{p}p_\theta(y_k\mid y_{<k},q),\\
    p_\theta(y_k\mid y_{<k},q)
      &=\varphi([T(y;q)]_k)
          \frac{\partial[T(y;q)]_k}{\partial y_k},
\end{aligned}
\label{eq:maf_autoregressive_density}
\end{equation}
where $\varphi$ is the standard univariate Gaussian density. Drawing $\epsilon\sim\mathcal{N}(0,I)$ and evaluating $T^{-1}(\epsilon;q)$ produces a sample from this conditional density.
\end{corollary}

\begin{proof}
The prefix property holds for each layer by Eq.~\eqref{eq:maf_layer_appendix}. If the first $k$ coordinates of $z^{(j-1)}$ depend only on $y_{\leq k}$, the same holds for the first $k$ coordinates of $z^{(j)}$. Induction over layers proves the claim for $T$. Equation~\eqref{eq:maf_inverse_appendix} gives the corresponding property for each inverse layer, and hence for their composition. Products of lower triangular matrices are lower triangular, and their diagonal entries multiply. Therefore,
\begin{equation}
    \frac{\partial[T(y;q)]_k}{\partial y_k}
      =\exp\!\left(\sum_{j=1}^{K}
          a_{j,k}(z^{(j-1)}_{<k},q)\right)>0.
    \label{eq:maf_monotonicity}
\end{equation}
For a fixed $y_{<k}$, all intermediate prefixes $z^{(j)}_{<k}$ are fixed. The map from $y_k$ to $[T(y;q)]_k$ is then a composition of affine maps with positive slopes, so it is strictly increasing and onto $\mathbb{R}$. The inverse is also coordinatewise increasing: its diagonal Jacobian entries are the reciprocals of those in Eq.~\eqref{eq:maf_monotonicity}, evaluated at $y=T^{-1}(\epsilon;q)$.

The standard Gaussian base factorizes over coordinates, and the triangular determinant is the product of the diagonal derivatives. Substitution into Eq.~\eqref{eq:maf_density_appendix} yields the product in Eq.~\eqref{eq:maf_autoregressive_density}. Each factor depends only on $y_{\leq k}$ and $q$, and a one-dimensional change of variable gives
\begin{equation}
    \int_{\mathbb{R}}\varphi([T(y;q)]_k)
       \frac{\partial[T(y;q)]_k}{\partial y_k}\,dy_k
       =\int_{\mathbb{R}}\varphi(u)\,du=1
    \label{eq:maf_conditional_normalization}
\end{equation}
for every fixed prefix. Integrating the later coordinates in reverse order confirms that these factors are the conditional densities of the joint model. Finally, for any measurable set $A\subseteq\mathbb{R}^p$, the change of variables $\epsilon=T(y;q)$ gives
\[
    \Pr\{T^{-1}(\epsilon;q)\in A\}
       =\int_A p_\epsilon(T(y;q))|\det D_yT(y;q)|\,dy
       =\int_A p_\theta(y\mid q)\,dy,
\]
which proves the sampling claim.
\end{proof}

\paragraph{Computational consequences.}
When evaluating a known target patch, every input coordinate to a layer is available. A masked MLP therefore produces all $p$ pairs of affine parameters in one forward pass, and density evaluation uses $K$ such passes plus the sum of $Kp$ log-scales. The latter sum does not require constructing a dense Jacobian; its cost is separate from the MLP evaluation cost. During generation, Eq.~\eqref{eq:maf_inverse_appendix} requires the recovered prefix before the next coordinate can be computed. A direct implementation has $p$ sequential coordinate updates per inverse layer, applied through the $K$ layers in reverse order. Independent draws with fixed conditioning can still be batched. These are the density-evaluation and sampling trade-offs of MAF~\citep{maf}.

\paragraph{Scope and regularity.}
The ordering established above is conditional on $q$ and concerns coordinates within a patch; it is not a claim of causal identification. The normalized density and exact likelihood concern complete $p$-dimensional patches. Truncating an output patch gives a marginal distribution on the retained coordinates and does not make the truncation invertible. If the parameter networks are locally Lipschitz rather than continuously differentiable, both the forward and recursive inverse maps are locally Lipschitz; the Jacobian and density identities then hold almost everywhere by the change-of-variables formula for such bijections. Finally, mathematical invertibility does not by itself guarantee good numerical conditioning: extremely small scales or large off-diagonal derivatives can amplify roundoff errors. The propositions establish structural properties of the specified flow, not universal approximation of arbitrary conditional densities.

%% file: Sections/Additional-Related-Work.tex
\section{Additional Related Work}
\label{app:additional-related-work}

\paragraph{Learning compact temporal representations.}
FLAME's Legendre memory centers on retaining predictive information in a compact state. The broader forecasting literature examines representation design through selective patch spaces~\citep{wu2025srsnet} and supervision through decomposition-based losses~\citep{qiu2025DBLoss}. The search for economical designs also extends to irregular observations, where a simple forecasting baseline has been studied~\citep{liu2026rethinking}. Joint time--frequency representations~\citep{qiu2026bridging} and adaptive spatio-temporal graph interactions~\citep{liu2026astgi} offer further perspectives on encoding such data. These approaches place temporal memory within a wider set of choices about what information a forecaster represents and how that representation is learned.

\paragraph{Predictive uncertainty and available information.}
The distribution of future values depends on the information available to the predictor. GinAR+~\citep{yu2025ginarp} considers forecasting from incomplete multivariate series, and Merlin~\citep{yu2025merlin} develops multi-view representations for varying missing rates. Additional information enters through exogenous variables in DAG's dual correlation network~\citep{qiu2026dag}, GCGNet's graph-consistent generative network~\citep{li2026gcgnet}, and KITE's knowledge-guided probabilistic model~\citep{cheng2026kite}. At the application level, point--interval battery lifetime prediction~\citep{cheng2026metagnsdformer} illustrates how uncertainty estimates accompany a specific prognostic target. These settings motivate studying efficient distributional prediction under different information constraints, alongside FLAME's generative forecasting head.

\paragraph{Temporal structure across objectives.}
Dependencies among variables are relevant to both prediction and detection. CCD uses deformable convolutions to capture forecasting cross-correlations~\citep{cheng2026ccd}, whereas CATCH investigates channel-aware frequency patches for identifying anomalies~\citep{wu2024catch}. Foundation-model adaptation likewise spans multivariate forecasting in CoRA~\citep{cheng2026cora} and anomaly detection in STAR~\citep{cheng2026star}. TAB supplies a unified benchmark for the latter task~\citep{qiu2025tab}; TimeART broadens the task landscape through agentic reasoning augmented with tools~\citep{wu2026timeart}. This literature raises complementary questions about which temporal regularities can be reused when the learning objective changes.

\paragraph{Generation with explicit controls.}
FLAME models numerical futures, while research on visual sequences studies other forms of conditional generation. The controllable video generation survey~\citep{ma2025controllable} provides an overview of this field. Individual examples include Follow Your Pose for pose-guided synthesis~\citep{ma2024followpose} and Follow-Your-Emoji for expressive portrait animation~\citep{ma2024followyouremoji}. LiveLight addresses interactive relighting of streaming video in real time~\citep{ma2026livelight}. Their output modalities and control mechanisms give a broader perspective on generative modeling, in which computational cost and the specification of desired outputs are also central design considerations.

%% file: Sections/checklist.tex
\newpage
\clearpage
\section*{NeurIPS Paper Checklist}

\begin{enumerate}

\item {\bf Claims}
    \item[] Question: Do the main claims made in the abstract and introduction accurately reflect the paper's contributions and scope?
    \item[] Answer: \answerYes{} 
    \item[] Justification: The main claims made in the abstract and introduction accurately reflect the paper's contributions and scope.

\item {\bf Limitations}
    \item[] Question: Does the paper discuss the limitations of the work performed by the authors?
    \item[] Answer: \answerYes{} 
    \item[] Justification: We discuss the limitations in Appendix~\ref{app:limitations}.

\item {\bf Theory assumptions and proofs}
    \item[] Question: For each theoretical result, does the paper provide the full set of assumptions and a complete (and correct) proof?
    \item[] Answer: \answerYes{} 
    \item[] Justification: Appendix~\ref{app:legt_proof} proves Theorem~\ref{theory: 1}. Appendix~\ref{app:maf_properties} states the assumptions and proves the conditional MAF's invertibility, Jacobian formula, normalized likelihood, and autoregressive sampling properties.

    \item {\bf Experimental result reproducibility}
    \item[] Question: Does the paper fully disclose all the information needed to reproduce the main experimental results of the paper to the extent that it affects the main claims and/or conclusions of the paper (regardless of whether the code and data are provided or not)?
    \item[] Answer: \answerYes{} 
    \item[] Justification: We set a fixed random seed to ensure deterministic results across all runs.

\item {\bf Open access to data and code}
    \item[] Question: Does the paper provide open access to the data and code, with sufficient instructions to faithfully reproduce the main experimental results, as described in supplemental material?
    \item[] Answer: \answerYes{} 
    \item[] Justification: We provide an anonymous repository link for releasing the model code, and the complete reproducible model code is already available.

\item {\bf Experimental setting/details}
    \item[] Question: Does the paper specify all the training and test details (e.g., data splits, hyperparameters, how they were chosen, type of optimizer) necessary to understand the results?
    \item[] Answer: \answerYes{} 
    \item[] Justification: We provide full experimental settings in Appendix~\ref{app: exp}.

\item {\bf Experiment statistical significance}
    \item[] Question: Does the paper report error bars suitably and correctly defined or other appropriate information about the statistical significance of the experiments?
    \item[] Answer: \answerNo{} 
    \item[] Justification: We use a fixed random seed but do not report error bars or statistical significance tests across independent runs.

\item {\bf Experiments compute resources}
    \item[] Question: For each experiment, does the paper provide sufficient information on the computer resources (type of compute workers, memory, time of execution) needed to reproduce the experiments?
    \item[] Answer: \answerYes{} 
    \item[] Justification: We provide the corresponding information in Appendix~\ref{app: exp}.
    
\item {\bf Code of ethics}
    \item[] Question: Does the research conducted in the paper conform, in every respect, with the NeurIPS Code of Ethics \url{https://neurips.cc/public/EthicsGuidelines}?
    \item[] Answer: \answerYes{} 
    \item[] Justification: We ensure that our research conforms with the NeurIPS Code of Ethics.

\item {\bf Broader impacts}
    \item[] Question: Does the paper discuss both potential positive societal impacts and negative societal impacts of the work performed?
    \item[] Answer: \answerYes{} 
    \item[] Justification: We discuss broader impacts in Appendix~\ref{app:impact}.
    
\item {\bf Safeguards}
    \item[] Question: Does the paper describe safeguards that have been put in place for responsible release of data or models that have a high risk for misuse (e.g., pre-trained language models, image generators, or scraped datasets)?
    \item[] Answer: \answerNA{} 
    \item[] Justification: Our research does not involve these concerns.

\item {\bf Licenses for existing assets}
    \item[] Question: Are the creators or original owners of assets (e.g., code, data, models), used in the paper, properly credited and are the license and terms of use explicitly mentioned and properly respected?
    \item[] Answer: \answerYes{} 
    \item[] Justification: We follow the licences of open-source code and datasets.

\item {\bf New assets}
    \item[] Question: Are new assets introduced in the paper well documented and is the documentation provided alongside the assets?
    \item[] Answer: \answerYes{} 
    \item[] Justification: New assets introduced in the paper are well documented.

\item {\bf Crowdsourcing and research with human subjects}
    \item[] Question: For crowdsourcing experiments and research with human subjects, does the paper include the full text of instructions given to participants and screenshots, if applicable, as well as details about compensation (if any)? 
    \item[] Answer: \answerNA{} 
    \item[] Justification: Our research does not involve these concerns.

\item {\bf Institutional review board (IRB) approvals or equivalent for research with human subjects}
    \item[] Question: Does the paper describe potential risks incurred by study participants, whether such risks were disclosed to the subjects, and whether Institutional Review Board (IRB) approvals (or an equivalent approval/review based on the requirements of your country or institution) were obtained?
    \item[] Answer: \answerNA{} 
    \item[] Justification: Our research does not involve these concerns.

\item {\bf Declaration of LLM usage}
    \item[] Question: Does the paper describe the usage of LLMs if it is an important, original, or non-standard component of the core methods in this research? Note that if the LLM is used only for writing, editing, or formatting purposes and does \emph{not} impact the core methodology, scientific rigor, or originality of the research, declaration is not required.
    \item[] Answer: \answerYes{} 
    \item[] Justification: Appendix~\ref{app:llm_use} discloses the use of LLMs for language editing and assistance with mathematical exposition and proofs.

\end{enumerate}